\documentclass[default,iicol]{sn-jnl}

\usepackage{amsmath,amssymb,amsfonts}
\usepackage{graphicx}

\usepackage{textcomp}
\usepackage{wrapfig}
\usepackage{makecell}
\usepackage{soul}
\usepackage{multirow}

\usepackage[utf8]{inputenc} 
\usepackage[T1]{fontenc}    
\usepackage{hyperref}       
\usepackage{url}            
\usepackage{booktabs}       
\usepackage{amsfonts}       
\usepackage{nicefrac}       
\usepackage{microtype}      
\usepackage{lipsum}
\usepackage{graphicx}
\usepackage{multirow}
\usepackage{array}
\usepackage{subcaption}
\usepackage{color, colortbl}
\usepackage{placeins}
\usepackage{enumitem}
\usepackage{gensymb}

\usepackage{tabularx}

\usepackage{url}
\jyear{2022}%

\theoremstyle{thmstyleone}%
\theoremstyle{thmstyletwo}%

\theoremstyle{thmstylethree}%

\begin{document}

\title[A Survey on Infrared Image \& Video Sets]{A Survey on Infrared Image \& Video Sets}

\author*[1]{\fnm{Kevser Irem} \sur{Danaci}}\email{kiremdanaci@sivas.edu.tr}

\author[2]{\fnm{Erdem} \sur{Akagunduz}}\email{akaerdem@metu.edu.tr}

\affil*[1]{\orgdiv{Department of Electrical Eng.}, \orgname{Sivas University of Science And Technology}, \orgaddress{
\country{Turkey}}}

\affil[2]{\orgdiv{The Graduate School of Informatics}, \orgname{Middle East Technical University}, \orgaddress{
\country{Turkey}}}


\abstract{In this survey, we compile a list of publicly available infrared image and video sets for artificial intelligence and computer vision researchers. We mainly focus on IR image and video sets, which are collected and labelled for computer vision applications such as object detection, object segmentation, classification, and motion detection. We categorise {109} publicly available or private sets according to their sensor types, image resolution, and scale. We describe each set in detail regarding their collection purpose, operation environment, optical system properties, and application area. We also cover a general overview of fundamental concepts related to IR imagery, such as IR radiation, IR detectors, IR optics and application fields. We analyse the statistical significance of the entire corpus from different perspectives. This survey will be a guideline for computer vision and artificial intelligence researchers who want to delve into working with the spectra beyond the visible domain.}


\maketitle

\section{Introduction}
\label{sec:introduction}
{Artificial intelligence is based on data, which is the new defining element of science. In the past decade, machine learning techniques have evolved to the point where they are now capable of processing larger data sets than humans could ever imagine or possess. Particularly in the field of computer vision, large-scale data sets improve machine learning performance so dramatically that deep neural networks are able to perform as well as humans on especially high-quality images} \cite{DodgeK17b}. {The amount of labelled visual data available for various computer vision tasks (such as image classification, segmentation, detection, tracking, etc.) has reached billions of high-quality images} \cite{Zhang13MS} {available worldwide for use by researchers and engineers.}

{Visual data that is publicly accessible comes in a variety of formats. Although the available data is overwhelmingly composed of the visible band, or in other words, ``RGB'' images; public access to images of other modalities, such as multi/hyperspectral, magnetic resonance (MR), computerised tomography (CT), synthetic aperture radar (SAR), to name a few, is also possible. One relatively less public imaging modality is the infrared (IR) imagery, which corresponds to images constructed with the radiation of an invisible portion of the electromagnetic spectrum, known as the infrared band.}

 All kinds of objects emit infrared radiation \cite{HAMAMATSUPHOTONICSK.K.2011}. With its low radiation absorption, high contrast, and capacity for hot target detection, the IR band is popular and practical for use in civil and military applications \cite{pineiro2014target}. IR imaging is used in many applications, such as object detection, object segmentation, classification, motion detection, etc. However, in contrast to visible band imagery, IR images are difficult to access for several reasons. To begin with, the technology of most IR imaging systems is relatively expensive for use in consumer electronics. Besides, since most IR vision applications are utilised for military or medical applications, they are inaccessible due to either security reasons or intellectual property rights. As a result, the publicly available infrared image and video sets are limited compared to high-scale labelled visible band image and video sets.

The primary purpose of this article is to compile a list of publicly available infrared image and video sets for artificial intelligence and computer vision researchers. We mainly focus\footnote{Multispectral image sets collected with satellites are left out of the scope of this survey paper. We believe that multispectral satellite imagery is a category that requires a unique focus due to differences in IR imaging in vision practices, perspective, atmospheric effects and applications.} on IR image and video sets which are collected and labelled for computer vision applications such as object detection, object segmentation, classification, and motion detection. We categorize {109} different publicly available or private sets according to their sensor types, image resolution, and scale. We describe each and every set in detail regarding their collection purpose, operation environment, optical system properties, and area of application. 

{The number of survey studies on IR vision algorithms and IR vision technologies is increasing} \cite{teutsch2021computer,DengBaoyuan2021Imva,Perpetuini2021,Hu2022}. {However, to the best of our knowledge, no published survey studies that review IR image or video sets exist. Our aim is to compile a collection of sets so that researchers in the fields of computer vision and deep learning can identify a visual corpus with necessary properties and compare it with other sets already available. As a result, we believe that the survey can contribute to new algorithms in deep learning and vision research using the spectra beyond the visible spectrum. By scanning public academic sources, we compile this list of image and video sets collected using IR imaging equipment. What is more,} for the reader to completely evaluate the different properties of IR image and video sets, we also provide a background on the fundamentals of infrared imagery, including topics such as principles of infrared radiation, infrared sensors, infrared optics, and application fields of IR imagery.

The remainder of this paper is organised as follows: Section 2 covers a general overview of IR radiation, IR detectors, IR optics and related applications. Section 3 starts with an analysis of the statistical significance of the entire corpora and follows by providing the compiled sets as a list with brief descriptions. Finally, Section 4 procures conclusions and sets future directions for the paper.

\section{Fundamentals of Infrared Imagery}

\subsection{Infrared Radiation}
The discovery of IR radiation dates back to an experiment by Frederick William Herschel more than 200 years ago using prisms and basic temperature sensors to measure the wavelength distribution of the stellar spectra \cite{clerke1908popular}. However, its widespread use is relatively new, starting by the early 20{th} century with the understanding of Plank's law and blackbody radiation, and also with the help of modern physics and quantum theory \cite{karim2013infrared,buser1997historical}. Today it is almost common knowledge that according to specific known laws of physics,  objects emit unique radiation in a broad region of wavelengths called the electromagnetic spectrum (ES). The IR region of this spectrum corresponds to wavelengths from the nominal red edge of the visible spectrum around 700 nanometers to 1 millimetre. IR wavelengths in this region are conventionally categorised into five spectral sub-bands. The wavelength region of 0.7$\mu$m to 1.4 $\mu$m is called the near-infrared (NIR), 1.4$\mu$m to 5$\mu$m: the short-wave infrared (SWIR), 3$\mu$m to 8$\mu$m: the mid-wave infrared (MWIR),  8 $\mu$m to 15 $\mu$m the long-wave infrared (LWIR), and finally 15$\mu$m to 1000$\mu$m the far-infrared (FIR) (see Table \ref{tab1}).

\begin{figure}[t]\centering
\includegraphics[trim=120 0 100 0,width=1.0\columnwidth]{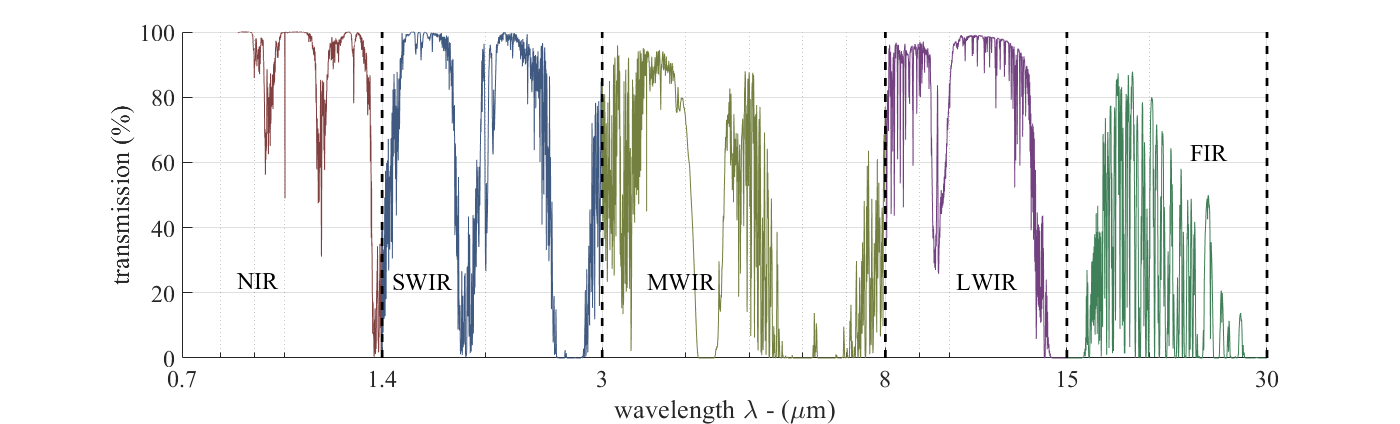}
\caption[]{Synthetic spectrum of atmospheric transmission between 0.7-30 $\mu$m, created with ATRAN module \cite{ATRAN1992}. }
\label{Figure1}
\end{figure}

The conventional categorization of IR sub-bands defined in Table \ref{tab1} is correlated with how IR radiation is absorbed, reflected or transmitted by the atmosphere. The region of the IR spectrum, where there is relatively little absorption of terrestrial thermal radiation by atmospheric gases, is called the IR atmospheric window, which is roughly between 1 to 15 $\mu$m. The absorption of IR radiation depends on various atmospheric conditions such as altitude, latitude, solar Zenith angle, water vapour, etc. In Figure \ref{Figure1}, a synthetically created spectrum of atmospheric transmission between 0.7-30 $\mu$m, using the ATRAN module\footnote{ATRAN module input parameters are selected as, observatory altitude: 13800 feet (Mauna Kea (red) at an altitude of 13.8K feet and 3.4 mm water vapour), observatory latitude: 39 degrees, water vapour overburden: 0 microns, standard atmosphere with 2 Layers, Zenith angle: 45 degrees, smoothing resolution: 1000.} \cite{ATRAN1992} is depicted. For instance, as seen in Figure \ref{Figure1}, atmospheric transmittance of the NIR spectrum band is relatively high, which makes this sub-band an effective spectrum for \emph{active} (i.e. a radiation source illuminating the scene) night vision systems.

\begin{table}
\centering
\caption{The IR Spectrum}
\label{Table1}
\begin{tabular}{p{90pt}|p{80pt}}
\toprule
Wavelength     & Designation      \\\midrule
    $10^-6$$\mu$m to $10^-2$$\mu$m  & x rays      \\
    $10^-2$$\mu$m to $0.4$$\mu$m   & ultraviolet  \\
    0.4$\mu$m to 0.7 $\mu$m & visible \\ \hline
    \textbf{0.7$\mu$m to 1.4$\mu$m} & \textbf{NIR} \\ \hline
    \textbf{1.4$\mu$m to 3$\mu$m}    & \textbf{SWIR} \\ \hline
    \textbf{3$\mu$m to 8$\mu$m}    & \textbf{MWIR} \\ \hline
    \textbf{8$\mu$m to 15$\mu$m}    & \textbf{LWIR} \\ \hline
    \textbf{15$\mu$m to 1mm}    &\textbf{FIR} \\ \hline
    1mm to 1m & microwaves \\
    1m to 10km & radiowaves \\ \bottomrule
\end{tabular}
\label{tab1}
\end{table}

It is also seen in Figure \ref{Figure1} that much of the IR spectrum is not suitable for everyday applications because IR radiation is absorbed by water or carbon dioxide in the atmosphere. However, there are a number of wavelength bands with low absorption, which actually create the IR sub-bands known as the short, medium and long-wavelength IR bands, abbreviated as SWIR, MWIR and LWIR respectively.

 Visible, NIR or SWIR light (0.35-3 $\mu$m) corresponds to a high atmospheric transmission band and peak solar illumination. This is why most optical systems usually include detectors sensitive to these bands for the best clarity and resolution. However, without moonlight or artificial illumination, SWIR imaging systems are known to provide poor or no imagery of objects below 300K temperatures. SWIR imaging systems predominantly use reflected light. Accordingly, they are comparable to grey-scale visible images in resolution and detail.

The MWIR (also referred to as the `MIR') band also provides partial regions of lossless atmospheric transmission with the added benefit of reduced ambient and background noise. This region is referred to as the ``thermal infrared''. The radiation in this sub-band is emitted from the object itself; hence passive imaging is utilised. Two principal factors determine how bright an object appears in the MWIR spectrum: the object’s temperature and its emissivity (E). Emissivity is a physical property of materials that describes how efficiently it radiates the absorbed radiation.

The LWIR band spans roughly between 8-15 $\mu$m, with almost no atmospheric absorption between the 9-12 $\mu$m region. Because LWIR sensors can construct an image of a scene based on passive thermal emissions only and hence require no active illumination, this region is also considered as ``thermal infrared''. LWIR band is better than MWIR for imaging through smoke or atmospheric particles (aerosols). Therefore, surveillance applications usually prefer LWIR technology. On the other hand, for very long-range detection (such as 10km or more), MWIR has greater atmospheric transmission than LWIR in most atmospheric conditions.

Although the FIR spectrum is defined between 0.75$\mu$m and 1mm,  the atmosphere absorbs almost all IR radiation with wavelengths above 25$\mu$m. Hence, atmospheric FIR spectroscopy can only be effectively utilised for wavelengths in the limited spectrum between 0.75 to 25 $\mu$m. This region is also an atmospheric thermal band, which we can experience in the form of heat waves. For astronomical observation outside of the atmosphere, the entire FIR spectrum is utilised.

For a general overview of the subject and the fundamentals of radiometry, the reader may refer to \cite{parr2005optical}. 

\subsection{Infrared Detectors}
One of the fundamental parts of an IR electro-optical system is the detecting sensor. In order to capture the IR signature of a scene, a detector sensitive to IR radiation is needed. IR-sensitive detectors capture the IR radiation emitted by the objects and the scene, and convert it into electrical signals. Objects that have different temperatures and emissivity, emit different levels of radiation so that the camera produces electrical signals that have different amplitudes. These electrical signals are used to produce the IR image.

Detectors are the core of an IR imaging system. Historically IR detectors can be scrutinised in three generations. The first generation consists of single-cell detectors. In order to create an image plane, the infrared beam emitted from a scene reaches a reflective surface (i.e. mirror). As the position of the mirror is deflected by two-dimensional rotary actuators, the focused infrared beam creates a two-dimensional pattern of the target image plane. In contrast, the second-generation systems comprise an array of detectors with an optical mirror system that rotates only on a single axis. Finally, the modern third-generation IR optical systems have two-dimensional array detectors, known as focal plane arrays (FPA), so that the system does not need a mirror system to scan different parts of the scene \cite{bagavathiappan2013infrared,HAMAMATSUPHOTONICSK.K.2011}.Third-generation IR detectors are quite similar to modern digital photographing machines in principle. 

In order to measure IR detector performance, three principle metrics are utilised: photosensitivity (or responsivity), noise-equivalent-power (NEP), and Detectivity (D*). 

Photosensitivity or responsivity is defined as the output signal per Watts of incident energy. The output may vary according to the type of detector, for example, while the output signals in photovoltaic detectors are usually photocurrent (i.e. Amperes), the output signals in photoconductor detectors are obtained as voltage. Photosensitivity is related to the magnitude of the sensor's response and is expressed as follows;

\begin{table}
\footnotesize
\caption{Types of Infrared Detectors}
\label{Table2}
\begin{tabular}{p{34pt}|p{70pt}|p{76pt}} \toprule
\makecell{}&\makecell{Type} & \makecell{Properties}\\
    \midrule
    \makecell{Thermal \\ Detectors} 
    & \makecell{Thermocouples, \\ Thermopiles,\\ Bolometers,\\ Pneumatic cells,\\ Pyroelectric \\Detectors}  
    & \makecell{room temperature \\operation, low cost,\\ low sensitivity, \\slow response.}\\ \hline
    \makecell{Photon \\Detectors} & \makecell{Photoconductors, \\Photodiodes,\\ Schottky Barrier\\ Detectors,\\ Quantum Well \\IR Photodetectors} & \makecell{low temperature\\ operation, higher \\cost, higher \\sensitivity, \\fast response.}\\ \bottomrule
\end{tabular}
\end{table}

\begin{equation}
R=\frac{S}{PA}
\label{responsivity}
\end{equation}

where S is signal output, P is incident energy and A is the detector's active area \cite{vollmer2017infrared,HAMAMATSUPHOTONICSK.K.2011}.

The signal-to-noise ratio (SNR) for a given input flux level is an important parameter used to determine IR image sensitivity \cite{boreman1998basic}. NEP is the quantity of incident light when the SNR is 1 and expressed as follows:

\begin{equation}
NEP=\frac{PA}{S/N\cdot\sqrt{\Delta}}
\end{equation}

where N is the noise output and $\Delta$ is the noise bandwidth (and S,P and A are the same as in {eq.} \ref{responsivity}).

Detectivity D* (\emph{normalised} detectivity) is the photosensitivity per unit active area of a detector and is expressed as follows:

\begin{equation}
D^*=\frac{\sqrt{A}}{NEP} 
\label{detec}
\end{equation}

Technologically, IR detectors are classified into two main groups: thermal detectors and photon (quantum) detectors (see Table \ref{Table2}) \cite{rogalski2002infrared}. Thermal detectors include thermocouples, thermopiles, pyrometers and bolometers that use infrared energy for detection. They are constructed using metal compounds or semiconductor materials and are low-cost. These detectors operate at room temperature. Their sensitivity is independent of wavelengths. Consequently, they are capable of capturing scenes in all IR sub-bands. However, they suffer from slow response times, low sensitivity, and low resolution.

In contrast to thermal detectors, photon detectors simply count photons of IR radiation.  There are different technologies that operationalize these types of sensors such as photoconductors, photodiodes, Schottky Barrier Detectors, and Quantum Well detectors \cite{rogalski2002infrared}. Compared to thermal sensors, they are more sensitive and operate faster. However, these types of detectors do not operate at room temperature but require a cooling capability. In addition,  they are made from materials such as InSb, HgCdTe, and GaAs/AIGaAs whose sensitivity depends on photon absorption and, therefore are more expensive. They also have a limited IR spectrum. Photon detectors are usually utilised when a high-sensitivity response is required at a specific wavelength.

Comparative studies on thermal and photon detectors show that both sensor types have their pros and cons \cite{hudson1976infrared,vollmer2017infrared,rogalski2002infrared}. Photon detectors are favoured at specific wavelengths and lower operating temperatures, whereas thermal detectors are favoured at a very long spectral range \cite{Rogalski1997}. Photon detectors are fundamentally limited by generation-recombination noise arising from photon exchange with a radiating background. Thermal detectors are fundamentally limited by temperature fluctuation noise arising from radiant power exchange with a radiating background \cite{KRUSE1995869}.

\begin{figure}[t]
\centering
\begin{subfigure}{0.40\columnwidth}	
	\includegraphics*[trim=0 0 0 0,clip=false,width=1\textwidth]{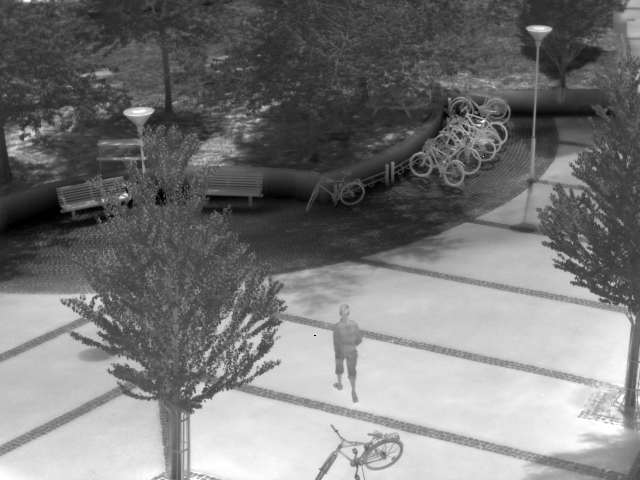}
	\caption{}	
\end{subfigure}
\begin{subfigure}{0.40\columnwidth}	
	\includegraphics*[trim=0 0 0 0,clip=true,width=1\textwidth]{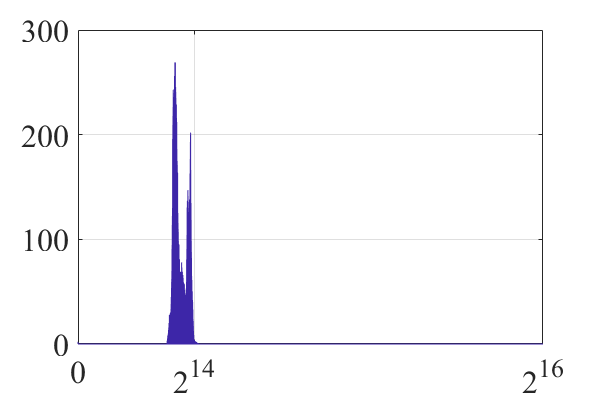}
	\caption{}	
\end{subfigure}
\begin{subfigure}{0.43\columnwidth}	
	\includegraphics*[trim=-0 0 0 0,clip=false,width=1\textwidth]{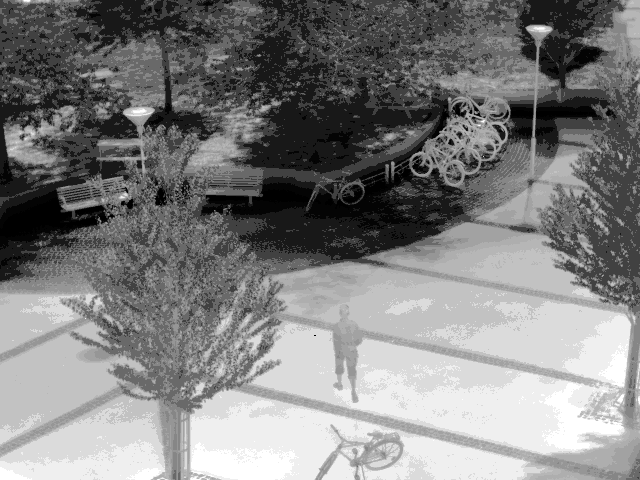}
	\caption{}	
\end{subfigure}
\begin{subfigure}{0.40\columnwidth}	
	\includegraphics*[trim=0 0 0 0,clip=false,width=1\textwidth]{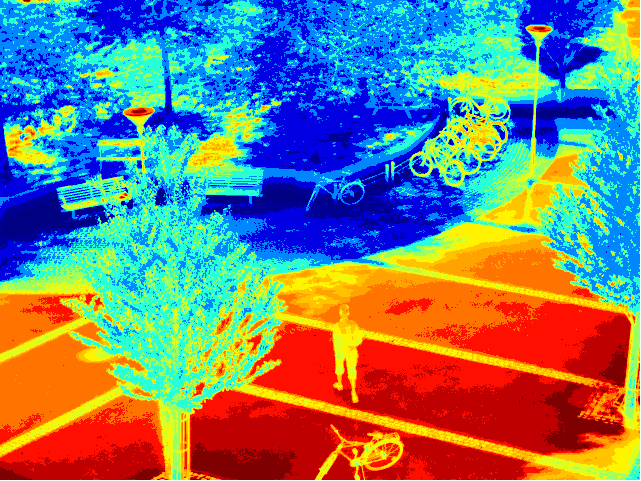}
	\caption{}	
\end{subfigure}
\caption{(a) A 16-bit raw IR image, (b) its 16-bit raw pixel histogram (x-axis has logarithmic scale), (c) the enhanced image and (d) the false-colour image are depicted. (The picture is taken from The LTIR Dataset \cite{ltir2015}) }
\label{rawimage}
\end{figure}

\begin{figure*}[t]\centering
\includegraphics[trim=100 0 70 0,clip=true,width=1\textwidth]{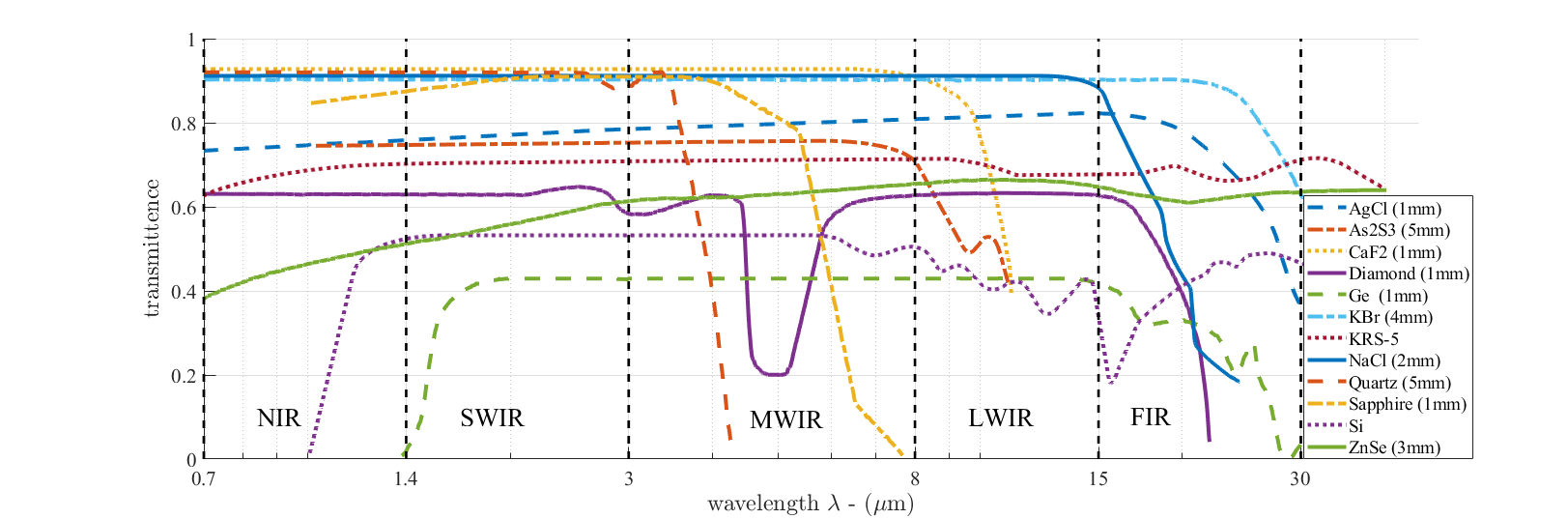}
\caption[]{Transmittance of different materials in IR sub-bands.}
\label{transmit}
\end{figure*}

\subsubsection{IR Detector Raw Output}
The raw pixel output of an IR detector is the irradiance (i.e. the flux of infrared energy per unit area) transformed into quantised n-bit values. These values are within the limits of the so-called ``dynamic range'', which is the difference between the largest and smallest signal value the detector can record or reproduce. Hence, the raw pixel values are usually not uniformly distributed within the dynamic range. In practice, a raw IR detector output is usually confined to a very limited range. In Figure \ref{rawimage}, a 16bit IR detector raw output (taken from \cite{ltir2015}), its 16-bit raw pixel histogram and the enhanced image are depicted.

IR electro-optical systems that provide a visual output for human users, enhance the raw detector output using contrast-enhancing histogram shaping methods \cite{Soundrapandiyan2022}. These types of systems usually provide 8-bit contrast-enhanced images as output. {The aim of such a process is to increase the contrast of the raw IR image for the human observer. As seen in Figure} \ref{rawimage}a, {the raw image is barely visible to the human eye. Due to the irreversibility of most image enhancement algorithms, the bit range decreases with the price of sacrificing information. This enhancement is usually a default process for visible band cameras. On the other hand, systems that provide intelligent IR image processing algorithms, such as tracking, detection, recognition, etc., utilize the raw output of pixels; since the raw output is representative of the actual irradiance values collected from the scene and has a higher dynamic bit range.} The raw output of the electro-optical system usually has the same bit-depth as the IR detector, such as 11-bits or 14-bits. In the following section, when analyzing the various image and video sets, information regarding the raw or enhanced nature of pixel values for a given set is specifically indicated. 

Some thermal cameras utilize false colours for their 8-bit contrast-enhanced output. This is usually done for temperature mapping for cameras that are used for temperature measurement. In Figure \ref{rawimage}d,  an example of a false-colour contrast-enhanced infrared image is depicted.  

\subsection{Infrared Optics}
IR imaging technology was founded in the late 1920s with the understanding of photon emission, and improvements continue even today \cite{buser1997historical}. IR imaging is based on a fundamental concept in geometrical optics called the ray model. A ray model ignores the diffraction and assumes that light travels in straight lines from a source point. Each location in the scene can be assumed as a source point, and the source points emit different levels of radiation that create the IR scene.\cite{boreman1998basic}. 

In geometrical optics an image is constructed via an optical material, by focusing the rays collected from the scene onto an image plane. Hence, the optical material used in an infrared system needs to be transparent (i.e. with transmittance closer to 1.0) at the wavelength the detector is sensitive to. The percentage of incident light that passes through a material for a given wavelength of radiation is defined as electromagnetic transmission, also known as transmittance.

When choosing the correct optical material for an IR imaging system, there are three main points to consider. The first is the thermal properties of the material. Optical materials are typically placed in environments with varying temperatures, and as a result, they can generate a significant amount of heat. To ensure that the user receives the desired performance, the coefficient of thermal expansion (CTE) of the material should be evaluated. Secondly, as mentioned above, sufficient transmittance of the material for the given wavelength is a must. In Figure \ref{transmit}, the transmittance of different materials in IR sub-bands is depicted. For example, if the system is intended to operate in the LWIR band, germanium (Ge) optics with a thickness of 1mm are preferable to sapphire optics with the same thickness.

Another factor in choosing a suitable optical material is the refractive index, which is the measure of how fast radiation travels through a material. IR refractive index varies among materials, allowing more flexibility in system design. As a solution, anti-reflection coatings are applied to materials used for IR optics, which also limits them to a desired band within the IR spectrum.

For more information on the subject, the reader may refer to \cite{fundIRoptic91}.

\begin{table*}
\centering
\footnotesize
\caption{A selection of commercial NIR electro-optical systems and their properties.}
\label{tab:table4}
\begin{tabular}{p{46pt}|p{62pt}|p{45pt}|p{40pt}|p{37pt}|p{68pt}|p{63pt}}
\toprule
    \makecell{Cameras $\longrightarrow$\\ Parameters $\downarrow$}     & \makecell {FLIR\\A35}  & \makecell {FLIR\\T560}  & \makecell {FLIR\\A655SC}   & \makecell {FLIR\\T1010}   & \makecell {FLIR\\A65}  & \makecell {FLIR\\E8-XT}  \\
\midrule
\makecell{\emph{Spectral}\\\emph{Range}}  & \makecell {7.5–13 $\mu m$ }&\makecell{7.5-14$\mu m$}   &\makecell{7.5–14$\mu m$}   &\makecell {7.5–14$\mu m$}  &\makecell {7.5–13$\mu m$ } &\makecell {7.5–13 $\mu m$ }     \\
    \hline
   \makecell{\emph{Detector}\\\emph{Type}}  & 
\makecell {Uncooled  \\VOx µ-bol.} & 
\makecell {Uncooled \\µ-bol.} & 
\makecell {Uncooled \\µ-bol. }& 
\makecell {Uncooled \\µ-bol. }&  
\makecell {Uncooled VOx \\µ-bol.}  &  
\makecell {Uncooled\\ µ-bol. }  \\
   \hline
    \makecell{\emph{Field of }\\\emph{View}\\(\emph{lens size}\\(\emph{if available}} & \makecell{63°x50°(7.5mm)\\48°x39°(9mm)\\24°x19.2°(19mm)\\13°x10.8°(35mm)\\7.6°x6.08°(60mm)} &
    \makecell{14°x10° } & 
    \makecell{15°x11°}&  
    \makecell {12°x9°} & \    \makecell {90°x69°(7.5mm)\\ 45°x37°(13mm) \\25°x20°(25mm) \\12.4°x9.92°(50mm) \\6.2°x4.96°(100mm)}  &  
    \makecell {45°x34°} \\
    \hline
    \makecell{\emph{Thermal}\\\emph{Sensitivity}\\\emph{(NETD)}}    & 
    \makecell {<0.05°C @ 30°C \\ / 50 mK }&  
    \makecell {<50 mK\\ @ 30°C}  &   
    \makecell {<30 mK } &    
    \makecell {<25 mK\\ @ 30°C }  &   
    \makecell {< 50 mK\\@ 30°C} &
    \makecell{<0.05°C\\/<50 mK}
 \\
    \hline
    \makecell{\emph{Object}\\\emph{Temperature}\\\emph{Range}}    & 
    \makecell{–25$\leftrightarrow$135°C \\ –40$\leftrightarrow$550°C }  &
    \makecell{–20$\leftrightarrow$120°C\\0$\leftrightarrow$650°C\\300$\leftrightarrow$1500°C}&  
    \makecell{–40$\leftrightarrow$150°C \\ 100$\leftrightarrow$650°C }  &  
    \makecell{–40$\leftrightarrow$650°C } &  
    \makecell{–25$\leftrightarrow$135°C \\ –40$\leftrightarrow$550°C } &
    \makecell{–20$\leftrightarrow$550°C \\(-4$\leftrightarrow$1022°F)}\\
    \hline
    \makecell{\emph{Measure-}\\\emph{ment}\\\emph{Accuracy}}& \makecell{± 5°C (± 9°F) \\or 5 \% \\of reading}   & \makecell{±2°C \\ or ±2\% \\of reading} & \makecell{±2°C \\or ±2 \% \\of reading} & \makecell{± 2 °C \\or ± 2\% \\of reading}  & \makecell{± 5°C (± 9°F)\\or 5 \% \\of reading} & \makecell{±2°C (±3.6°F) \\or ±2\% \\of reading}\\
    \hline
    \makecell{\emph{Operating}\\\emph{Temperature}\\\emph{Range}}& 
    \makecell {–15$\leftrightarrow$60°C } &  
    \makecell {–40$\leftrightarrow$75°C } & 
    \makecell {–15$\leftrightarrow$50°C }  &  
    \makecell {–15$\leftrightarrow$50°C } &  
    \makecell {–15$\leftrightarrow$60°C}  & 
    \makecell {–15$\leftrightarrow$50°C } \\
\bottomrule
\end{tabular}
\end{table*}

\subsection{IR Electro-Optical System Properties }

There are some important parameters used in selecting appropriate equipment and characterising the performance of IR systems. The parameters that measure the performance of an IR electro-optical system depend on its ability to detect IR radiation and resolve the temperature differences in the scene. The contrast in an IR image occurs due to variations in temperature and emissivity. The parameters that may affect the performance of an IR electro-optical system, in general, include spectral range, normalised detectivity, temperature range, absolute accuracy, repeatability, frame rate, spatial resolution and thermal sensitivity \cite{venkataraman2003performance}. Below these parameters are briefly explained: 

\begin{itemize}[leftmargin=*]
    \item \emph{Spectral range}: refers to the wavelength range in which the IR system will operate. 
    \item \emph{Normalised detectivity} (D*): as defined in {eq.} (\ref{detec}), is one of the widely used parameters to compare the performance of IR detectors.
    \item \emph{Temperature range}: or the \emph{operating temperature}, is the minimum and maximum temperatures that can be measured by the IR electro-optical system. It has a unit of K, C$\degree$, or F$\degree$.
    \item \emph{Absolute Accuracy}: is a measure of how accurately the system detects the actual temperature and is denoted by temperature units. Related to this measure, \emph{Repeatability} is defined as the consistency of the system accuracy.
    \item \emph{Frame rate}: is the number of frames displayed per second. For monitoring moving objects, higher frame rate cameras are mostly preferred \cite{bagavathiappan2013infrared}. It has a unit of Hz. 
    \item \emph{Spatial resolution}: also referred to as the ``instantaneous field-of-view'' (IFOV), is the imaging system's ability to differentiate the details of objects within a single pixel-sized FOV. It is a measure of solid angle, hence represented by steradians. As spatial resolution increases, so does the image qualityç \cite{bagavathiappan2013infrared}.
    \item \emph{Thermal sensitivity}: is the smallest temperature change detected by the IR imaging system. There are three most common parameters used as a measure of thermal sensitivity, namely ``Noise Equivalent Temperature Difference'' (NEDT), ``Minimum Resolvable Temperature Difference'' (MRDT) and ``Minimum Detectable Temperature Difference'' (MDTD) \cite{venkataraman2003performance}. It has a unit of temperature (i.e. K, C$\degree$, or F$\degree$).
\end{itemize}

In order to choose the right camera for the right application, all of the aforementioned parameters should be taken into account. There are numerous commercial IR electro-optical systems available in the market. In Table \ref{tab:table4}, we provide a selection of six different near-infrared electro-optical systems, with their comparative parameters, so as to give the reader a sense of the systems engineering perspective of IR electro-optical system selection.

\subsection{Applications of IR Electro-Optical Systems }
The development in IR sensing technologies has resulted in countless applications, which we divided into four major categories: military $\&$ surveillance, industrial, medical, and scientific. Each category title is briefly explained below. The IR image and video sets provided in the next section are categorised according to these application titles. 

\begin{itemize}[leftmargin=*]
\item \emph{Military $\&$ Surveillance Applications}: The military and surveillance field, which also encapsulates law enforcement and rescue applications, cover a wide variety of applications utilised in all IR sub-bands. Warfare applications include target tracking/detection/acquisition in various platforms such as missile seeker heads, forward-looking infrared (FLIR) systems, infrared search and track (IRST) systems, and directional countermeasure (DIRCM) systems. Regarding law enforcement and rescue applications, night vision systems, reconnaissance and surveillance, fire fighting and rescue in smoke, identification of earthquake victims' locations, forest fire detection, and radiation thermometer are prime examples.
\item \emph{Industrial Applications}: Industrial applications of IR imaging systems include the utilization of IR sensing technology in various industrial fields, such as infrared heating in process control, nondestructive inspection of thermal insulators, hidden piping location detection, diseased tree and crop detection, hot spot detection, brake lining, industrial temperature measurement, clear-air turbulence detection, pipeline leak and petrol spill detection, just to name a few. 

\item \emph{Medical Applications}: In medicine, IR technology is fundamentally used for diagnosis, such as early cancer detection, determining the optimum site for amputation, determining the location of the placenta, detecting strokes and vein blockages before they occur, monitoring wound healing, and detecting infection. Due to its non-invasive nature, IR technology in medicine provides information about conditions that are directly or indirectly related to the focused region of the body (such as hands \cite{SOUSA2017315}), as well as facilitating the assessment of treatment. 

\begin{figure*}[t]
\centering
\begin{subfigure}{5.1cm}	
	\includegraphics*[trim=0 70 0 20,clip=true,width=1\textwidth]{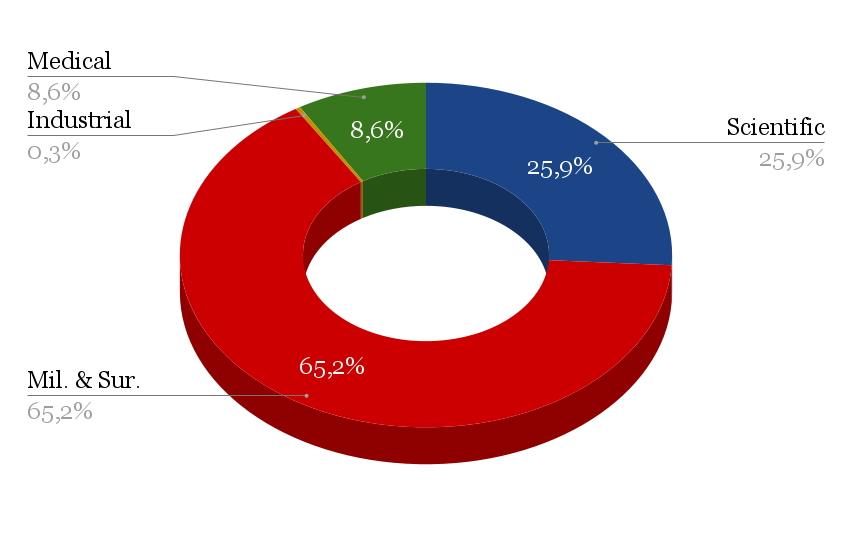}
	\caption{Application fields}
\end{subfigure}
\begin{subfigure}{5.1cm}	
	\includegraphics*[trim=0 70 0 -20,clip=true,width=1\textwidth]{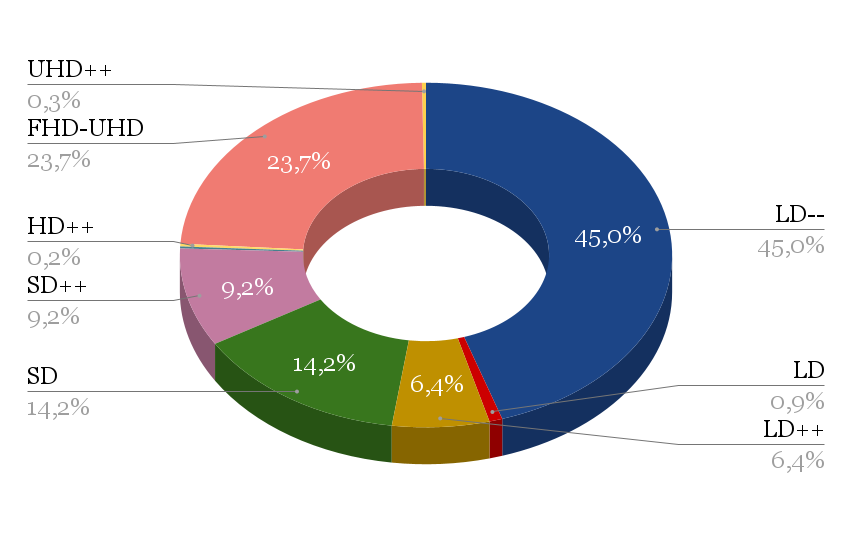}
	\caption{Image/frame resolutions}
\end{subfigure}
\begin{subfigure}{5.1cm}	
	\includegraphics*[trim=0 70 0 0,clip=true,width=1\textwidth]{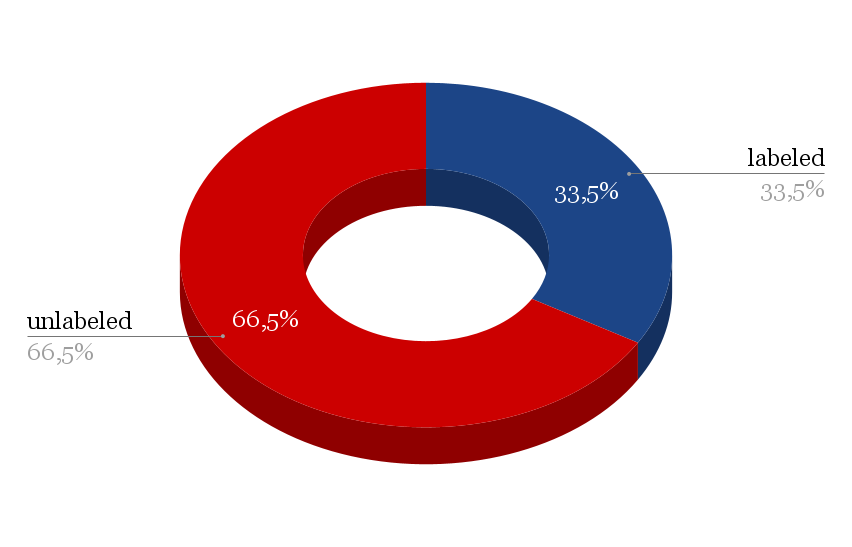}
	\caption{Ground truth labels}
\end{subfigure}
\begin{subfigure}{5.1cm}	
	\includegraphics*[trim=0 70 0 0,clip=false,width=1\textwidth]{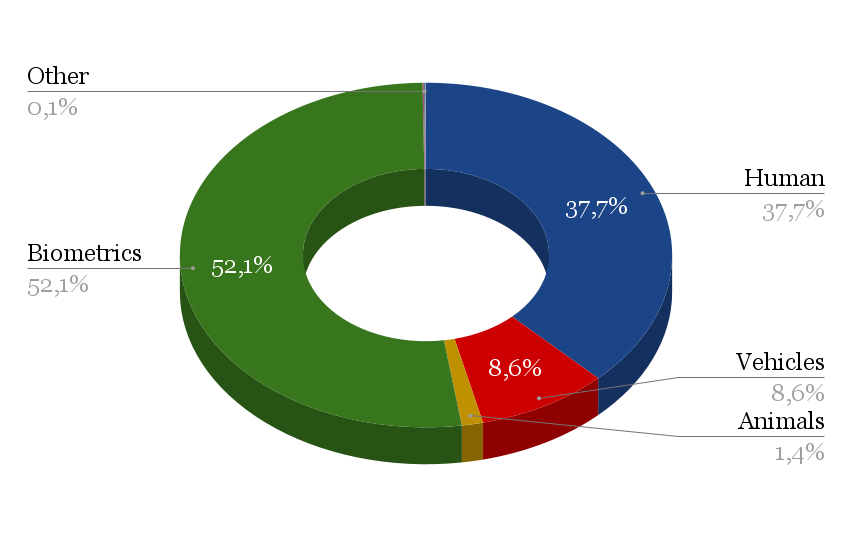}
	\caption{Object categories}	
\end{subfigure}
\begin{subfigure}{5.1cm}	
	\includegraphics*[trim=0 70 0 0,clip=true,width=1\textwidth]{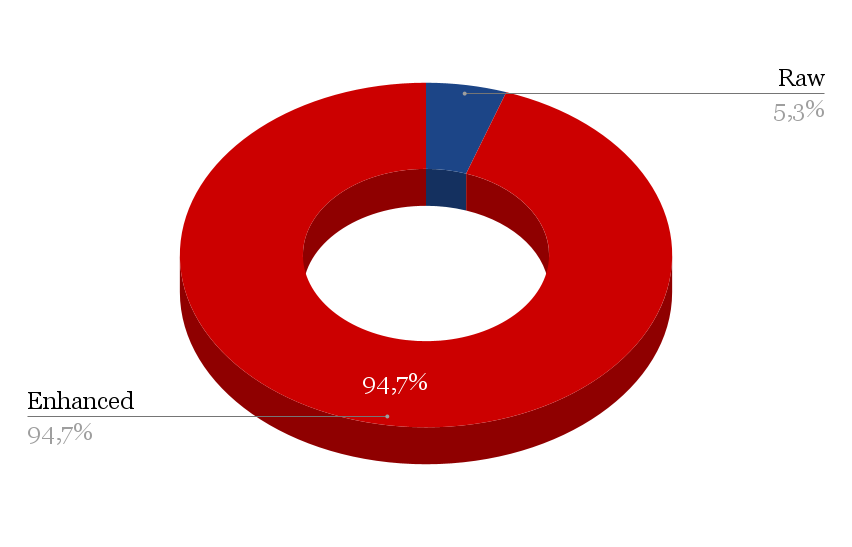}
	\caption{Image Enhancement}
\end{subfigure}
\caption{Distributions of various attributes (application field, resolution, ground truth labelling, object category and image enhancement) in the entire collected image/frame corpus are depicted in pie charts. }
\label{piecharts}
\end{figure*}

\item \emph{Scientific Applications}: In nearly every scientific field, from remote sensing and meteorology to material science and microbiology, from engineering to biology, IR imaging technologies are used. {In this paper, when categorizing a set as "scientific", we took into account its use outside of the other categorised sectors, namely military, surveillance, industrial, and medical. An image or video set, for example, is classified as both Military $\&$ Surveillance and Scientific if it has the capacity to support both types of applications.} 
\end{itemize}

\section{IR Image \& Video Sets}
\label{sec:others}
The paper analyses {109} IR image and/or video sets and provides a list of the sets in Table \ref{datasetList}. {A total of 77 are public, in other words, they offer public download links, while 3 are private and require payment. The remaining 29 sets can be downloaded for free but require manual registration by contacting the institution that owns them.} The entire corpus of sets includes nearly 20 million still images and video frames. In the following, we provide the statistical details of the compiled list of IR image and video sets in terms of application fields, included object categories, resolution, annotation, and preprocessing, before presenting the list with brief descriptions.

Table \ref{datasetList} provides a sample image and a brief description for every set. There are also separate columns for technical details, such as types of annotated classes, number of total frames, image resolutions, sensor types, image bit depths, and application fields. {The description section additionally specifies whether the collection is accessible to everyone (pub), accessible to paid users only (pri), or needs registration (rr). For more details, we suggest that the reader consult the References section for an online link to the image set.}

\subsection{Application Fields} As mentioned in the previous section, application fields for IR image and video sets are scrutinised in 4 main titles: Military \& Surveillance (Mil. \& Sur.), Industrial, Medical and Scientific.  As seen in Figure \ref{piecharts}a, Military \&  Surveillance comprises 65.2\% of the total volume of images and video frames, clearly demonstrating the importance of IR imaging in this industry. Sets collected for scientific applications cover 25.9\% of the corpus, while medical applications cover 8.6\%, most likely due to the legal challenges involved in collecting or publishing health informatics data. 
Industrial applications account for a marginal share, which is probably due to the fact that they do not publish their data in public domain. In Table \ref{datasetList},  the application fields for every individual set are indicated in the right-most column (titled ``\emph{App.}'').

\subsection{Resolution and Sensor}
IR image and video sets listed in this survey range from \emph{lower}-definition (LD-{}-), which corresponds to resolutions lower than LD, to ultra-\emph{higher}-definition (HD++), which corresponds to resolutions higher than UHD. Depending on the application, the resolution plays a significant role. Most surveillance systems require HD or better resolutions for accuracy. On the other hand, LD and standard definition (SD) systems are ideal when the computational capabilities of the system are limited. Figure \ref{piecharts}b shows that despite three-quarters of the corpora being SD++ or worse, the rest are almost UHD or better. It is important to note that sets with UHD or better resolutions are recent sets showing a clear future trend. In Table \ref{datasetList}, the actual resolutions for every individual set are indicated in column four (titled ``\emph{res}''). 

{In addition, the optical equipment used to collect each set is provided in column five (titled ``\emph{Sensor}''). When compared to RGB cameras, IR optical systems are capable of different kinds of calibration, and they are capable of producing characteristic output that may not be replicated with similar equipment. The reason for this is that today's RGB cameras usually use the same preprocessing and aim at producing almost the same output, whereas, with IR vision, it becomes important to know the parameters of the equipment in order to recreate similar scenes or images. Therefore, in Table} \ref{datasetList} {column titled ``\emph{Sensor}'', we provide details regarding the collection equipment for the sets, which openly specify these details.}

\subsection{Annotations and Object Categories}
Many computer vision applications annotate data with labels for certain purposes, such as detection, tracking, and recognition. Data annotation/labelling is an expensive effort, which provides means for supervised learning, and hence deep learning if the annotated data are sufficiently large in scale. Similarly, some of the IR sets listed in this survey are annotated with various labels. As shown in Figure \ref{piecharts}c, about 33.5\% of the entire corpus is annotated. For some sets, these annotations are black-box locations for objects, whereas for others they are global labels for entire images. {A majority of the corpora are not labelled, but we believe that most annotations may not be shared publicly due to their commercial implications. Once again, it is important to note that sets with labels are recent sets showing another future trend.} 

In Table \ref{datasetList}, (in column three, titled ``\emph{Classes}''), categories for any existing annotation of a given set are provided. The entire collection of sets includes a wide range of object annotation categories. The objects are categorised under {seven} titles in Figure \ref{piecharts}d, namely biometrics, environments, humans, vehicles, animals, unknown and uncategorised. Biometrics annotations include IR images of faces, irises, ears and/or fingerprints, and cover the majority of the annotations with a 52.1\% share. Human annotations, including pedestrians, runners, sportsmen, etc cover 37.7\% of these annotations. Vehicles of different sorts such as cars, bicycles, motorcycles, aircraft, boats, etc, are also included and cover 8.6\% of label annotations.  There are a small number of animal class annotations that take 1.4\%. There is also a marginal share of annotations that are related to environmental objects, including terrain, roads, clouds, or various objects like food, or uncategorised application-specific labels. 

\subsection{Image Enhancement}
As mentioned previously in Section 2.2.1, IR electro-optical systems that provide a visual output for human users, usually enhance the raw detector output using contrast-enhancing histogram shaping methods. {However, IR image processing systems that utilize algorithms such as tracking, detection, recognition, etc., utilize the raw output of pixels, which usually has the same bit-depth of the IR detector. The histogram-enhanced image is, in most cases, the only accessible output of an IR optical system. For such systems, the details of the enhancement algorithms are rarely provided to the user. Most systems apply different  algorithms that suit their design requirements such as level of contrast, and real-time operation, just to name a few.} 
As seen in Figure \ref{piecharts}e, only a minority of 5.3\% of the entire corpora of collected frames are raw detector outputs. The ``bit'' column in Table \ref{datasetList} gives information about the bit depth of an image/frame for a given set. The number (8, 11, 16, etc.) corresponds to image bit-depth. For some sets, the bit depth is indicated by ``8*'' showing that the images/frames are of 24bit RGB (i.e 8bit per channel) format.  The abbreviation ``HE'' is to indicate the existence of any histogram enhancement process, whereas ``RAW'' suggests the accessibility of the raw detector output. {The type of enhancement technique is not indicated in the table, because this information is not available for most of the collection equipment.}

\begin{center}

\begin{table*}[p]
  \centering
  \footnotesize
  \caption{From left to right, the columns depict the name and the reference, a sample image, the included object classes (if any), total number of frames, pixel resolution of images, the optical system (if specified), pixel bit depth (and if any histogram equalization - HE applied), a brief description and the application fields of the given dataset, respectively.}
    \begin{tabular}{p{30pt}|p{43pt}|p{38pt}|p{12pt}|p{15pt}|p{26pt}|p{11pt}|p{150pt}|p{15pt}}
    \toprule
    \makecell{Name} & 
    \makecell{Sample} &
    \makecell{Classes} & 
    \makecell{\#\\\emph{fr.}} & 
    res. & 
    \makecell{\emph{Sensor}} & 
    \makecell{\emph{bit}} & 
    \makecell{\emph{Description}} & 
    App. \\
    \midrule    

    \makecell{\rotatebox[]{90}{AAlart Data} \rotatebox[]{90}{\cite{zhang2018novel}}} &
    $\vcenter{\includegraphics[width=1.0\linewidth]{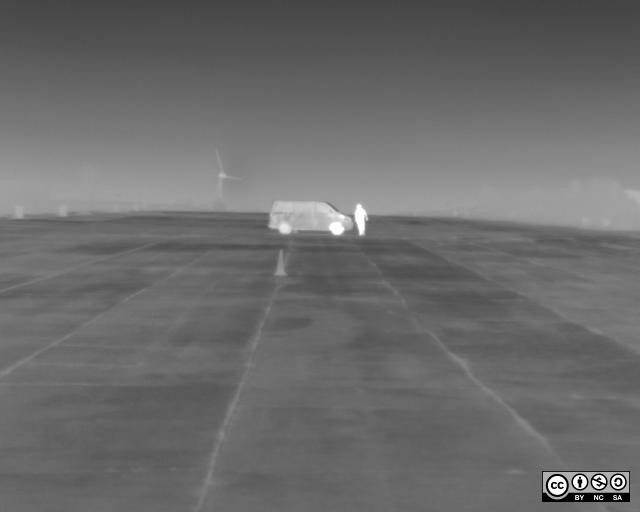}}$ &
    {pedestrian, vehicle} & 
    \makecell{\rotatebox[]{90}{771}} & 
    \makecell{\rotatebox{90}{640x513}} & 
    \makecell{\rotatebox[]{90}{\begin{tabular}[c]{@{}c@{}}Catherine \\ MP LWIR\end{tabular}}} & 
    \makecell{\rotatebox[]{90}{8 HE}} & 
    $\vcenter{{(\emph{pub})} The dataset includes thermal infrared video sequences comprising of pedestrians and various types of vehicles for an infrared surveillance system that can be widely applied in military applications.}$ &
    \makecell{\rotatebox[]{90}{Mil. \& Sur.}} \\
    \midrule 
    
    \makecell{\rotatebox[]{90}{AAU RainSnow} \rotatebox[]{90}{\cite{bahnsen2018rain}}}  & 
    $\vcenter{ \includegraphics[width=1.0\linewidth]{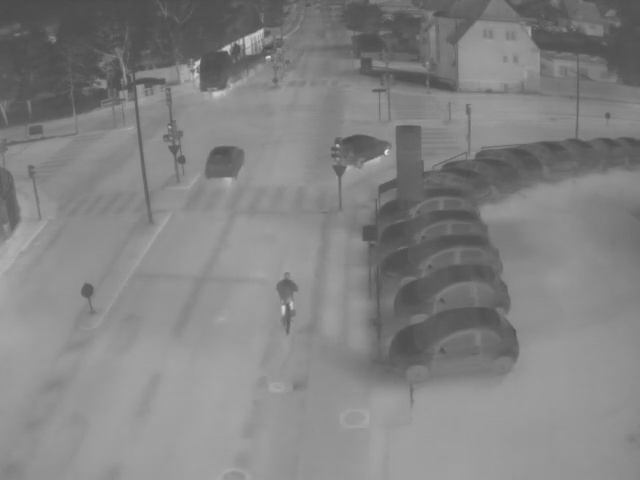}}$  & 
    {vehicle}  & 
    \makecell{\rotatebox[]{90}{4.5K} } & 
    \makecell{\rotatebox[]{90}{640x480} } & 
    \makecell{\rotatebox[]{90}{not specified} } & 
    \makecell{\rotatebox[]{90}{8* HE} } & 
    $\vcenter{{(\emph{pub})} The image set includes traffic surveillance  images (RGB+thermal) and 22 video pairs in rainfall and snowfall from seven different traffic intersections at the Danish cities of Aalborg and Viborg. The purpose of the dataset is detection and classification under challenging weather conditions. The dataset includes pixel-level annotations of 2.200 frames, containing 13,297 objects.}$  & 
    \makecell{\rotatebox[]{90}{Mil. \& Sur.}  } \\
    \midrule
    
    \makecell{\rotatebox[]{90}{AAU-PD-T} \rotatebox[]{90}{\cite{DiverseData}}} &
    $\vcenter{\includegraphics[width=1.0\linewidth]{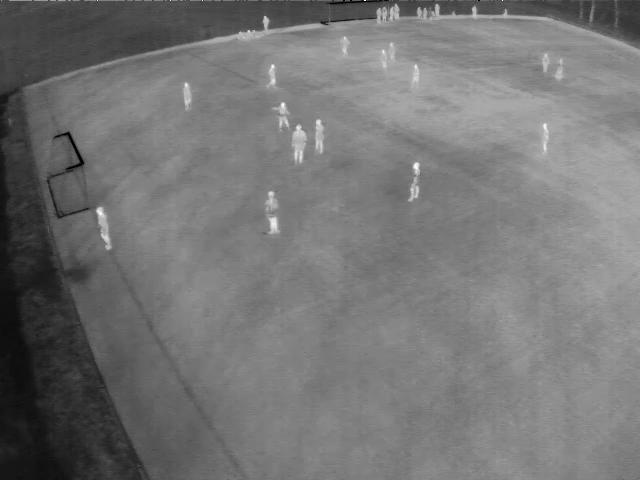}}$ &
    {pedestrian} & 
    \makecell{\rotatebox[]{90}{3K}} & 
    {\rotatebox[]{90}{\begin{tabular}[c]{@{}c@{}}384×288\\640×480\end{tabular}}} & 
    \makecell{\rotatebox[]{90}{\begin{tabular}[c]{@{}c@{}}Axis Q1921\\Axis 1922\end{tabular}}} & 
    \makecell{\rotatebox[]{90}{8* HE}} & 
    $\vcenter{{(\emph{pub})} The image set contains approximately 3k thermal pedestrian images with a total of 5590 person annotations. Training data is divided into 9 categories, such as good weather, far viewpoint, occlusions, snow, wind, just to name a few.}$ &
    \makecell{\rotatebox[]{90}{Mil. \& Sur.}} \\
    \midrule 

    \makecell{\rotatebox[]{90}{{All-Ther}} \rotatebox[]{90}{\cite{ALL_Ther}}} &
    $\vcenter{\includegraphics[width=1.0\linewidth]{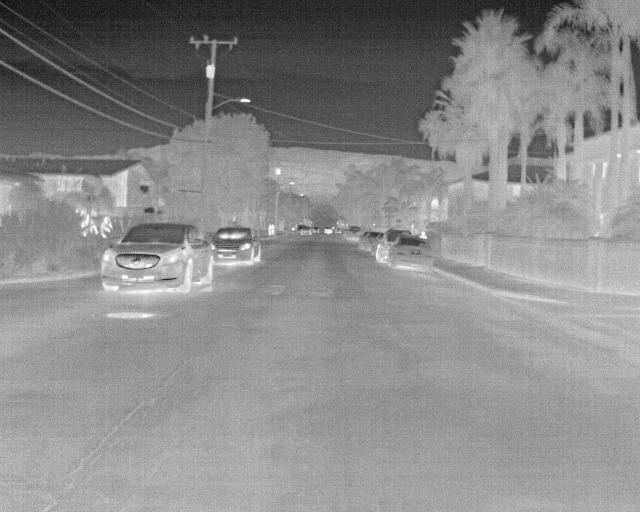}}$ &
    {vehicle, pedestrian} & 
    \makecell{\rotatebox[]{90}{20K}} & 
    {\rotatebox[]{90}{\begin{tabular}[c]{@{}c@{}}640x512\\1280x1024\end{tabular}}} & 
    \makecell{\rotatebox[]{90}{\begin{tabular}[c]{@{}c@{}}not specified\end{tabular}}} & 
    \makecell{\rotatebox[]{90}{8* HE}} & 
    $\vcenter{{(\emph{pub})} The thermal dataset contains vehicle and pedestrian infrared images in traffic and parking land scenarios. The set provides bounding box annotations.}$ &
    \makecell{\rotatebox[]{90}{Mil. \& Sur. }} \\
    \midrule 
    
    \makecell{\rotatebox[]{90}{ASL-TIR} \rotatebox[]{90}{\cite{6907094}}}  & 
      $\vcenter{ \includegraphics[width=1.0\linewidth]{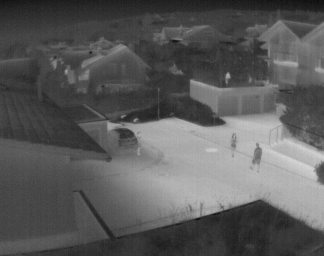}}$  & 
      {human, cat, horse}  & 
      \makecell{\rotatebox[]{90}{4381}  } & 
      \makecell{\rotatebox[]{90}{324x256} } & 
      \makecell{\rotatebox[]{90}{\begin{tabular}[c]{@{}c@{}}FLIR Tau\\320\end{tabular}}} &
      \rotatebox[]{90}{\makecell{8/16\\HE/RAW}} &
      $\vcenter{{(\emph{pub})} ASL thermal IR dataset includes 8 sequences of thermal images of humans, cats and horses, with annotations.  The images are captured at indoor and  outdoor environments.}$  & 
     \makecell{\rotatebox[]{90}{Mil. \& Sur.}  } \\
     \midrule

    \makecell{\rotatebox[]{90}{{Baracca}} \rotatebox[]{90}{\cite{baracca}}}  & 
     $\vcenter{ \includegraphics[width=1.0\linewidth]{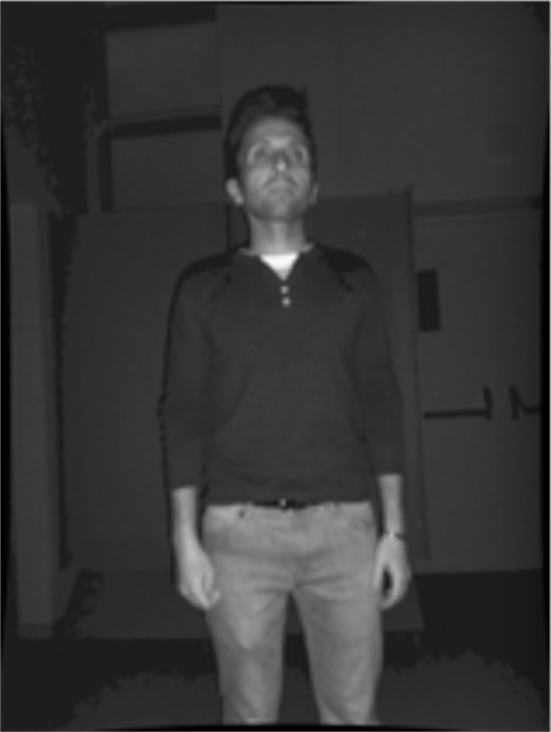}}$  & 
      {person}  & 
      \makecell{\rotatebox[]{90}{9600} } & 
      {\rotatebox[]{90}{\begin{tabular}[c]{@{}c@{}}160x120\\640×480\end{tabular}}} & 
      {\rotatebox[]{90}{\begin{tabular}[c]{c@{}c@{}}Pico Zense\\DCAM710\\PureThermal 2\end{tabular}}} &
      \makecell{\rotatebox[]{90}{8 HE} } & 
      $\vcenter{{(\emph{rr})} The dataset presents depth, thermal and RGB images for anthropometric measurements such as height, shoulder width, and forearm with annotations.}$  & 
     \makecell{\rotatebox[]{90}{Scientific} } \\
     \midrule

    \makecell{\rotatebox[]{90}{BCT} \rotatebox[]{90}{\cite{Breast_Cancer_thermography}}}  & 
     $\vcenter{ \includegraphics[width=1.0\linewidth]{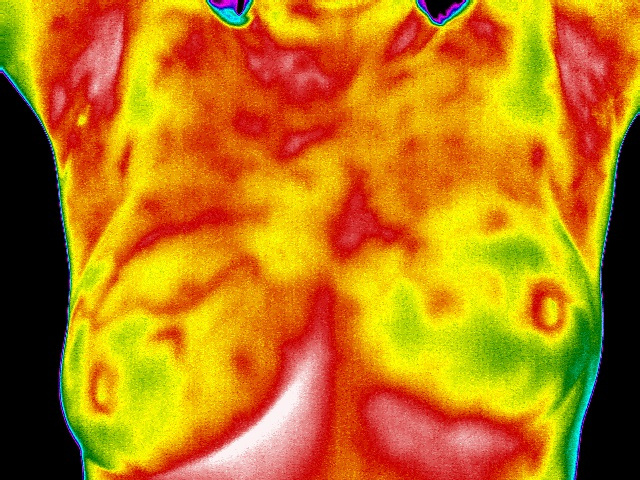}}$  & 
      {breast}  & 
      \makecell{\rotatebox[]{90}{60} } & 
      \makecell{\rotatebox[]{90}{640x480} } & 
      \makecell{\rotatebox[]{90}{Infrec  R500} } & 
      \makecell{\rotatebox[]{90}{8 HE} } & 
      $\vcenter{{(\emph{pub})} The dataset contains thermal breast images of 60 patients for breast cancer detection.}$  & 
     \makecell{\rotatebox[]{90}{Medical} } \\
     \midrule

 \makecell{\rotatebox[]{90}{BERTIN} \rotatebox[]{90}{\cite{robin}}}  & 
      $\vcenter{ \includegraphics[width=1.0\linewidth]{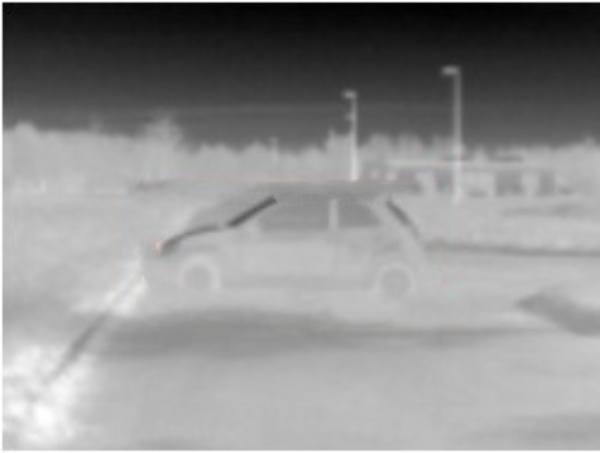}}$  & 
      {vehicle, pedestrian}  & 
      \makecell{\rotatebox[]{90}{more  than  10K} } & 
      \makecell{\rotatebox[]{90}{320x240} } & 
      \makecell{\rotatebox[]{90}{FLIR  A20M} } & 
      \makecell{\rotatebox[]{90}{16 RAW} } & 
      $\vcenter{{(\emph{rr})} The dataset includes both visible and  thermal images of pedestrians and vehicles, captured with a static and a moving camera. Contains training, validation  and test sets with their ground truths. } $ & 
     \makecell{\rotatebox[]{90}{Mil. \& Sur.} } \\
     \midrule

     \makecell{\rotatebox[]{90}{{Bird}} \rotatebox[]{90}{\cite{bird}}}  & 
      $\vcenter{ \includegraphics[width=1.0\linewidth]{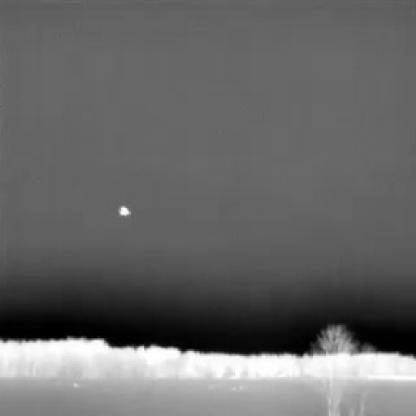}}$  & 
      {bird}  & 
      \makecell{\rotatebox[]{90}{302} } & 
      \makecell{\rotatebox[]{90}{416x416} } & 
      \makecell{\rotatebox[]{90}{not specified} } & 
      \makecell{\rotatebox[]{90}{8* HE} } & 
      $\vcenter{{(\emph{pub})} The dataset presents thermal images that include flying birds with annotations.} $ & 
     \makecell{\rotatebox[]{90}{Scientific}} \\
     \midrule

    \end{tabular}%
  \label{datasetList}%
\end{table*}%

\begin{table*}[p]
  \centering
  \footnotesize
    \begin{tabular}{p{30pt}|p{43pt}|p{38pt}|p{16pt}|p{15pt}|p{26pt}|p{17pt}|p{135pt}|p{15pt}}
    \toprule
    Table \ref{datasetList}   &  continued... & 
    \makecell{} & 
    \makecell{} & 
    & 
    \makecell{} & 
    \makecell{} & 
    \makecell{} & 
    \makecell{} \\
   \midrule  

   \makecell{\rotatebox[]{90}{BIRDSAI} \rotatebox[]{90}{\cite{bondi2020birdsai}}}  & 
      $\vcenter{ \includegraphics[width=1.0\linewidth]{dataset_images/4-birdsai.jpg}} $ & 
      {human,   animal}  & 
      \makecell{\rotatebox[]{90}{160K} } & 
      \makecell{\rotatebox[]{90}{640x480} } & 
      \makecell{\rotatebox[]{90}{\begin{tabular}[c]{@{}c@{}}FLIR  Vue Pro  640 \\ Tamarisk  640\end{tabular}}} & 
      \makecell{\rotatebox[]{90}{8 HE} } & 
      $\vcenter{{(\emph{pub})} The dataset consists of two parts: real and synthetically-created, with annotations. The real data includes 48 thermal infrared  videos sequences of  humans and animals,  captured with a thermal camera mounted on a UVA, that flies over African landscapes. The synthetic data, generated with MS AirSim simulation platform, include 124 thermal infrared video sequences.}$  & 
      \makecell{\rotatebox[]{90}{Scientific}  } \\
     \midrule

 \multirow{2}{=}[-5ex]{\makecell{\rotatebox[]{90}{BU-TIV } \rotatebox[]{90}{\cite{6909984}}}} &
    $\vcenter{ \includegraphics[width=1.0\linewidth]{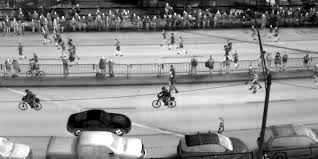}}$ & 
    \makecell[l]{motorcycle, \\ runner, \\ car, \\ bicycle, \\ pedestrian} & 
    \makecell[l]{35K \\2999\\6000\\1275\\1282} & 
    \multirow{2}{=}{\makecell{\rotatebox[]{90}{up to 1024x1024}}} & 
    \multirow{2}{=}{\makecell{\rotatebox[]{90}{FLIR SC8000}}} & 
    \multirow{2}{=}{\makecell{\rotatebox[]{90}{16 RAW}}} 
    & $\vcenter{{(\emph{pub})} The dataset contains various tasks, such as single  object detection, multi-object detection, motion detection, counting, that describes real world scenarios such as a marathon runner, people walking down  a hall, etc., with ground truth data.  In addition, the set includes images of bats for tracking and counting purposes.}$ &
    \makecell{\rotatebox[]{90}{Mil. \& Sur.}}\\
    \cline{2-4} \cline{9-9}& 
    $\vcenter{\includegraphics[width=0.8\linewidth]{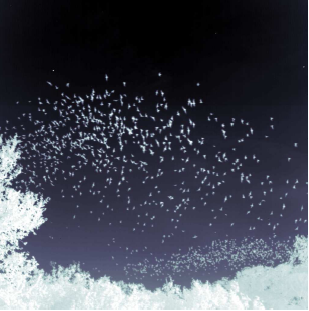}}$ & 
    {bat} & \makecell{19K} &  & \makecell &  &  &\makecell{\rotatebox[]{90}{Scientific}}\\
    \midrule

   \makecell{\rotatebox[]{90}{CAMEL} \rotatebox[]{90}{\cite{gebhardt2018camel}}}   & 
      $\vcenter{ \includegraphics[width=1.0\linewidth]{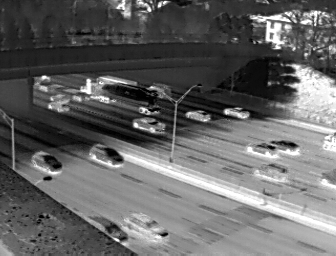}}$  & 
      {biker,   vehicle,  pedestrian}  & 
      \makecell{\rotatebox[]{90}{44.5k} } & 
      \makecell{\rotatebox[]{90}{336x256} } & 
      \makecell{\rotatebox[]{90}{FLIR  Vue Pro} } & 
      \makecell{\rotatebox[]{90}{8* HE} } & 
      $\vcenter{{(\emph{pub})} The dataset contains 26 video sequences,  captured in an urban environment,  in the visible and thermal infrared  domains, with their annotations.}$  & 
      \makecell{\rotatebox[]{90}{Mil. \& Sur.}  } \\
     \midrule
    
    \makecell{\rotatebox[]{90}{Carl  Database} \rotatebox[]{90}{\cite{CARL} }} & 
      $\vcenter{ \includegraphics[width=1.0\linewidth]{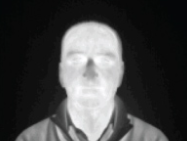}}$  & 
      {face}   & 
      \makecell{\rotatebox[]{90}{7380 } } & 
      \makecell{\rotatebox[]{90}{640x480} } & 
      \makecell{\rotatebox[]{90}{TESTO 880-3} } & 
      \makecell{\rotatebox[]{90}{8* HE}   } & 
      $\vcenter{{(\emph{rr})} Contains visible and thermal images of human faces, with a constant 135 cm distance from the sensor, in front of a matt black background.}$  & 
       \makecell{\rotatebox[]{90}{Mil. \& Sur.}  } \\
     \midrule
    
    \makecell{\rotatebox[]{90}{CATS} \rotatebox[]{90}{\cite{Treible2017CVPR}}}  & 
     $\vcenter{ \includegraphics[width=1.0\linewidth]{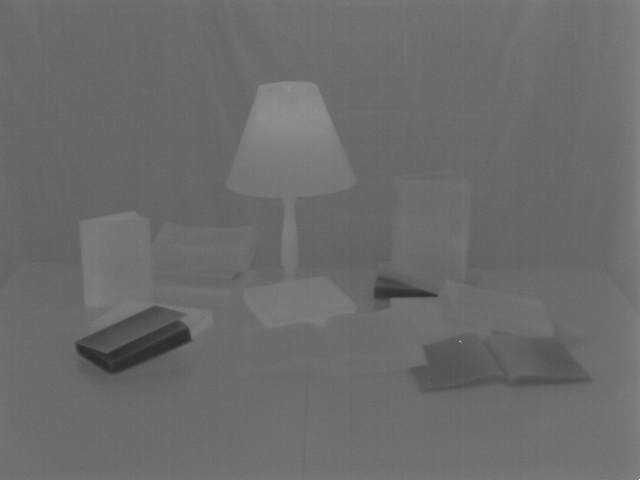}}$  & 
      {vehicles, pedestrians, other}  & 
      \makecell{\rotatebox[]{90}{1520} } & 
      \makecell{\rotatebox[]{90}{640x480} } & 
      \makecell{\rotatebox[]{90}{\begin{tabular}[c]{@{}c@{}}Xenics Gobi\\640 GigEs\end{tabular}}} & 
      \makecell{\rotatebox[]{90}{16 RAW} } & 
      $\vcenter{{(\emph{pub})} The dataset contains stereo thermal, stereo  colour, and cross-modality images of pedestrians, vehicles, and some other objects in different weather conditions.}$  & 
      \makecell{\rotatebox[]{90}{Mil. \& Sur.}} \\
     \midrule
     
     \makecell{\rotatebox[]{90}{CBSR NIR } \rotatebox[]{90}{\cite{li2007illumination} }}  & 
      $\vcenter{ \includegraphics[width=1.0\linewidth]{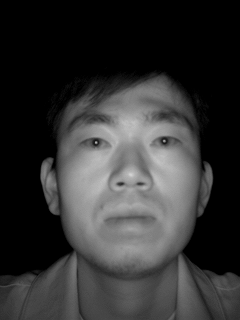}}$  & 
      {face}  & 
      \makecell{\rotatebox[]{90}{4K}    } & 
      \makecell{\rotatebox[]{90}{480x640} } & 
      \makecell{\rotatebox[]{90}{not  specified} } & 
      \makecell{\rotatebox[]{90}{8 HE}} & 
     $\vcenter{{(\emph{pub})} The dataset contains NIR images from 197 people, collected for face detection, eye detection and face recognition tasks.}$   & 
      \makecell{\rotatebox[]{90}{Mil. \& Sur.}  } \\
     \midrule
     
     \makecell{\rotatebox[]{90}{CDW 2014} \rotatebox[]{90}{\cite{wang2014cdnet}}} & 
     $\vcenter{ \includegraphics[width=1.0\linewidth]{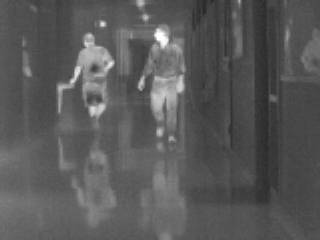}}$  & 
      {pedestrian}  & 
      \makecell{\rotatebox[]{90}{21K}} & 
      {\rotatebox[]{90}{\begin{tabular}[c]{@{}c@{}}320x240 $\rightarrow$ \\ 720x576\end{tabular}}} &
      \makecell{\rotatebox[]{90}{not specified} } & 
      \makecell{\rotatebox[]{90}{8* HE}} & 
      $\vcenter{{(\emph{pub})} The 2014 CDnet dataset, which is extended  version of CDnet 2012 dataset, contains 11 video categories, one of which  is the Thermal category, that contains 5 video sequences  (3 outdoor + 2 indoor) shot  by a far infrared camera annotated with pedestrian bounding boxes.}$  & 
      \makecell{\rotatebox[]{90}{Mil. \& Sur.}} \\
     \midrule
     
    \makecell{\rotatebox[]{90}{Cheetah} \rotatebox[]{90}{\cite{roboflow_2020}}} & 
      $\vcenter{ \includegraphics[width=1.0\linewidth]{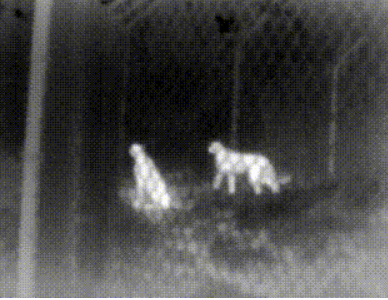}}$  & 
      {cheetah}  & 
      \makecell{\rotatebox[]{90}{129}   } & 
      \makecell{\rotatebox[]{90}{416x416} } &
      {\rotatebox[]{90}{\begin{tabular}[c]{c@{}c@{}}Seek\\Compact\\XR\end{tabular}}} &
      \makecell{\rotatebox[]{90}{8 HE}     } & 
      $\vcenter{{(\emph{pub})} This dataset is a collection of thermal infrared images and video frames of cheetahs.}$  & 
     \makecell{\rotatebox[]{90}{Mil. \& Sur.}  } \\
     \midrule

    \end{tabular}%
\end{table*}%

\begin{table*}[p]
  \centering
  \footnotesize
    \begin{tabular}{p{30pt}|p{43pt}|p{38pt}|p{16pt}|p{15pt}|p{26pt}|p{17pt}|p{135pt}|p{15pt}}
    \toprule
    Table \ref{datasetList} 
    &  continued... &
    \makecell{} & 
    \makecell{} & 
    & 
    \makecell{} & 
    \makecell{} & 
    \makecell{} & 
    \makecell{} \\
   \midrule

    \makecell{\rotatebox[]{90}{Comp-Mat} \rotatebox[]{90}{\cite{PMID:32995401}}}  & 
      $\vcenter{ \includegraphics[width=1.0\linewidth]{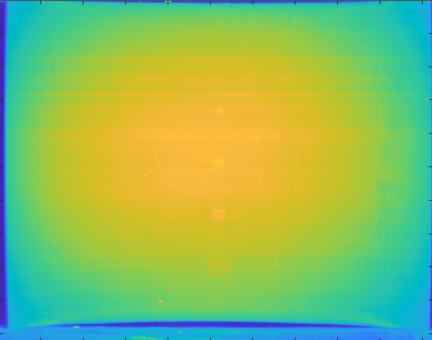}}$  & 
      { composite  materials }  & 
      \makecell{\rotatebox[]{90}{2000}  } & 
      \makecell{\rotatebox[]{90}{512X512} } & 
      \makecell{\rotatebox[]{90}{FLIR  X6900} } & 
      \makecell{\rotatebox[]{90}{8 HE}     } & 
      $\vcenter{{(\emph{pub})} The dataset consists of 12 image sequences collected for locating laminar defective regions from composite materials. The IR camera records the thermal evolution of both  sides of the sample for a few seconds for 3 stages, before a heat pulse, during  the heat pulse and after the heat pulse.}$  & 
     \makecell{\rotatebox[]{90}{Industrial}  } \\
     \midrule

   \makecell{\rotatebox[]{90}{CSIR CSIO} \rotatebox[]{90}{\cite{akula2014adaptive}}}  & 
      $\vcenter{ \includegraphics[width=1.0\linewidth]{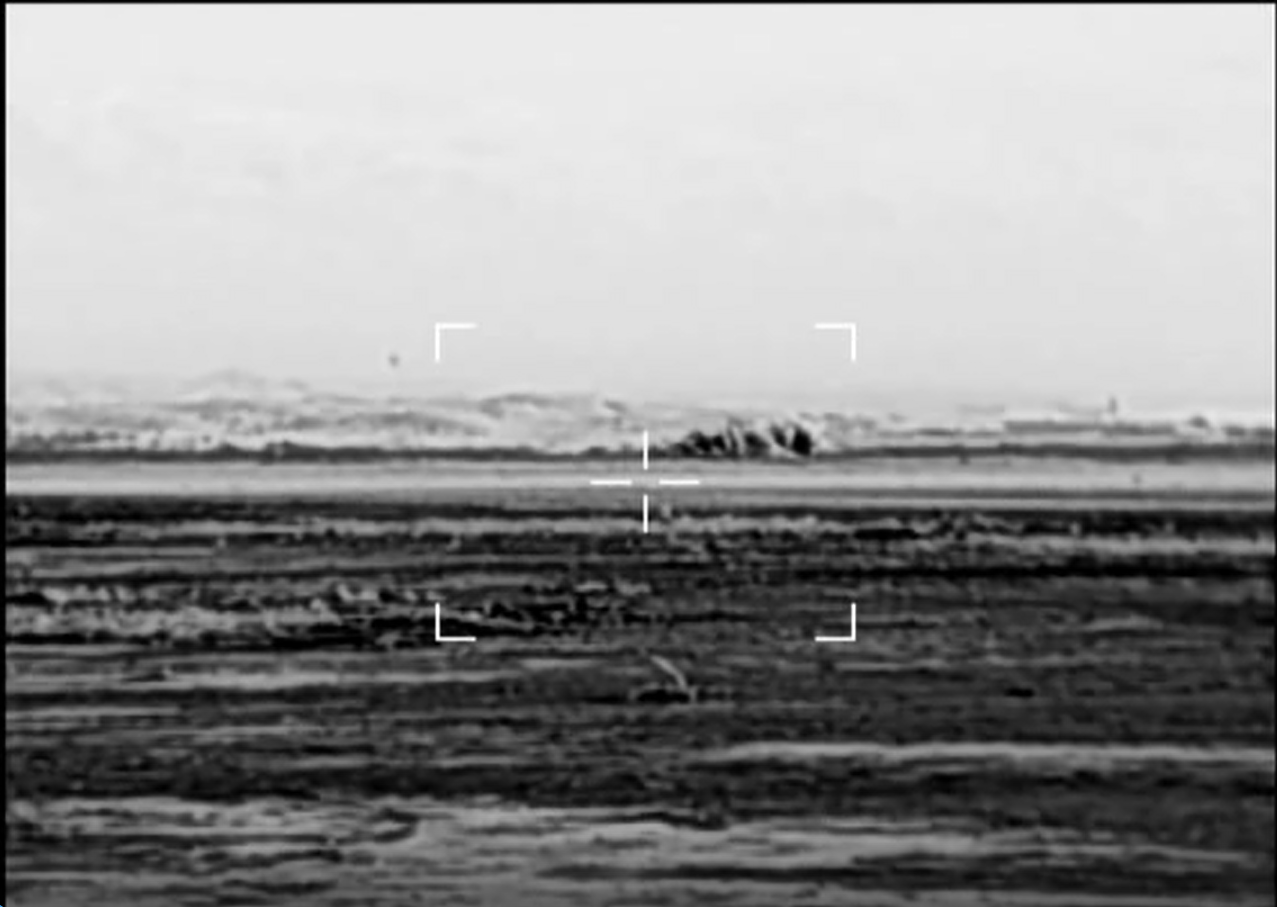}}$  & 
      {vehicle, human, dog, bird}  &
      \makecell{\rotatebox[]{90}{\begin{tabular}[c]{@{}c@{}} \centering{3650}\end{tabular}}} &
      \makecell{\rotatebox[]{90}{640x480}} & 
      \makecell{\rotatebox[]{90}{\begin{tabular}[c]{@{}c@{}}Uncool. µ-bol\end{tabular}}} &
      \makecell{\rotatebox[]{90}{8* HE} } & 
      $\vcenter{{(\emph{pub})} A moving object detection dataset which  has 18 sequences of moving targets, four, three and two-wheelers, pedestrians and some animals. Thermal infrared videos were captured at the coastal area of the Bay of Bengal in Southern India {with a sampling rate of 10hz.}}$  & 
      \makecell{\rotatebox[]{90}{Mil. \& Sur.}  } \\
     \midrule

    \makecell{\rotatebox[]{90}{CVC-09} \rotatebox[]{90}{\cite{CVC-09}}}  & 
      $\vcenter{ \includegraphics[width=1.0\linewidth]{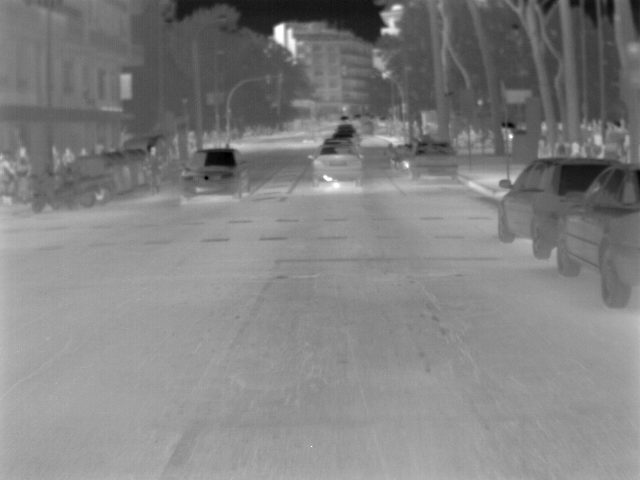}}$  & 
      {pedestrian}  & 
      \makecell{\rotatebox[]{90}{10K}} & 
      \makecell{\rotatebox[]{90}{640x480} } & 
      \makecell{\rotatebox[]{90}{not specified} } & 
      \makecell{\rotatebox[]{90}{8 HE}} & 
      $\vcenter{{(\emph{pub})} FIR Sequence Pedestrian Dataset include two sequences, day and night, captured at urban areas. Each sequence has about five thousand frames, divided into distinct train and test sets.}$  & 
      \makecell{\rotatebox[]{90}{Mil. \& Sur.}  } \\
     \midrule
     
     \makecell{\rotatebox[]{90}{CVC-14} \rotatebox[]{90}{\cite{CVC-14}}}  & 
      $\vcenter{ \includegraphics[width=1.0\linewidth]{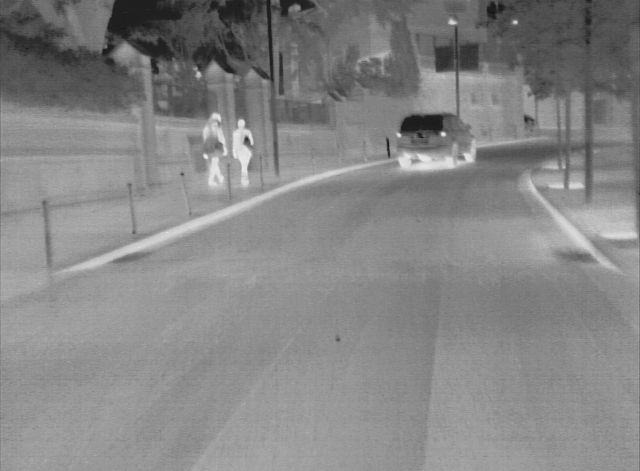}}$  & 
      {pedestrian}  & 
      \makecell{\rotatebox[]{90}{8400}  } & 
      \makecell{\rotatebox[]{90}{471x640} } & 
      \makecell{\rotatebox[]{90}{not  specified} } & 
      \makecell{\rotatebox[]{90}{8 HE} } & 
      $\vcenter{{(\emph{pub})} Visible-FIR Day-Night Pedestrian Sequence dataset includes two sequences of 3695  day images and 3390 night images,  with pedestrians annotated, again divided into distinct train and test sets.}$ & 
      \makecell{\rotatebox[]{90}{Mil. \& Sur.}  } \\
     \midrule

    \makecell{\rotatebox[]{90}{DDPM} \rotatebox[]{90}{\cite{DDPM}}}  & 
      $\vcenter{ \includegraphics[width=1.0\linewidth]{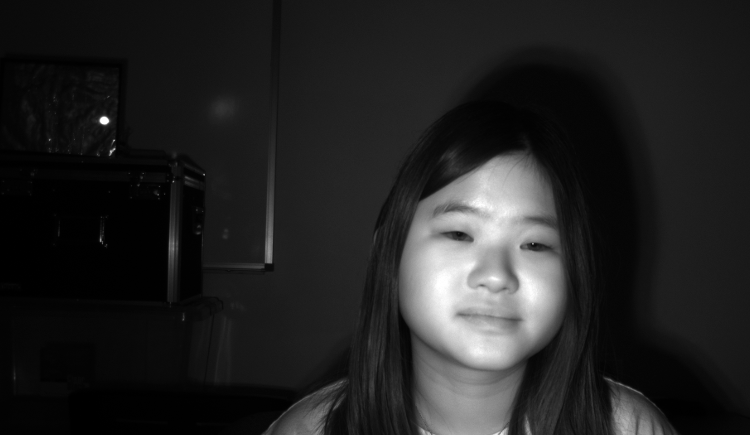}}$  & 
       {face}  & 
      \makecell{\rotatebox[]{90}{8M}   } & 
      {\rotatebox[]{90}{\begin{tabular}[c]{@{}c@{}}80x60\\ 1920×1080\end{tabular}}} &
      {\rotatebox[]{90}{\begin{tabular}[c]{@{}c@{}}DMK\\  33UX290\\FLIR C2\end{tabular}}} &
      \makecell{\rotatebox[]{90}{8 HE}     } & 
      $\vcenter{{(\emph{rr})} The dataset includes RBG and IR face videos of 13 hours of recordings of 70 subjects, for an interview scenario, in which the interviewee attempts to deceive the interviewer on selected responses.  Besides the RGB and IR  videos, the set includes various biometric sensor records.}$  & 
      \makecell{\rotatebox[]{90}{Mil. \& Sur.}  } \\
     \midrule
   
    \makecell{\rotatebox[]{90}{DIAST   } \rotatebox[]{90}{\cite{7763611} }} & 
      $\vcenter{ \includegraphics[width=1.0\linewidth]{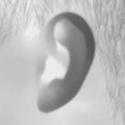}}$  & 
      {ear}    & 
      \makecell{\rotatebox[]{90}{2200}  } & 
      \makecell{\rotatebox[]{90}{125x125} } & 
      \makecell{\rotatebox[]{90}{FLIR  E60} } & 
      \makecell{\rotatebox[]{90}{8 HE}  } & 
      $\vcenter{{(\emph{pub})} The DIAST dataset contains visible and thermal ear images taken from a side face profile, from  55 subjects, collected for the task of ear recognition.}$  & 
      \makecell{\rotatebox[]{90}{Mil. \& Sur.}  } \\
     \midrule

     \makecell{\rotatebox[]{90}{Dim Small Air.} \rotatebox[]{90}{\cite{sciencedb_2019}}}  & 
      $\vcenter{ \includegraphics[width=1.0\linewidth]{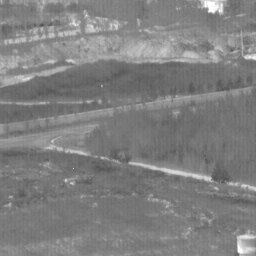}}$  & 
      {aircraft}  & 
      \makecell{\rotatebox[]{90}{16K}   } & 
      \makecell{\rotatebox[]{90}{256×256} } & 
      \makecell{\rotatebox[]{90}{not  specified} } & 
      \makecell{\rotatebox[]{90}{8* HE}   } & 
      $\vcenter{{(\emph{pub})} The dataset includes 22 sequences with sky and open-field backgrounds for infrared aircraft detection and recognition tasks.}$  & 
      \makecell{\rotatebox[]{90}{Mil. \& Sur.}   } \\
     \midrule
     
    \makecell{\rotatebox[]{90}{DMR-IR} \rotatebox[]{90}{\cite{breast_cancer}}}  & 
      $\vcenter{ \includegraphics[width=1.0\linewidth]{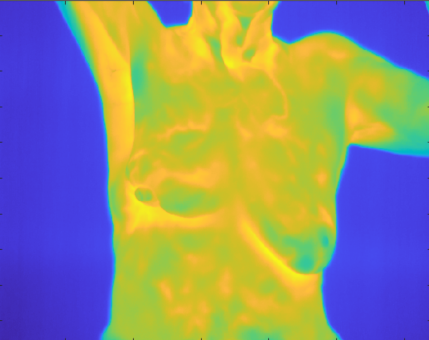}}$  & 
      {breast}  & 
      \makecell{\rotatebox[]{90}{1522}   } & 
      \makecell{\rotatebox[]{90}{640x480}  } & 
      \makecell{\rotatebox[]{90}{FLIR  SC-620}  } & 
      \makecell{\rotatebox[]{90}{8 HE}     } & 
      $\vcenter{{(\emph{pub})} The dataset consists of thermal breast images to be used for breast cancer  diagnosis. }$  & 
     \makecell{\rotatebox[]{90}{Medical} } \\
     \midrule
     
     \makecell{\rotatebox[]{90}{Dogs} \rotatebox[]{90}{\cite{nelson_2020}}}  & 
      $\vcenter{\includegraphics[width=0.7\linewidth]{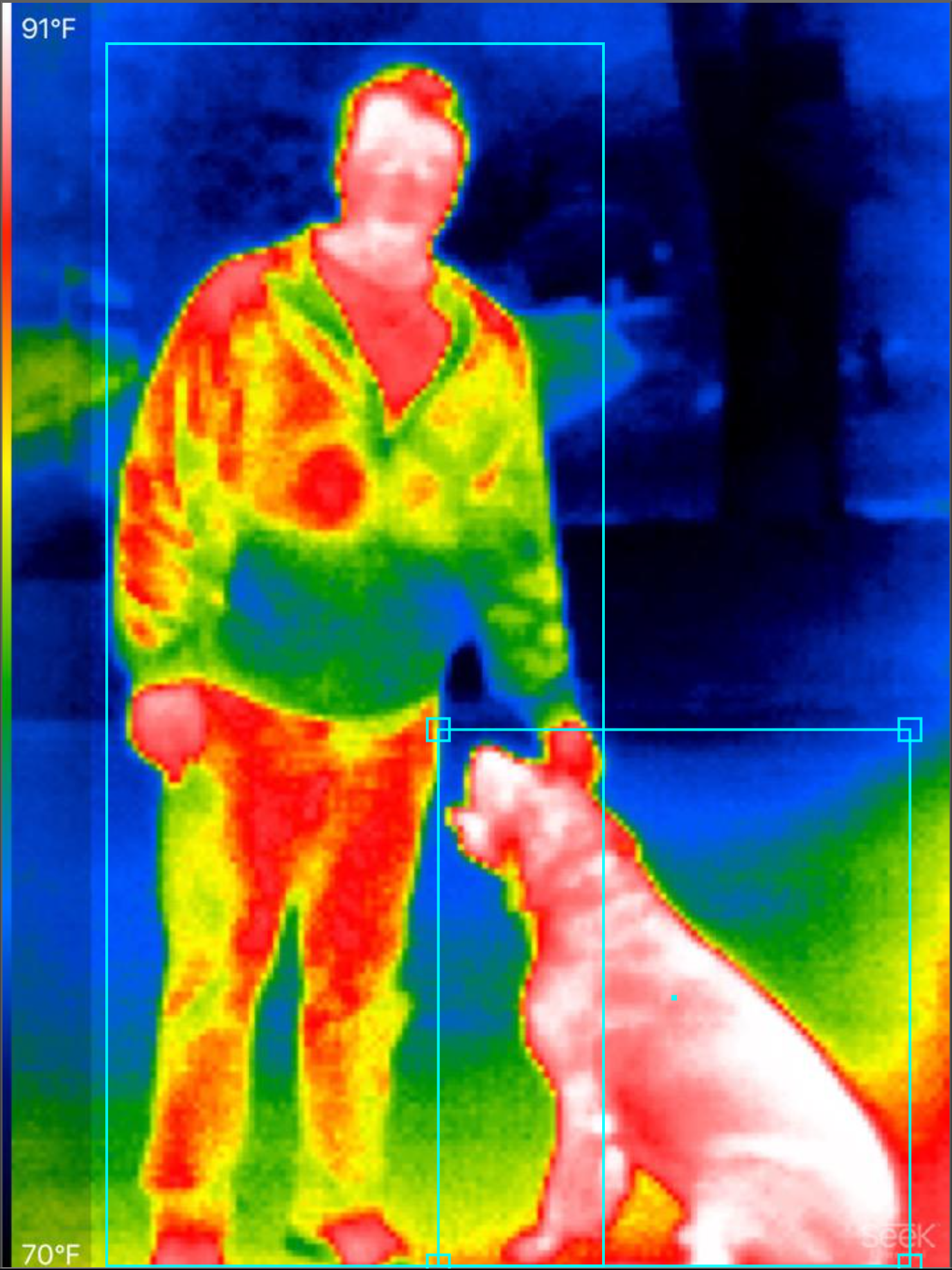}}$  & 
      {pedestrian, dog}  & 
      \makecell{\rotatebox[]{90}{203} } & 
      \makecell{\rotatebox[]{90}{416x416} } & 
      $\vcenter{\rotatebox[]{90}{\begin{tabular}[c]{c@{}c@{}}Seek\\Compact\\XR\end{tabular}}}$ &
      \makecell{\rotatebox[]{90}{8 HE} } & 
     $\vcenter{{(\emph{pub})} The dataset includes thermal infrared person and dog images captured in outdoor  environments.}$  & 
     \makecell{\rotatebox[]{90}{Mil. \& Sur.} } \\
     \midrule

     \makecell{\rotatebox[]{90}{{Drones}} \rotatebox[]{90}{\cite{drones}}}  & $\vcenter{\includegraphics[width=0.7\linewidth]{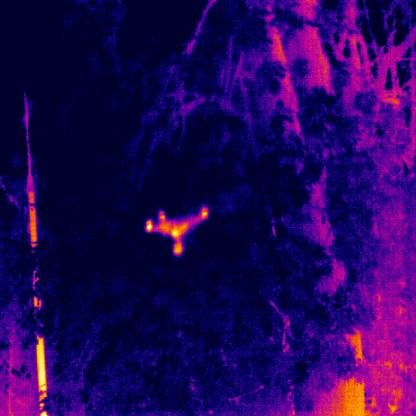}}$  & 
      {drone}  & 
      \makecell{\rotatebox[]{90}{150} } & 
      $\vcenter{\rotatebox[]{90}{\begin{tabular}[c]{c@{}c@{}}416x416\\640x480\end{tabular}}}$ & 
      \makecell{\rotatebox[]{90}{not  specified} } &
      \makecell{\rotatebox[]{90}{8* HE} } & 
     $\vcenter{{(\emph{pub})} The dataset contains thermal images for detecting UAVs. It provides ground truth bounding boxes for drones. }$  & 
     \makecell{\rotatebox[]{90}{Mil. \& Sur.} } \\
     \midrule

    \end{tabular}%
\end{table*}%

\begin{table*}[p]
  \centering
  \footnotesize
    \begin{tabular}{p{30pt}|p{43pt}|p{38pt}|p{16pt}|p{15pt}|p{26pt}|p{17pt}|p{135pt}|p{15pt}}
    \toprule
    Table \ref{datasetList}   &  continued... & 
    \makecell{} & 
    \makecell{} & 
    & 
    \makecell{} & 
    \makecell{} & 
    \makecell{} & 
    \makecell{} \\
   \midrule

    \makecell{\rotatebox[]{90}{EADS} \rotatebox[]{90}{\cite{robin} } } & 
      $\vcenter{ \includegraphics[width=1.0\linewidth]{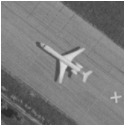}}$  & 
      \makecell[l]{plane, \\ vehicle} & 
      \makecell{\rotatebox[]{90}{not specified} } & 
      \makecell{\rotatebox[]{90}{not specified} } & 
      \makecell{\rotatebox[]{90}{not specified} } & 
      \makecell{\rotatebox[]{90}{8 HE} } & 
      $\vcenter{{(\emph{rr})} The dataset contains aerial images of vehicles, planes etc, for object detection tasks. Training and validation sets are provided with ground truths.}$ & 
     \makecell{\rotatebox[]{90}{Mil. \& Sur.}  } \\
     \midrule

     \makecell{\rotatebox[]{90}{{ESPOL FIR}} \rotatebox[]{90}{\cite{espol_fir} }}  & 
      $\vcenter{ \includegraphics[width=1.0\linewidth]{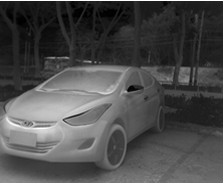}}$  & 
      \makecell[l]{person, \\vehicle, \\objects} & 
      \makecell{\rotatebox[]{90}{101} } & 
      \makecell{\rotatebox[]{90}{640x512} } & 
      \makecell{\rotatebox[]{90}{FLIR TAU 2} } & 
      \makecell{\rotatebox[]{90}{8 HE} } & 
      $\vcenter{{(\emph{pub})} The dataset contains infrared images taken in outdoor and indoor environments during the day. The dataset is primarily intended for super-resolution problems in the IR domain.}$ & 
     \makecell{\rotatebox[]{90}{Mil. \& Sur.}  } \\
     \midrule

   \makecell{\rotatebox[]{90}{Focus-Obj} \rotatebox[]{90}{\cite{thermal_focus}}}   & 
      $\vcenter{ \includegraphics[width=1.0\linewidth]{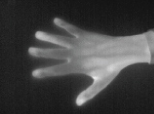}}$  & 
      $\vcenter{circuits, heater,  face, hand, etc.}$  & 
      \makecell{\rotatebox[]{90}{960}   } & 
      \makecell{\rotatebox[]{90}{160×120} } & 
      \makecell{\rotatebox[]{90}{\begin{tabular}[c]{@{}c@{}}TESTO\\880-3\end{tabular}}} &
      \makecell{\rotatebox[]{90}{\begin{tabular}[c]{@{}c@{}}{8 HE}\end{tabular}}} &
      $\vcenter{{(\emph{rr})} The database consists of several image sets. In each set, the camera acquires one image of the scene at 96 different lens positions.}$  & 
     \makecell{\rotatebox[]{90}{Industrial}  } \\
     \midrule

 \makecell{\rotatebox[]{90}{FREE FLIR   } \rotatebox[]{90}{\cite{FREE_FLIR}}}  & 
      $\vcenter{ \includegraphics[width=1.0\linewidth]{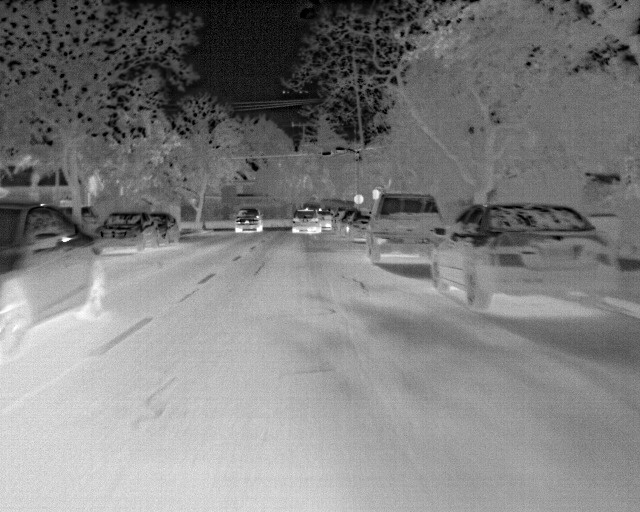}}$  & 
      \makecell[l]{vehicle\\ bicycle\\ dog\\ pedestrian}  & {\rotatebox[]{90}{\begin{tabular}[c]{@{}c@{}}14K images\\10k short videos\end{tabular}}} &
      \makecell{\rotatebox[]{90}{up to 1280x1024} } &  \makecell{\rotatebox[]{90}{\begin{tabular}[c]{@{}c@{}}IR Tau2\\FLIR  BlackFly\end{tabular}}} &
      \makecell{\rotatebox[]{90}{8/14 HE/RAW} } & 
      $\vcenter{{(\emph{rr})} The dataset contains 8 bit RGB and 8/14 bit thermal images images,  with annotations. The images are  captured on streets and highways of Santa Barbara, CA area, in different weather conditions, during day and night.}$   & 
      \rotatebox[]{90}{\begin{tabular}[c]{@{}c@{}}Mil.\& Sur.\\ Industrial\end{tabular}} \\
     \midrule
     
     \makecell{\rotatebox[]{90}{FV-USM}} \rotatebox[]{90}{\cite{FV-USM}}  & 
      $\vcenter{ \includegraphics[width=1.0\linewidth]{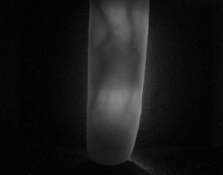}}$  & 
      {Finger  Vein}  & 
      \makecell{\rotatebox[]{90}{5904}   } & 
      \makecell{\rotatebox[]{90}{640x480} } & 
      \makecell{\rotatebox[]{90}{not  specified} } & 
      \makecell{\rotatebox[]{90}{8* HE}   } & 
      $\vcenter{{(\emph{rr})} The dataset contains infrared finger images of 83 male and 40 female volunteers for finger vein recognition tasks.}$  & 
      \makecell{\rotatebox[]{90}{Mil. \& Sur.}   } \\
     \midrule

   \makecell{\rotatebox[]{90}{HGRD} \rotatebox[]{90}{\cite{hand_gesture}}}   & 
      $\vcenter{ \includegraphics[width=1.0\linewidth]{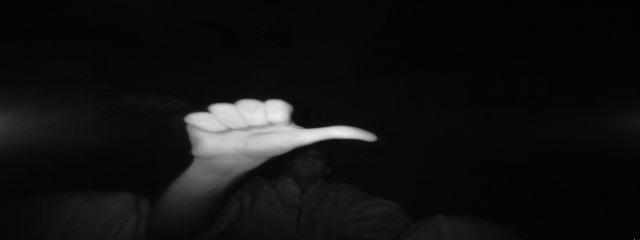}}$  & 
      {hand}  & 
       \makecell{\rotatebox[]{90}{20K}   } & 
      \makecell{\rotatebox[]{90}{640×240}  } & 
      \makecell{\rotatebox[]{90}{not specified} } & 
      \makecell{\rotatebox[]{90}{8 HE}    } & 
      $\vcenter{{(\emph{pub})} The Hand Gesture Recognition Dataset (HGRD) includes NIR images  of 10 different gestures, performed  by 10 different subjects (5 men and 5  women). It has 200 frames for each subject and each hand gesture.}$  & 
     \makecell{\rotatebox[]{90}{Industrial} } \\
     \midrule
     
     \makecell{\rotatebox[]{90}{HS-NIR} \rotatebox[]{90}{\cite{kaggle_2020}}}  & 
      $\vcenter{\includegraphics[width=1.0\linewidth]{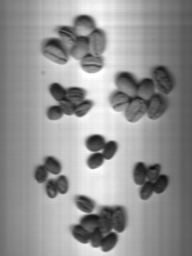}}$  & 
      $\vcenter{coffee  bean,  sugar,  floor,  salt}$ & 
      \makecell{\rotatebox[]{90}{480}   } & 
      \makecell{\rotatebox[]{90}{192x256} } & 
      \makecell{\rotatebox[]{90}{AHS- U20MIR} } & 
      \makecell{\rotatebox[]{90}{8 HE}     } & 
      $\vcenter{{(\emph{pub})} The dataset provides a total of 480 images of 5 objects. For each object, there are 96 different NIR hyperspectral images.}$  & 
      \makecell{\rotatebox[]{90}{Industrial}  } \\
     \midrule

   \makecell{\rotatebox[]{90}{IIT Delhi} \rotatebox[]{90}{\cite{iit_delhi}}}   & 
      $\vcenter{ \includegraphics[width=1.0\linewidth]{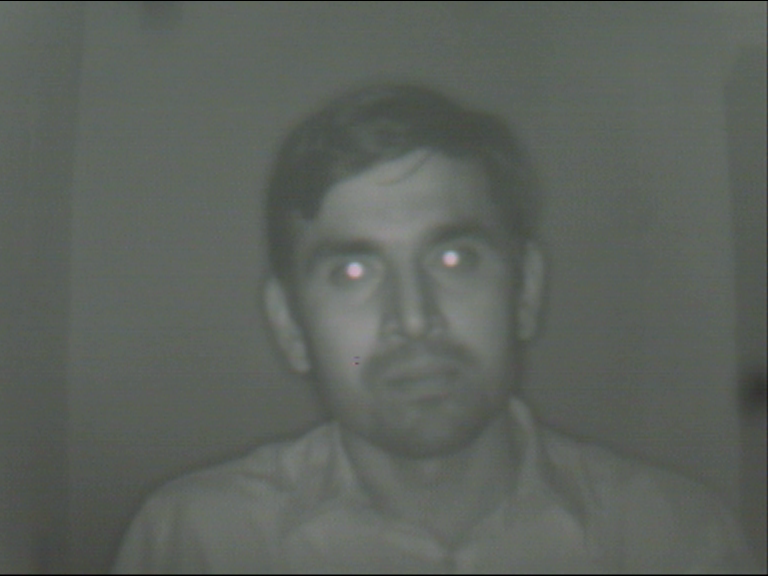}}$  & 
      {face}    & 
      \makecell{\rotatebox[]{90}{612}    } & 
      \makecell{\rotatebox[]{90}{768x576}  } & 
      \makecell{\rotatebox[]{90}{not  specified} } & 
      \makecell{\rotatebox[]{90}{8 HE}} & 
      $\vcenter{{(\emph{rr})} The database contains face images, captured at IIT Delhi campus with a webcam, shot with NIR-only illumination. }$   & 
      \makecell{\rotatebox[]{90}{Mil. \& Sur.}  } \\
     \midrule
     
     \makecell{\rotatebox[]{90}{Illegal Fishers } \rotatebox[]{90}{\cite{fishers}}}  & 
       $\vcenter{ \includegraphics[width=1.0\linewidth]{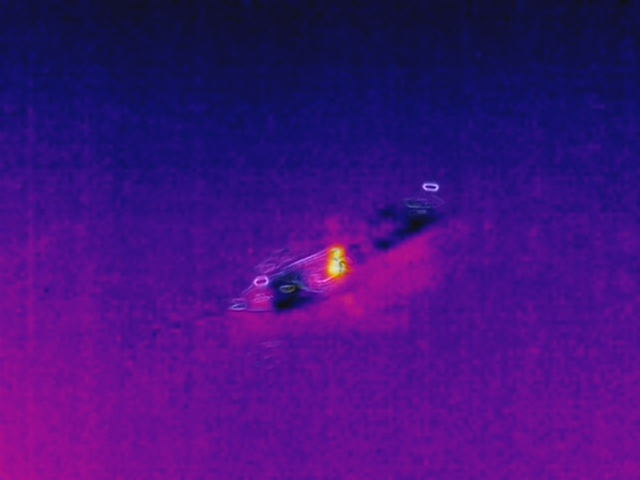}}$  & 
      {fishing  gear, ship}   & 
       \makecell{\rotatebox[]{90}{84} } & 
       {\rotatebox[]{90}{\begin{tabular}[c]{@{}c@{}} 640x480\\4056x3040\end{tabular}}} &
       \makecell{\rotatebox[]{90}{not  specified} } & 
       \makecell{\rotatebox[]{90}{8* HE}  } & 
       $\vcenter{{(\emph{pub})} The dataset has raw thermal, visible, and  night vision images captured by a drone  with a thermal camera to detect the  illegal fishing in Kuwaiti Bay.}$  & 
      \makecell{\rotatebox[]{90}{Mil. \& Sur.}  } \\
     \midrule

     \makecell{\rotatebox[]{90}{{Indoor-Outdoor IR }} \rotatebox[]{90}{\cite{indoor-outdoor}}}  & 
       $\vcenter{ \includegraphics[width=1.0\linewidth]{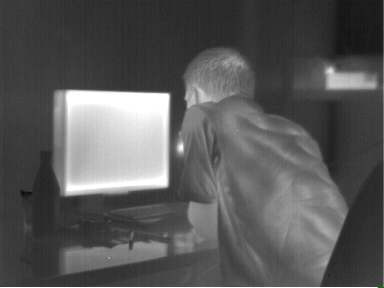}}$  & 
      {person, vehicle, objects}   & 
       \makecell{\rotatebox[]{90}{400} } & 
       \makecell{\rotatebox[]{90}{384x288}  } &
       \makecell{\rotatebox[]{90}{\begin{tabular}[c]{@{}c@{}}Thermoteknix\\Miricle 110KS\end{tabular}}}
        & 
       \makecell{\rotatebox[]{90}{8* HE}  } & 
       $\vcenter{{(\emph{pub})} The dataset presents indoor and outdoor IR-visible image pairs. The authors captured the images to compare the statistics of the visible and infrared images.}$  & 
      \makecell{\rotatebox[]{90}{Mil. \& Sur.}  } \\
     \midrule

    
    \end{tabular}%
  \label{tab:datasetList}%
\end{table*}%

\begin{table*}[p]
  \centering
  \footnotesize
    \begin{tabular}{p{30pt}|p{43pt}|p{38pt}|p{16pt}|p{15pt}|p{26pt}|p{17pt}|p{135pt}|p{15pt}}
    \toprule
    Table \ref{datasetList}   &  continued... & 
    \makecell{} & 
    \makecell{} & 
    & 
    \makecell{} & 
    \makecell{} & 
    \makecell{} & 
    \makecell{} \\
   \midrule  

   \makecell{\rotatebox[]{90}{{InfAR}} \rotatebox[]{90}{\cite{Infar}}}  & 
       $\vcenter{ \includegraphics[width=1.0\linewidth]{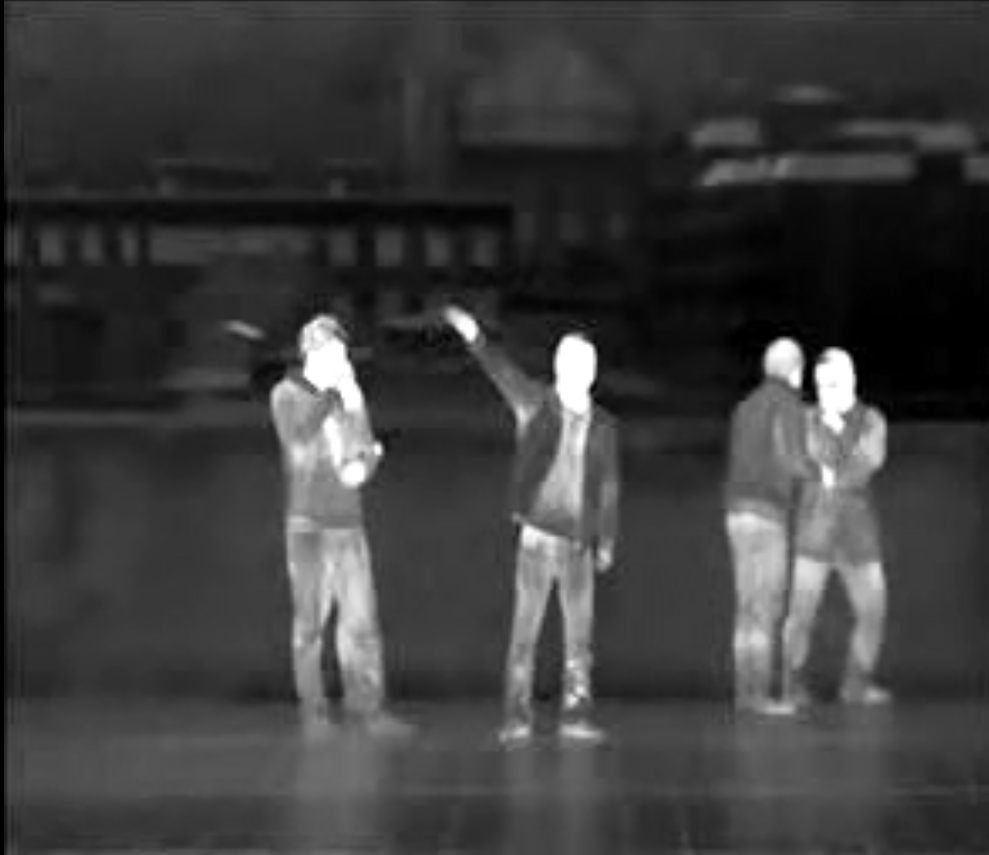}}$  & 
      {person}   & 
       {\rotatebox[]{90}{\makecell{3.6M\\(600 vid.)}}} & 
       \makecell{\rotatebox[]{90}{293×256}  }
        &
       \makecell{\rotatebox[]{90}{\begin{tabular}[c]{@{}c@{}}GUIDIR\\IR300\end{tabular}}} & 
       \makecell{\rotatebox[]{90}{8* HE}  } & 
       $\vcenter{{(\emph{pub})} The first publicly available dataset for infrared action recognition contains video action classes such as walking, hopping, fighting, etc. There are 12 human action classes, each with 50 videos.}$  & 
      \rotatebox[]{90}{\begin{tabular}[c]{@{}c@{}}Mil.\& Sur.\\ Scientific\end{tabular}} \\
     \midrule
     
     \makecell{\rotatebox[]{90}{IPATCH   } \rotatebox[]{90}{\cite{ipatch}}}  & 
      $\vcenter{ \includegraphics[width=1.0\linewidth]{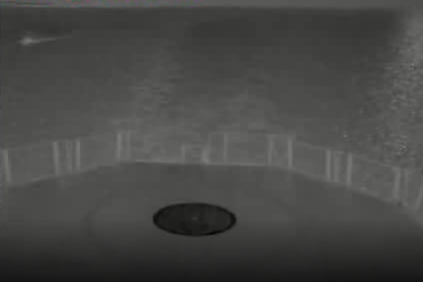}}$  & 
      {boat}   & 
       \makecell{\rotatebox[]{90}{not  specified} } & 
       {\rotatebox[]{90}{\begin{tabular}[c]{@{}c@{}}640x480\\640x512\end{tabular}}} &
       \makecell{\rotatebox[]{90}{\begin{tabular}[c]{@{}c@{}}FLIR SC655\\FLIR  A65\end{tabular}}} &
       \makecell{\rotatebox[]{90}{not  specified}     } & 
       $\vcenter{{(\emph{pri})} The IPATCH dataset contains 14 thermal videos from the coast of Brest, France, for object detection and tracking, event detection, and threat recognition tasks of maritime piracy. There are selected sequences of abnormal events, such as boat speeding up, boat loitering, boat moving around vessel, etc.}$  & 
     \makecell{\rotatebox[]{90}{Mil. \& Sur.} } \\
     \midrule

   \makecell{\rotatebox[]{90}{IRIS  } \rotatebox[]{90}{\cite{IRIS}}}  & 
      $\vcenter{ \includegraphics[width=1.0\linewidth]{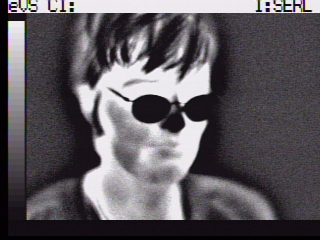}}$  & 
       {face}   & 
      \makecell{\rotatebox[]{90}{4228}  } & 
      \makecell{\rotatebox[]{90}{320x240} } & 
      \makecell{\rotatebox[]{90}{\begin{tabular}[c]{@{}c@{}}Raytheon\\Palm IR Pro\end{tabular}}} &
      \makecell{\rotatebox[]{90}{8* HE}   } & 
      $\vcenter{{(\emph{pub})} The dataset consists of thermal and visible face images with various expressions,  poses and illumination conditions.}$  & 
      \makecell{\rotatebox[]{90}{Mil. \& Sur.}   } \\
     \midrule

 \makecell{\rotatebox[]{90}{IRShips  } \rotatebox[]{90}{\cite{westlake2020deep}}}  & 
      $\vcenter{ \includegraphics[width=1.0\linewidth]{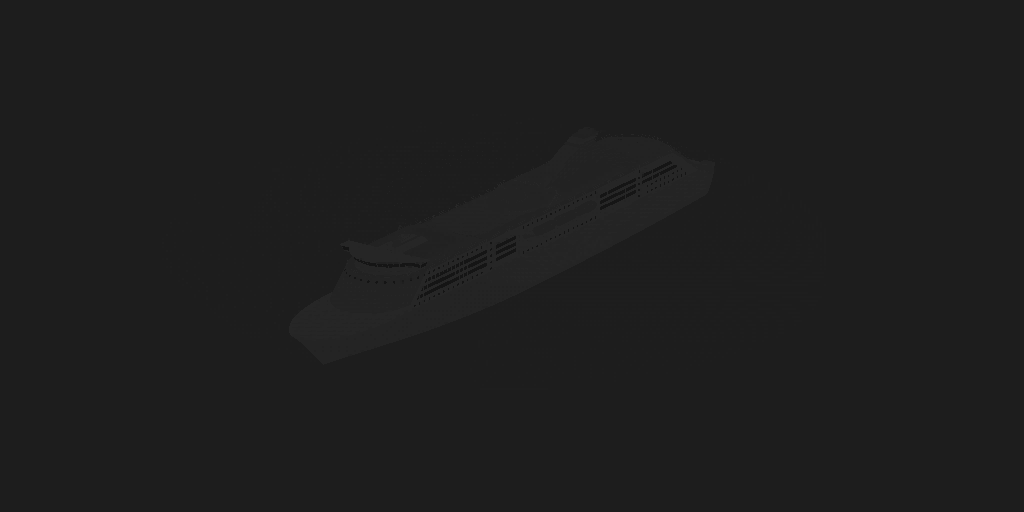}}$  & 
       {ship}   & 
      \makecell{\rotatebox[]{90}{972K}  } & 
      \makecell{\rotatebox[]{90}{1024x512} } & 
      \makecell{\rotatebox[]{90}{N/A} } & 
      \makecell{\rotatebox[]{90}{8 HE}    } & 
      $\vcenter{{(\emph{pub})} This dataset contains synthetically generated IR images of 10 different ships.}$   & 
      \makecell{\rotatebox[]{90}{Mil. \& Sur.}  } \\
     \midrule
     
     \makecell{\rotatebox[]{90}{JPL } \rotatebox[]{90}{\cite{JPL}}}  & 
      $\vcenter{ \includegraphics[width=1.0\linewidth]{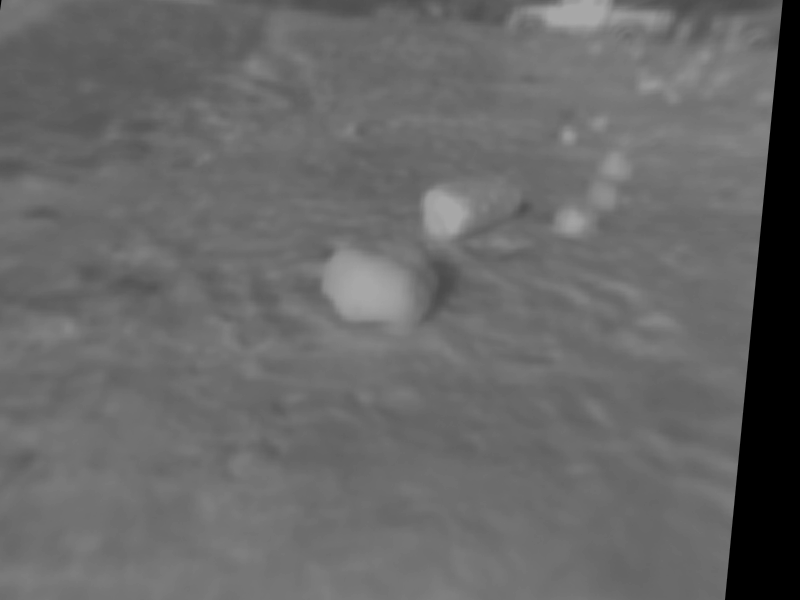}}$  & 
      $\vcenter{sand, soil, rocks, bedrock, rocky terrain, etc.} $  & 
      \makecell{\rotatebox[]{90}{1300}  } & 
      \makecell{\rotatebox[]{90}{800x600} } & 
      \makecell{\rotatebox[]{90}{FLIR  AX65} } & 
      \makecell{\rotatebox[]{90}{8 HE}   } & 
      $\vcenter{{(\emph{pub})} The dataset comprises IR and RGB images, in 2 parts: 1) the Semantic Dataset for Terrain Types and 2) the Virtual Sensor Dataset for deriving RGB-to-IR  mapping models. Semantic Dataset includes manually annotated 7 categories: unlabeled,  sand, soil, rocks, bedrock, rocky terrain, and ballast.}$  & 
      \makecell{\rotatebox[]{90}{ Scientific}   } \\
     \midrule

        \makecell{\rotatebox[]{90}{KAIST} \rotatebox[]{90}{\cite{hwang2015multispectral}} } & 
      $\vcenter{ \includegraphics[width=1.0\linewidth]{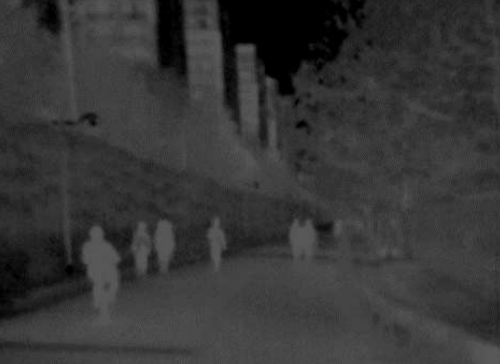}}$  & 
      {pedestrian}  & 
      \makecell{\rotatebox[]{90}{95K}} & 
      \makecell{\rotatebox[]{90}{320x256} } & 
      \makecell{\rotatebox[]{90}{FLIR  A35}} & 
      \makecell{\rotatebox[]{90}{8 HE}} & 
      $\vcenter{{(\emph{pub})} The dataset contains multispectral images of vehicles captured in day and  night traffic with annotations.}$  & 
      \makecell{\rotatebox[]{90}{Mil. \& Sur.}  } \\
     \midrule
     
     \makecell{\rotatebox[]{90}{Kayak Image Fusion}\rotatebox[]{90}{ \cite{toet2002detection}}}  & 
      $\vcenter{ \includegraphics[width=1.0\linewidth]{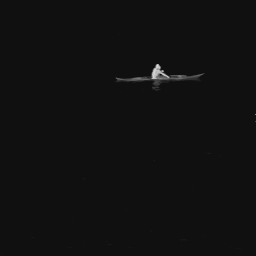}}$  & 
      {kayak}  & 
      \makecell{\rotatebox[]{90}{2541}  } & 
      \makecell{\rotatebox[]{90}{up to  752x582} } & 
      {\rotatebox[]{90}{\begin{tabular}[c]{@{}c@{}c@{}}Raytheon Rad.\\ HS, AEG AIM\\ 256  PLW,\\Philips  LTC500\end{tabular}}} &
      \makecell{\rotatebox[]{90}{8-8* HE}   } & 
      $\vcenter{{(\emph{pub})} This dataset includes sequences with several kayaks in a maritime environment. The images  are captured in MWIR, LWIR and visible domains.}$  & 
      \makecell{\rotatebox[]{90}{Mil. \& Sur.}  } \\
     \midrule
     
     \makecell{\rotatebox[]{90}{L-CAS FACE} \rotatebox[]{90}{\cite{8513201} }}  & 
      $\vcenter{ \includegraphics[width=1.0\linewidth]{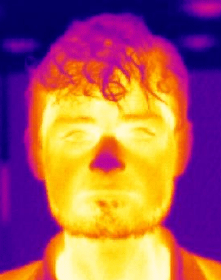}}$  & 
      {face}   & 
      \makecell{\rotatebox[]{90}{3000}   } & 
      \makecell{\rotatebox[]{90}{382x288}  } & 
      \makecell{\rotatebox[]{90}{Optris  PI-450}  } & 
      \makecell{\rotatebox[]{90}{8 HE}     } & 
      $\vcenter{{(\emph{rr})} The dataset comprises of thermal images of  moving faces, captured with a sensor that is mounted on the top of a robot to measure respiration and heartbeat rate for physiological monitoring. The dataset provides ground truth for respiration and  heartbeat. }$  & 
     \makecell{\rotatebox[]{90}{Mil. \& Sur.} } \\
     \midrule

   \makecell{\rotatebox[]{90}{L-CAS ReID} \rotatebox[]{90}{\cite{Cosar2019}}}  & 
      $\vcenter{ \includegraphics[width=1.0\linewidth]{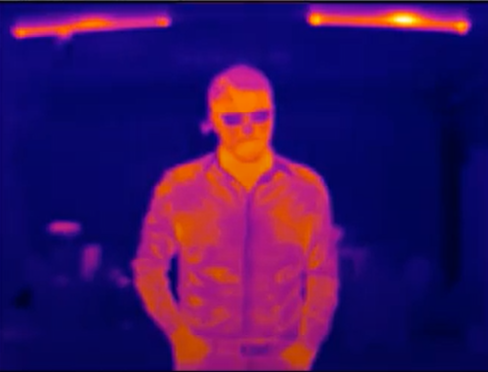}}$  & 
     {person}  & 
      \makecell{\rotatebox[]{90}{36K}   } & 
      \makecell{\rotatebox[]{90}{382×288} } & 
      \makecell{\rotatebox[]{90}{Optris  PI-450} } & 
      \makecell{\rotatebox[]{90}{8 HE}  } & 
      $\vcenter{{(\emph{rr})} The dataset includes RGB, depth, and thermal  images of people for  re-identification tasks. The set is collected via a thermal camera on a robot.}$  & 
     \makecell{\rotatebox[]{90}{Mil. \& Sur.}  } \\
     \midrule

    \makecell{\rotatebox[]{90}{Leaves} \rotatebox[]{90}{\cite{diseased}}}  & 
      $\vcenter{ \includegraphics[width=1.0\linewidth]{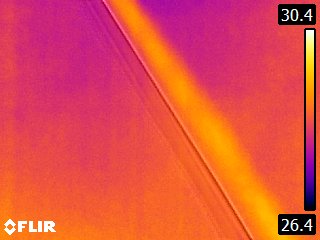}}$  & 
      {leaf}  & 
      \makecell{\rotatebox[]{90}{636}   } & 
      \makecell{\rotatebox[]{90}{320x240}  } &
      \makecell{\rotatebox[]{90}{\begin{tabular}[c]{@{}c@{}}FLIR\\E8\end{tabular}}} &
      \makecell{\rotatebox[]{90}{8 HE}     } & 
      $\vcenter{{(\emph{pub})} The dataset, composed of thermal leaf  images, is categorised into 6 categories: bacteria  leaf blight, blast, leaf spot, leaf folder,  hispa, healthy leaves. }$  & 
     \makecell{\rotatebox[]{90}{Industrial} } \\
     \midrule

    \end{tabular}%
  \label{tab:dataset13}%
\end{table*}%

\begin{table*}[p]
  \centering
  \footnotesize
    \begin{tabular}{p{30pt}|p{43pt}|p{38pt}|p{16pt}|p{15pt}|p{26pt}|p{17pt}|p{135pt}|p{15pt}}
    \toprule
    Table \ref{datasetList}   &  continued... & 
    \makecell{} & 
    \makecell{} & 
    & 
    \makecell{} & 
    \makecell{} & 
    \makecell{} & 
    \makecell{} \\
   \midrule 

    \makecell{\rotatebox[]{90}{{LLVIP}} \rotatebox[]{90}{\cite{llvip}}} & 
      $\vcenter{ \includegraphics[width=1.0\linewidth]{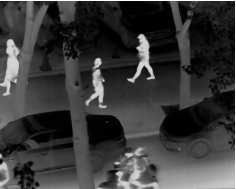}}$  & $\vcenter{person, vehicle}$ & 
      \makecell{\rotatebox[]{90}{31K} } & 
      {\rotatebox[]{90}{\begin{tabular}[c]{@{}c@{}}1920×1080\\1280×720\end{tabular}}}
       & 
      {\rotatebox[]{90}{\begin{tabular}[c]{@{}c@{}}Hikvision \\DS-2TD8166\\75C2F/V2\end{tabular}}} & 
      {\rotatebox[]{90}{\begin{tabular}[c]{@{}c@{}}8 HE \\16 RAW\end{tabular}}} & 
      $\vcenter{{(\emph{rr})} Contains infrared and visible images of pedestrians in low-light conditions. The images contain vehicles, but pedestrians are only labelled. Raw data can also be obtained.}$  & 
      \makecell{\rotatebox[]{90}{Mil. \& Sur.}  } \\
     \midrule

     \makecell{\rotatebox[]{90}{{LR-MR-HR FIR}} \rotatebox[]{90}{\cite{LRMRHR}}} & 
      $\vcenter{ \includegraphics[width=1.0\linewidth]{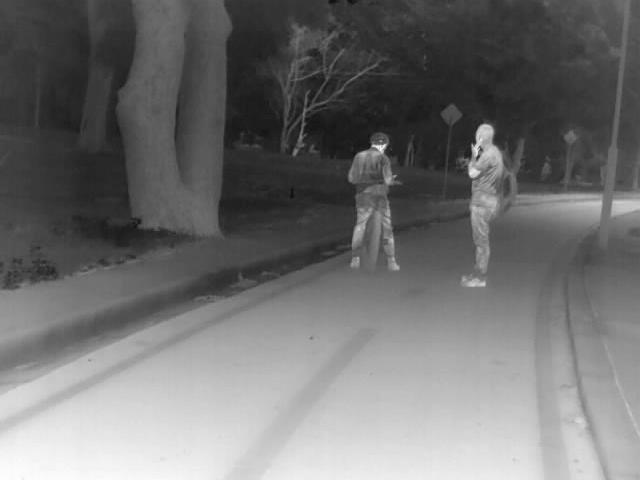}}$  & $\vcenter{person, animal, vehicle, objects}$ & 
      \makecell{\rotatebox[]{90}{3063} } & 
      {\rotatebox[]{90}{\begin{tabular}[c]{@{}c@{}}up to\\640x512\end{tabular}}}
       & 
      {\rotatebox[]{90}{\begin{tabular}[c]{@{}c@{}}Axis Domo P1290\\Axis Q2901-E\\FC-6320 FLIR\end{tabular}}} & 
      \makecell{\rotatebox[]{90}{8 HE}    } & 
      $\vcenter{{(\emph{pub})} The dataset contains thermal images at different resolutions for image super-resolution tasks. Various levels of resolution are available: low, medium and high resolution, which correspond to 160x120, 320x240, and 640x512 pixels, respectively. There are 1021 images per resolution.}$  & 
      \makecell{\rotatebox[]{90}{Mil. \& Sur.}  } \\
     \midrule

   \makecell{\rotatebox[]{90}{LSI  } \rotatebox[]{90}{\cite{LSI}}} & 
      $\vcenter{ \includegraphics[width=1.0\linewidth]{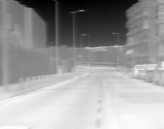}}$  & 
      {pedestrian}  & 
      \makecell{\rotatebox[]{90}{20K} } & 
      {\rotatebox[]{90}{\begin{tabular}[c]{@{}c@{}}32x64\\164x128\end{tabular}}} &
      \makecell{\rotatebox[]{90}{Indigo Omega} } & 
      \makecell{\rotatebox[]{90}{14 RAW}    } & 
      $\vcenter{{(\emph{pub})} The set consists of FIR pedestrian images in an outdoor urban environment, divided into train and test sets, with manually annotated bounding boxes.}$  & 
      \makecell{\rotatebox[]{90}{Mil. \& Sur.}  } \\
     \midrule

   \makecell{\rotatebox[]{90}{LSOTB TIR} \rotatebox[]{90}{\cite{LSOTB-TIR}}}  & 
      $\vcenter{ \includegraphics[width=1.0\linewidth]{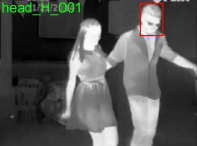}}$  & 
      $\vcenter{person, animal, vehicle, aircraft, boat}$  & 
      \makecell{\rotatebox[]{90}{600K} } & 
      \makecell{\rotatebox[]{90}{not  specified} } & 
      \makecell{\rotatebox[]{90}{not  specified} } & 
      \makecell{\rotatebox[]{90}{8 HE} } & 
      $\vcenter{{(\emph{pub})} Contains various object classes with 730K bounding box annotations, shot in various environments, including urban areas, forests, the sea, etc.}$  & 
      \makecell{\rotatebox[]{90}{Mil. \& Sur.}  } \\
     \midrule

\makecell{\rotatebox[]{90}{LTIR} \rotatebox[]{90}{\cite{ltir2015}}} & 
      $\vcenter{ \includegraphics[width=1.0\linewidth]{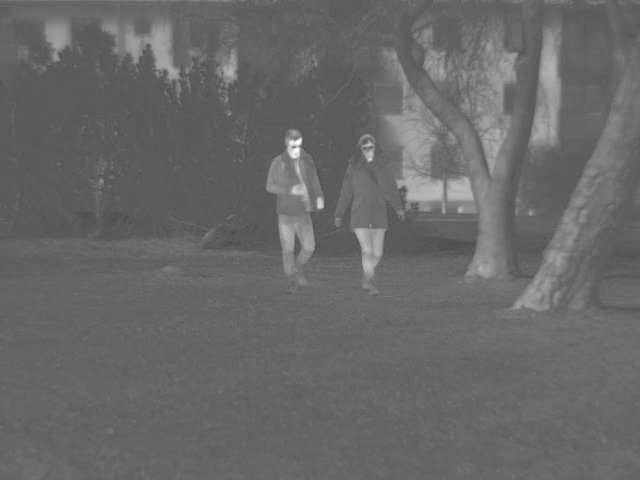}}$  & 
      $\vcenter{rhino, human, horse, car, quadrocopter, dog}$  & 
      \makecell{\rotatebox[]{90}{11K}   } & 
      \rotatebox[]{90}{\makecell{up to\\1920x480}} &
      \rotatebox[]{90}{\makecell{FLIR A35\\FLIR Tau320\\FLIR A655SC}} &
      \rotatebox[]{90}{\makecell{8/16\\HE/RAW}} &
      $\vcenter{{(\emph{pub})} The dataset includes 20 thermal IR sequences of different objects, captured at indoor and outdoor environments, with annotations.}$  & 
      \makecell{\rotatebox[]{90}{Mil. \& Sur.}   } \\
     \midrule
     
     \makecell{\rotatebox[]{90}{Mango  } \rotatebox[]{90}{\cite{Mango}}}  & 
      $\vcenter{ \includegraphics[width=1.0\linewidth]{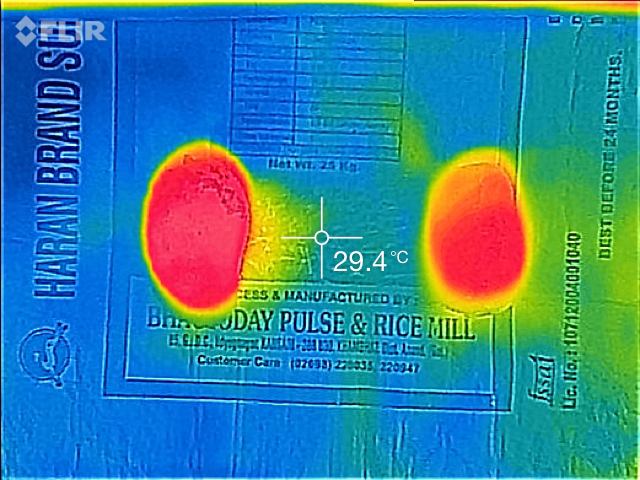}}$  & 
      {mango}  & 
      \makecell{\rotatebox[]{90}{309}   } & 
      \makecell{\rotatebox[]{90}{640x480} } & 
      \makecell{\rotatebox[]{90}{FLIR One} } & 
      \makecell{\rotatebox[]{90}{8* HE} } & 
      $\vcenter{{(\emph{pub})} The dataset consists of thermal mango images for object counting and colour feature extraction tasks.}$ & 
      \makecell{\rotatebox[]{90}{Industrial}  } \\
     \midrule
     
     \makecell{\rotatebox[]{90}{MBD } \rotatebox[]{90}{\cite{MBD}}}  & 
      $\vcenter{ \includegraphics[width=1.0\linewidth]{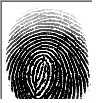}}$  & 
      $\vcenter{finger-print,  finger-vein, face, iris}$  & 
      \makecell{\rotatebox[]{90}{72} } & 
      \makecell{\rotatebox[]{90}{up to 640x480} } & 
      \makecell{\rotatebox[]{90}{not  specified} } & 
      \makecell{\rotatebox[]{90}{8/8*/32 HE} } & 
      $\vcenter{{(\emph{pub})} The set includes 45 thermal fingerprints, 9 thermal finger veins, 9 thermal iris,  and 9 thermal face images.}$  & 
      \makecell{\rotatebox[]{90}{Mil. \& Sur.} } \\
     \midrule
     
     \makecell{\rotatebox[]{90}{MBDA} \rotatebox[]{90}{\cite{robin}}}  & 
      $\vcenter{ \includegraphics[width=1.0\linewidth]{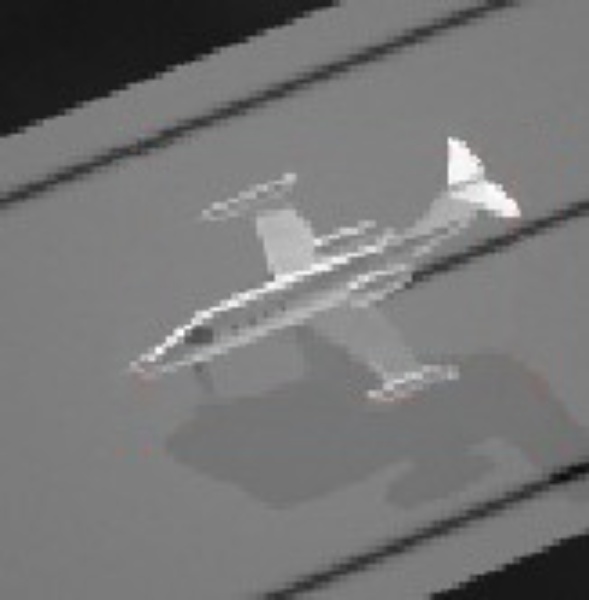}}$  & 
      $\vcenter{plane, helicopter, vehicle, tower}$  & 
      \makecell{\rotatebox[]{90}{>15K} } & 
      {\rotatebox[]{90}{\makecell{320x240\\256x256}}} &
      \makecell{\rotatebox[]{90}{N/A} } & 
      \makecell{\rotatebox[]{90}{14/16 RAW} } & 
      $\vcenter{{(\emph{rr})} The dataset contains  computer-generated aircraft and vehicle images. Train and validation sets are provided  with ground truths. }$  & 
     \makecell{\rotatebox[]{90}{Mil. \& Sur.} } \\
     \midrule
     
     \makecell{\rotatebox[]{90}{METU} \rotatebox[]{90}{\cite{kinect}}} & 
      $\vcenter{ \includegraphics[width=1.0\linewidth]{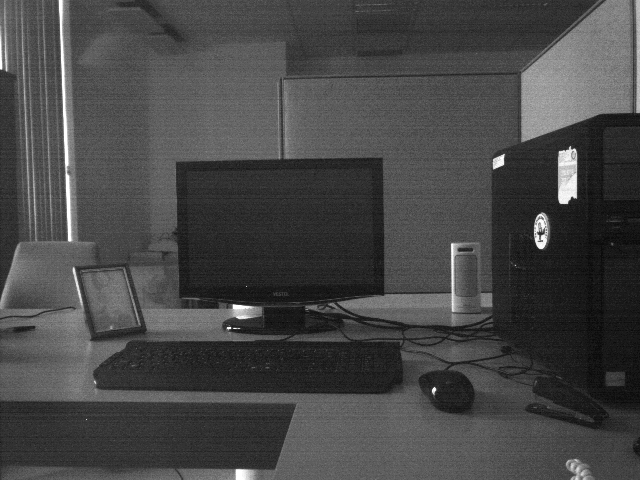}}$  & 
      $\vcenter{various  objects}$  & 
      \makecell{\rotatebox[]{90}{24}   } & 
      \makecell{\rotatebox[]{90}{640x480} } & 
      \makecell{\rotatebox[]{90}{Kinect} } & 
      \makecell{\rotatebox[]{90}{8 HE}} & 
      $\vcenter{{(\emph{pub})} METU Kinect dataset consists of IR and visible image pairs of various objects.}$  & 
      \makecell{\rotatebox[]{90}{Scientific}   } \\
     \midrule
     
     \makecell{\rotatebox[]{90}{MIntPAIN  } \rotatebox[]{90}{\cite{haque2018deep}}}  & 
      $\vcenter{ \includegraphics[width=1.0\linewidth]{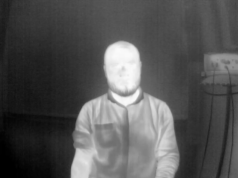}}$  & 
      {person, face}  & 
      {\rotatebox[]{90}{\makecell{2M\\(9366 vid.)}}} &
      \makecell{\rotatebox[]{90}{640x480} } & 
      \makecell{\rotatebox[]{90}{Axis Q  1922} } & 
      \makecell{\rotatebox[]{90}{8 HE}      } & 
      $\vcenter{{(\emph{rr})} The MIntPAIN is an RGB, Depth and Thermal (RGBDT) image set, which contains RGBDT videos of 20 subjects. Each  subject has 80 folders in visible, depth,  and thermal domains for pain level recognition tasks.}$ & 
      \makecell{\rotatebox[]{90}{Med.\& Sci.}  } \\
     \midrule

    \end{tabular}%
  \label{tab:dataset13}%
\end{table*}%

\begin{table*}[p]
  \centering
  \footnotesize
    \begin{tabular}{p{30pt}|p{43pt}|p{38pt}|p{16pt}|p{15pt}|p{26pt}|p{17pt}|p{135pt}|p{15pt}}
    \toprule
    Table \ref{datasetList}   &  continued... & 
    \makecell{} & 
    \makecell{} & 
    & 
    \makecell{} & 
    \makecell{} & 
    \makecell{} & 
    \makecell{} \\
   \midrule 

   \makecell{\rotatebox[]{90}{MS Focus} \rotatebox[]{90}{\cite{multipspec}}}  & 
      $\vcenter{ \includegraphics[width=1.0\linewidth]{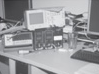}}$  & 
      $\vcenter{building, car, corridor, head, keyboard, office desk, pens}$  & 
      \makecell{\rotatebox[]{90}{420}   } & 
      \makecell{\rotatebox[]{90}{640x480} } & 
      \makecell{\rotatebox[]{90}{\begin{tabular}[c]{@{}c@{}}Canon EOS 350D\\FLIR SC660\end{tabular}}} &
      \makecell{\rotatebox[]{90}{8 HE}     } & 
      $\vcenter{{(\emph{pub})} Contains 420 images of 7 objects in different optical focus positions, captured in visible, near-infrared, and thermal spectrum.}$  & 
      \makecell{\rotatebox[]{90}{Industrial} } \\
     \midrule

   \makecell{\rotatebox[]{90}{MM-Bio} \rotatebox[]{90}{\cite{Biometrics}}}  & 
      $\vcenter{ \includegraphics[width=1.0\linewidth]{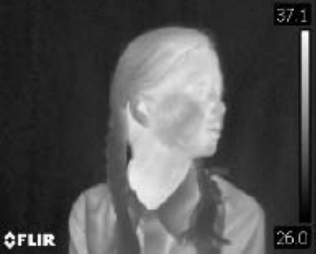}}$  & 
      {face}   & 
      \makecell{\rotatebox[]{90}{2500} } & 
      \makecell{\rotatebox[]{90}{not   specified}  } & 
      \makecell{\rotatebox[]{90}{FLIR  E40} } & 
      \makecell{\rotatebox[]{90}{8 HE}     } & 
      $\vcenter{{(\emph{rr})} A multi-model biometric dataset containing thermal and RGB images of 125 people.}$  & 
     \makecell{\rotatebox[]{90}{Mil. \& Sur.} } \\
     \midrule
     
     \makecell{\rotatebox[]{90}{MM-Hand} \rotatebox[]{90}{\cite{multi_modal}}}  & 
      $\vcenter{ \includegraphics[width=1.0\linewidth]{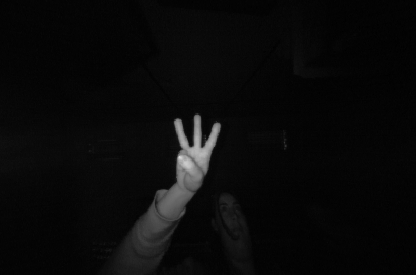}}$  & 
       {hand}     & 
       \makecell{\rotatebox[]{90}{65K}   } & 
      \makecell{\rotatebox[]{90}{412x273} } & 
      \makecell{\rotatebox[]{90}{not  specified} } & 
      \makecell{\rotatebox[]{90}{8 HE}     } & 
      $\vcenter{{(\emph{pub})} Multi-modal  Hand  Gesture  Recognition dataset presents near-infrared images  of 15 different hand gestures from 15  different subjects (5 women and 10 men).}$  & 
     \makecell{\rotatebox[]{90}{Scientific}  } \\
     \midrule
     
     \makecell{\rotatebox[]{90}{Mov-Tar} \rotatebox[]{90}{\cite{movingTarget}}}  & 
      $\vcenter{ \includegraphics[width=1.0\linewidth]{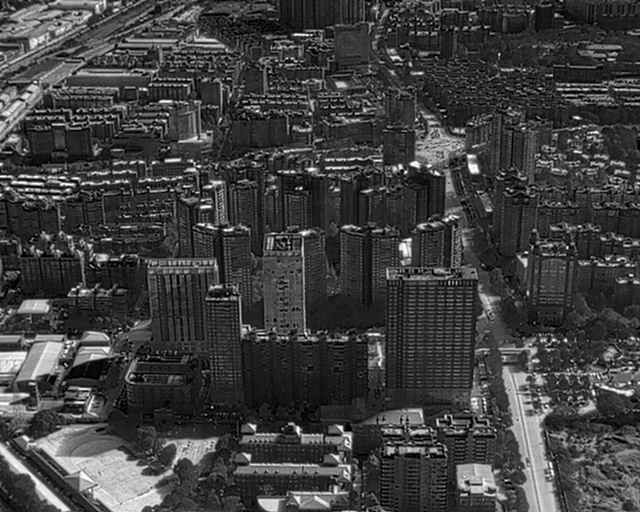}}$  & {vegetation, building}  & 
      \makecell{\rotatebox[]{90}{150K}   } & 
      \makecell{\rotatebox[]{90}{640x512} } & 
      \makecell{\rotatebox[]{90}{not specified} } & 
      \makecell{\rotatebox[]{90}{8 HE} } & 
      $\vcenter{{(\emph{pub})} This dataset contains synthetic small infrared targets embedded in IR background images to be used for target detection and tracking applications.}$  & 
      \makecell{\rotatebox[]{90}{Mil. \& Sur.}  } \\
     \midrule

\makecell{\rotatebox[]{90}{{MS-SS}} \rotatebox[]{90}{\cite{ms-ss}}}  & 
      $\vcenter{ \includegraphics[width=1.0\linewidth]{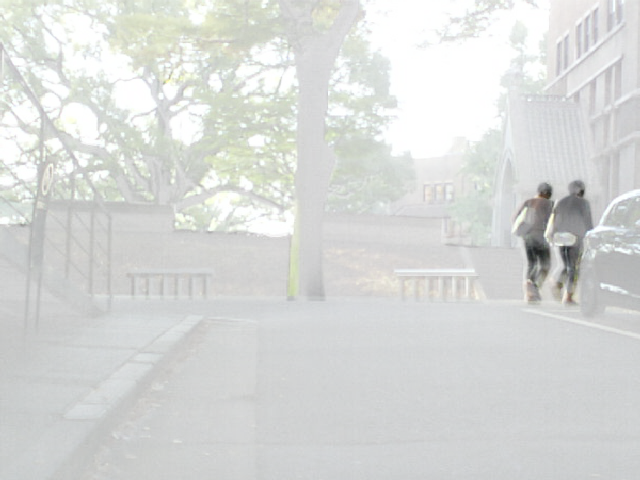}}$  & 
      $\vcenter{car, person, bike}$  & 
      \makecell{\rotatebox[]{90}{1569}   } & 
      \makecell{\rotatebox[]{90}{480×640} } & 
      \makecell{\rotatebox[]{90}{\begin{tabular}[c]{@{}c@{}}InfRec R500\\\end{tabular}}} &
      \makecell{\rotatebox[]{90}{8* HE}     } & 
      $\vcenter{{(\emph{pub})} Contains RGB-Thermal combined images of autonomous vehicles with pixel-level annotations.}$  & 
      \makecell{\rotatebox[]{90}{Industrial} } \\
     \midrule

   \makecell{\rotatebox[]{90}{{MSSpoof}} \rotatebox[]{90}{\cite{msspoof}}}  & 
      $\vcenter{ \includegraphics[width=1.0\linewidth]{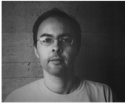}}$  & 
      {face}  & 
      \makecell{\rotatebox[]{90}{4.5K}   } & 
      \makecell{\rotatebox[]{90}{1280x1024} } & 
      \makecell{\rotatebox[]{90}{\begin{tabular}[c]{@{}c@{}}uEye camera\end{tabular}}} &
      \makecell{\rotatebox[]{90}{8 HE}     } & 
      $\vcenter{{(\emph{rr})} The dataset contains visible and NIR face images of 21 subjects, designed for spoofing attacks. It provides manual face annotations with 16 key point coordinates.}$  & 
      \makecell{\rotatebox[]{90}{Mil. \& Sur. }} \\
     \midrule

     \makecell{\rotatebox[]{90}{{MS-OD}} \rotatebox[]{90}{\cite{ms-ob}}}  & 
      $\vcenter{ \includegraphics[width=1.0\linewidth]{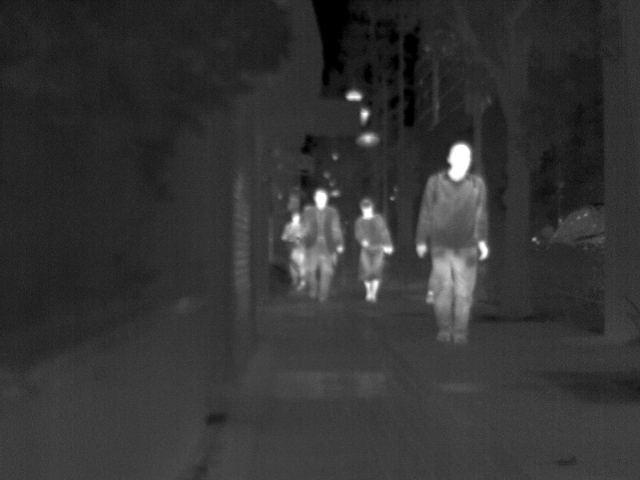}}$  & 
      $\vcenter{bike, car, person}$  & 
      \makecell{\rotatebox[]{90}{30K}   } & 
      \rotatebox[]{90}{\begin{tabular}[c]{@{}c@{}} 320×256\\640×480\end{tabular}} & 
      \makecell{\rotatebox[]{90}{\begin{tabular}[c]{@{}c@{}} InfReC R500\\ InfReC H8000\\Xeva-1.7-320\end{tabular}}} &
      \makecell{\rotatebox[]{90}{8 HE}     } & 
      $\vcenter{{(\emph{pub})} The Multi-Spectral Object Detection dataset contains RGB, NIR, MIR and FIR images for automatic mobile robot tasks in traffic. The authors also present the ground truth labels. There are 7512 images per domain.}$  & 
      \makecell{\rotatebox[]{90}{Industrial} } \\
     \midrule
   
    \makecell{\rotatebox[]{90}{Multi-Focus} \rotatebox[]{90}{\cite{benes2013multi}} } & 
      $\vcenter{ \includegraphics[width=1.0\linewidth]{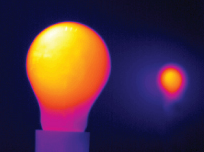}}$  & 
      {uncate-gorised}  & 
     \makecell{\rotatebox[]{90}{576}   } & 
      \makecell{\rotatebox[]{90}{320×240}} & 
      \makecell{\rotatebox[]{90}{\begin{tabular}[c]{@{}c@{}}TESTO \\882-3\end{tabular}}} &
      \makecell{\rotatebox[]{90}{8 HE}      } & 
      $\vcenter{{(\emph{rr})} The set consists of six image sets acquired at different lens positions, by manually moving the lens in 1 mm steps. Each set consists of 96 different images of a single scene.}$  & 
     \makecell{\rotatebox[]{90}{Industrial} } \\
     \midrule
     
     \makecell{\rotatebox[]{90}{ND X1} \rotatebox[]{90}{\cite{nd-x1}}}  & 
      $\vcenter{\includegraphics[width=1.0\linewidth]{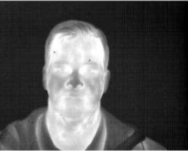}}$  & 
       {face}  & 
      \makecell{\rotatebox[]{90}{4584}   } & 
      \makecell{\rotatebox[]{90}{not specified} } & 
      \makecell{\rotatebox[]{90}{\begin{tabular}[c]{@{}c@{}c@{}}Merlin\\uncooled\\µ-bol.\end{tabular}}} &
      \makecell{\rotatebox[]{90}{8 HE}     } & 
      $\vcenter{{(\emph{rr})} The set includes 2292 IR and  2292 visible face images from 82 subjects.}$  & 
      \makecell{\rotatebox[]{90}{Mil. \& Sur.}  } \\
     \midrule
     
     \makecell{\rotatebox[]{90}{ND-NIVL} \rotatebox[]{90}{\cite{nd-nivl}}}  & 
      $\vcenter{\includegraphics[width=1.0\linewidth]{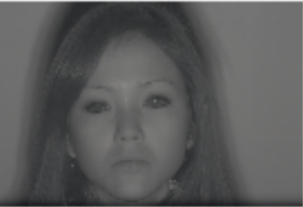}}$  & 
       {face}  & 
      \makecell{\rotatebox[]{90}{25K}   } &
      \rotatebox[]{90}{\begin{tabular}[c]{@{}c@{}}4770x3177\\4288x2848\end{tabular}} &
      \makecell{\rotatebox[]{90}{\begin{tabular}[c]{@{}c@{}}Nikon  D90\\Canon  EOS 50D\end{tabular}}} &
      \makecell{\rotatebox[]{90}{8 HE}     } & 
      $\vcenter{{(\emph{rr})} The dataset includes RGB face images obtained from 574 subjects and NIR face images obtained from 230 subjects. There  are 2341 frames in the visible domain and 22264  frames in the IR domain.}$  & 
      \makecell{\rotatebox[]{90}{Mil. \& Sur.}  } \\
     \midrule

     \makecell{\rotatebox[]{90}{{NOAA}} \rotatebox[]{90}{\cite{noaa}}}  & 
      $\vcenter{\includegraphics[width=1.0\linewidth]{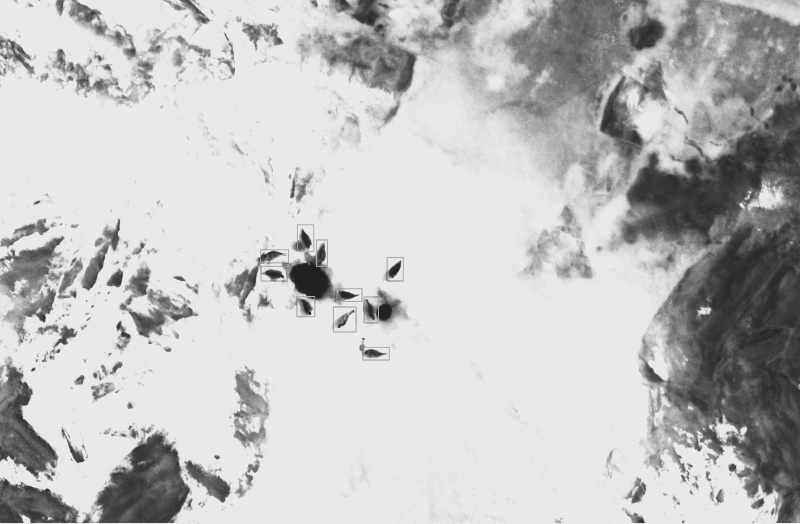}}$  & 
       {animal}  & 
      \makecell{\rotatebox[]{90}{80K}   } &
      \makecell{\rotatebox[]{90}{640x512} } &
      \makecell{\rotatebox[]{90}{\begin{tabular}[c]{@{}c@{}}FLIR A6751 \\SLS\end{tabular}}} &
      \makecell{\rotatebox[]{90}{14 RAW}     } & 
      $\vcenter{{(\emph{pub})} Contains visible and IR aerial images with annotations. There are additional images with no objects/animals. It can be used for detecting animals on sea ice. }$  & 
      \makecell{\rotatebox[]{90}{Scientific }} \\
     \midrule

    \end{tabular}%
  \label{tab:dataset13}%
\end{table*}%

\begin{table*}[p]
  \centering
  \footnotesize
    \begin{tabular}{p{30pt}|p{43pt}|p{38pt}|p{16pt}|p{15pt}|p{26pt}|p{17pt}|p{135pt}|p{15pt}}
    \toprule
    Table \ref{datasetList}   &  continued... &
    \makecell{} & 
    \makecell{} & 
    & 
    \makecell{} & 
    \makecell{} & 
    \makecell{} & 
    \makecell{} \\   
   \midrule 

   \makecell{\rotatebox[]{90}{NTU  RGB+D} \rotatebox[]{90}{\cite{shahroudy2016ntu}}}  & 
      $\vcenter{ \includegraphics[width=1.0\linewidth]{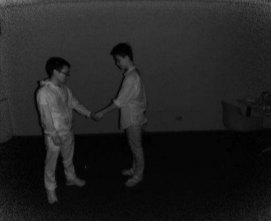}}$  & 
      {120  human  actions}  & 
      \makecell{\rotatebox[]{90}{114K}   } & 
      \makecell{\rotatebox[]{90}{512x424} } & 
      \makecell{\rotatebox[]{90}{Kinect  V2} } & 
      \makecell{\rotatebox[]{90}{8 HE} } & 
      $\vcenter{{(\emph{rr})} The dataset contains RGB videos, depth  map sequences, 3D skeletal data, and  infrared (IR) videos of 82 daily actions, 12 medical  conditions, and 26 mutual  actions.}$  & 
     \makecell{\rotatebox[]{90}{Scientific} } \\
     \midrule

     \makecell{\rotatebox[]{90}{OSU-CTD} \rotatebox[]{90}{\cite{davis2007background}}}  & 
      $\vcenter{ \includegraphics[width=1.0\linewidth]{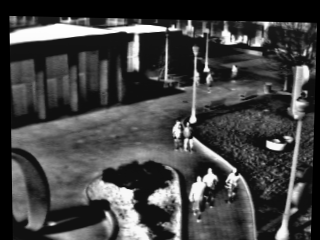}}$  & 
      {vehicle, pedestrian}   & 
      \makecell{\rotatebox[]{90}{17089} } & 
      \makecell{\rotatebox[]{90}{320x240} } & 
      \makecell{\rotatebox[]{90}{\begin{tabular}[c]{@{}c@{}c@{}}Raytheon\\PalmIR\\250D\end{tabular}}} &
      \makecell{\rotatebox[]{90}{8* HE} } & 
      $\vcenter{{(\emph{pub})} The OSU Color-Thermal Database (CDT) includes six colour/thermal sequences, collected at three locations, used for fusion and fusion-based object detection in colour or thermal imagery.} $ & 
      \makecell{\rotatebox[]{90}{Mil. \& Sur.}  } \\
     \midrule
   
  \makecell{\rotatebox[]{90}{OSU TPD} \rotatebox[]{90}{\cite{davis2005two}}} & 
      $\vcenter{ \includegraphics[width=1.0\linewidth]{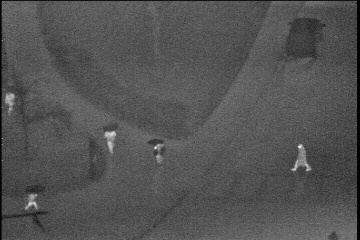}}$  & 
      {pedestrian}  & 
      \makecell{\rotatebox[]{90}{284}   } & 
      \makecell{\rotatebox[]{90}{360x240} } & 
      \makecell{\rotatebox[]{90}{\begin{tabular}[c]{@{}c@{}}Raytheon\\300D\end{tabular}}} &
      \makecell{\rotatebox[]{90}{8 HE}     } & 
      $\vcenter{{(\emph{pub})} The OSU Thermal Pedestrian Database (TPD) includes 10 sequences of videos with annotations of pedestrians who are 50\% visible in images.}$  & 
      \makecell{\rotatebox[]{90}{Mil. \& Sur.} } \\
     \midrule
   
    \makecell{\rotatebox[]{90}{OTCBVS-P} \rotatebox[]{90}{\cite{bilodeau2014thermal}}} & 
      $\vcenter{ \includegraphics[width=1.0\linewidth]{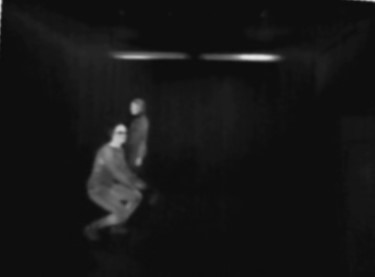}}$  & 
      {pedestrian}  & 
      \makecell{\rotatebox[]{90}{5390} } & 
      \makecell{\rotatebox[]{90}{480x360} } & 
      \makecell{\rotatebox[]{90}{\begin{tabular}[c]{@{}c@{}}FLIR\\A40M\end{tabular}}} &
      \makecell{\rotatebox[]{90}{8* HE}    } & 
      $\vcenter{{(\emph{pub})} Pedestrian Infrared/visible Stereo Video dataset contains 4 visible and infrared video sequences of several people walking with 206 annotated frames and 25819 ground-truth point pairs.}$  & 
       \makecell{\rotatebox[]{90}{Mil. \& Sur.} } \\
     \midrule

    \makecell{\rotatebox[]{90}{{Overhead}} \rotatebox[]{90}{\cite{overhead}}} & 
      $\vcenter{ \includegraphics[width=1.0\linewidth]{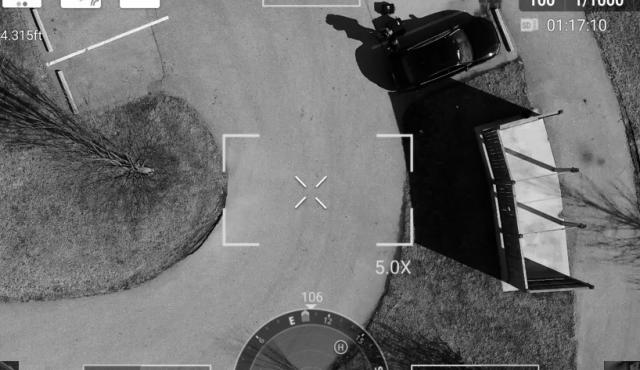}}$  & 
      {pedestrian}  & 
      \makecell{\rotatebox[]{90}{1.9K} } & 
      \makecell{\rotatebox[]{90}{640x370} } & 
      \makecell{\rotatebox[]{90}{\begin{tabular}[c]{@{}c@{}}FLIR Vue \\TZ20\end{tabular}}} &
      \makecell{\rotatebox[]{90}{8* HE}    } & 
      $\vcenter{{(\emph{pub})} The dataset presents thermal images from a UAV perspective, with object annotations. The images do not always include objects. }$  & 
       \makecell{\rotatebox[]{90}{Mil. \& Sur.} } \\
     \midrule

   \makecell{\rotatebox[]{90}{Parma} \rotatebox[]{90}{\cite{parmaTetra}}}  & 
      $\vcenter{ \includegraphics[width=1.0\linewidth]{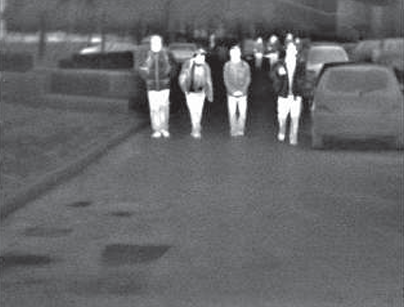}}$  & 
      {pedestrian}   & 
      \makecell{\rotatebox[]{90}{18K} } & 
      \makecell{\rotatebox[]{90}{320x240} } & 
      \makecell{\rotatebox[]{90}{\begin{tabular}[c]{@{}c@{}}not\\specified\end{tabular}}} &
      \makecell{\rotatebox[]{90}{\begin{tabular}[c]{@{}c@{}}not\\specified\end{tabular}}} &
      $\vcenter{{(\emph{pri})} Parma Tetravision dataset, includes visible and infrared images  captured from a car, with pedestrian  annotations.}$  & 
      \makecell{\rotatebox[]{90}{Mil. \& Sur.}  } \\
     \midrule

    \makecell{\rotatebox[]{90}{PETS 2005} \rotatebox[]{90}{\cite{pets2005}}}  & 
      $\vcenter{ \includegraphics[width=1.0\linewidth]{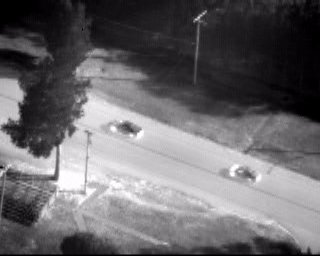}}$  & 
      {vehicle}  & 
      \makecell{\rotatebox[]{90}{5K}    } & 
      \makecell{\rotatebox[]{90}{320x256} } & 
      \makecell{\rotatebox[]{90}{\begin{tabular}[c]{@{}c@{}}not\\specified\end{tabular}}} &
      \makecell{\rotatebox[]{90}{8* HE}    } & 
      $\vcenter{{(\emph{pub})} This an aerial vehicle tracking dataset containing  6 visible and 3  thermal infrared video sequences with annotations.} $ & 
       \makecell{\rotatebox[]{90}{Mil. \& Sur.}  } \\
     \midrule

    \makecell{\rotatebox[]{90}{PTB-TIR  } \rotatebox[]{90}{\cite{PTB-TIR}}}  & 
      $\vcenter{ \includegraphics[width=1.0\linewidth]{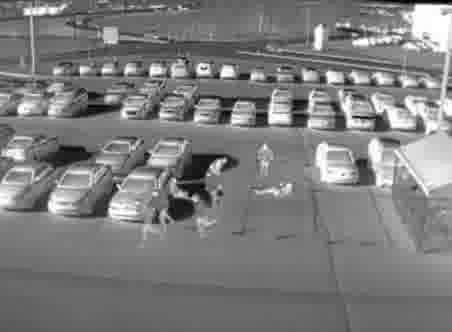}}$  & 
      {vehicle,  pedestrian}  & 
      \makecell{\rotatebox[]{90}{30128} } & 
      \makecell{\rotatebox[]{90}{up to  1280×720} } & 
      \makecell{\rotatebox[]{90}{8 dif. cams.} } & 
      \makecell{\rotatebox[]{90}{8 HE}     } & 
      $\vcenter{{(\emph{pub})} TIR pedestrian tracking dataset include  manually annotated 60 thermal sequences, divided into nine subsets with  different shooting properties. The images are captured indoor and outdoor environments with surveillance, hand-held, vehicle-mounted and drone cameras during day and night.}$  & 
      \makecell{\rotatebox[]{90}{Mil. \& Sur.}  } \\
     \midrule

    \makecell{\rotatebox[]{90}{RGB-NIR} \rotatebox[]{90}{\cite{RGB_NIR}}} & 
       $\vcenter{ \includegraphics[width=1.0\linewidth]{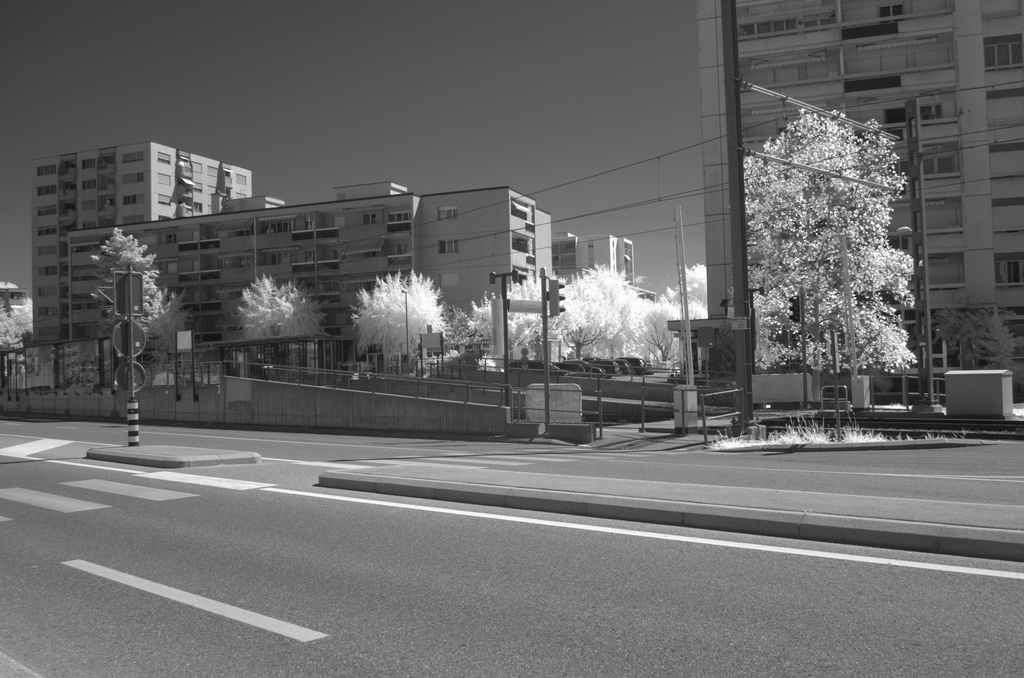}}$  & 
      $\vcenter{building, mountain, tree,  car, motorcy- cle, etc.}$  & 
      \makecell{\rotatebox[]{90}{477}  } & 
      \makecell{\rotatebox[]{90}{1024x768} } & 
      \makecell{\rotatebox[]{90}{Canon T1i} } & 
      \makecell{\rotatebox[]{90}{8* HE}  } & 
      $\vcenter{{(\emph{pub})} The dataset contains RGB and NIR images  of 9 categories: country, field, forest, indoor,  mountain, old buildings, street, urban, and water.  Some frames include objects like buildings,  trees, cars and motorcycles.}$  & 
     \makecell{\rotatebox[]{90}{Scientific} } \\
     \midrule
     
     \makecell{\rotatebox[]{90}{RGBT Salient} \rotatebox[]{90}{\cite{tu2020rgbt}}} & 
      $\vcenter{ \includegraphics[width=1.0\linewidth]{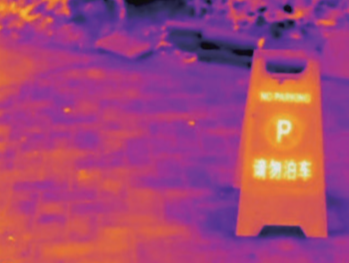}}$  & 
      {various  objects}  & 
      \makecell{\rotatebox[]{90}{5000}  } & 
      \makecell{\rotatebox[]{90}{640x480} } & 
      \makecell{\rotatebox[]{90}{\begin{tabular}[c]{@{}c@{}}FLIR T640\\T610\end{tabular}}} &
      \makecell{\rotatebox[]{90}{8 HE}     } & 
      $\vcenter{{(\emph{pub})} This dataset contains 5000 thermal/RGB image pairs with ground truth annotations of various objects like sneakers,  pontoon, stool, dustbin, headphones for salient object detection tasks.} $ & 
     \makecell{\rotatebox[]{90}{Scientific} } \\
     \midrule
     
     \makecell{\rotatebox[]{90}{RIFIR} \rotatebox[]{90}{\cite{RIFIR}}} & 
      $\vcenter{ \includegraphics[width=1.0\linewidth]{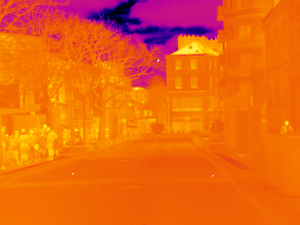}}$   & 
      {pedestrian}  & 
      \makecell{\rotatebox[]{90}{20K}   } & 
      \makecell{\rotatebox[]{90}{640X480} } & 
      \makecell{\rotatebox[]{90}{\begin{tabular}[c]{@{}c@{}}not\\specified\end{tabular}}} &
      \makecell{\rotatebox[]{90}{8* HE}} & 
      $\vcenter{{(\emph{pub})} The dataset contains sequences of images, captured in an urban environment with one FIR and two colour cameras. There are train and test sets, with annotated pedestrians in both visible and IR domains.}$ & 
      \makecell{\rotatebox[]{90}{Mil. \& Sur.} } \\
     \midrule

    \end{tabular}%
  \label{tab:dataset14}%
\end{table*}%

\begin{table*}[p]
  \centering
  \footnotesize
    \begin{tabular}{p{30pt}|p{43pt}|p{38pt}|p{16pt}|p{15pt}|p{26pt}|p{17pt}|p{135pt}|p{15pt}}
    \toprule
    Table \ref{datasetList}   &  continued... & 
    \makecell{} & 
    \makecell{} & 
    & 
    \makecell{} & 
    \makecell{} & 
    \makecell{} & 
    \makecell{} \\
   \midrule

   \makecell{\rotatebox[]{90}{{Roboflow-P}} \rotatebox[]{90}{\cite{roboflowpeople} }}  & 
      $\vcenter{ \includegraphics[width=1.0\linewidth]{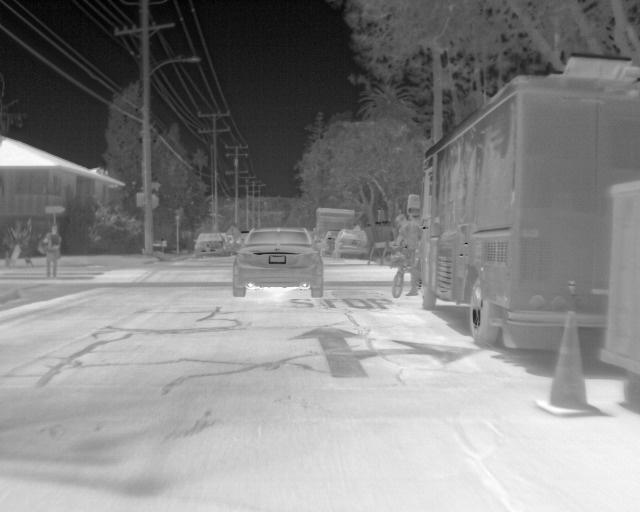}}$  & 
      {person} & 
      \makecell{\rotatebox[]{90}{13K} } & 
      \makecell{\rotatebox[]{90}{640x512} } & 
      \makecell{\rotatebox[]{90}{not specified} } & 
      \makecell{\rotatebox[]{90}{8* HE} } & 
      $\vcenter{{(\emph{pub})} The dataset contains IR images of people with annotations. Augmented images are also available.} $ & 
     \makecell{\rotatebox[]{90}{Mil. \& Sur.}  } \\
     \midrule
     
     \makecell{\rotatebox[]{90}{SAGEM} \rotatebox[]{90}{\cite{robin} }}  & 
      $\vcenter{ \includegraphics[width=1.0\linewidth]{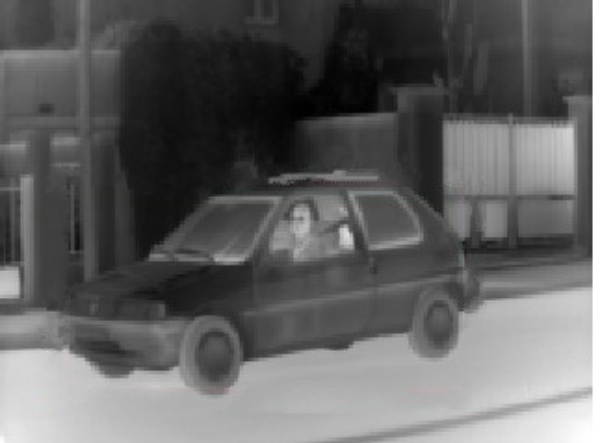}}$  & 
      $\vcenter{vehicle, etc.} $ & 
      \makecell{\rotatebox[]{90}{1400} } & 
      \makecell{\rotatebox[]{90}{384x256} } & 
      \makecell{\rotatebox[]{90}{MWIR  Matis} } & 
      \makecell{\rotatebox[]{90}{16 RAW} } & 
      $\vcenter{{(\emph{rr})} The dataset contains 16bits raw IR aerial images of vehicles for classification and  detection tasks.} $ & 
     \makecell{\rotatebox[]{90}{Mil. \& Sur.}  } \\
     \midrule

       \makecell{\rotatebox[]{90}{SCUT-FIR} \rotatebox[]{90}{\cite{xu2019}}} & 
      $\vcenter{ \includegraphics[width=1.0\linewidth]{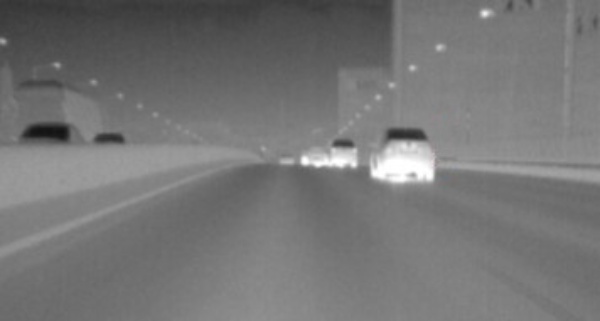}}$  & 
      {pedestrian}  & 
      \makecell{\rotatebox[]{90}{211K}  } & 
      \makecell{\rotatebox[]{90}{720×576} } & 
      \makecell{\rotatebox[]{90}{NV628} } & 
      \makecell{\rotatebox[]{90}{8 HE}   } & 
      $\vcenter{{(\emph{pub})} The dataset consists of approximately  11-hour-long image sequences captured  in various traffic scenarios. 11 road segments are captured in different environments, such as the city centre,  suburbs, highway, and campus.} $ & 
      \makecell{\rotatebox[]{90}{Mil. \& Sur.}  } \\
     \midrule
     
     \makecell{\rotatebox[]{90}{SDT}} \rotatebox[]{90}{\cite{Synthetic}} & 
      $\vcenter{ \includegraphics[width=1.0\linewidth]{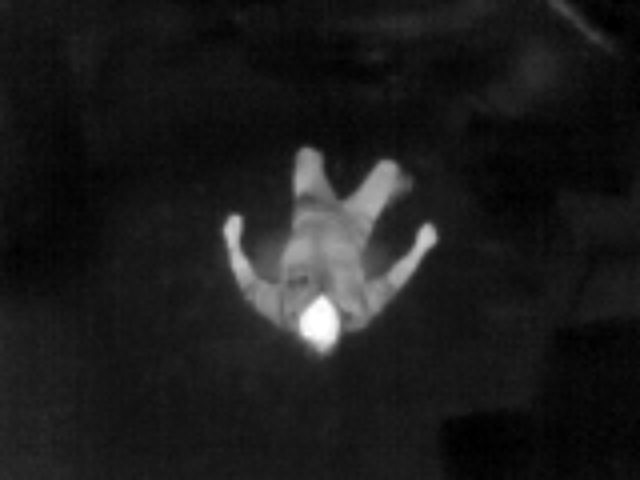}}$  &
      {person}  & 
      \makecell{\rotatebox[]{90}{48K}   } & 
      \makecell{\rotatebox[]{90}{640x480} } & 
      {\rotatebox[]{90}{\makecell{FLIR\\Lepton\\ 3.5}}} &
      \makecell{\rotatebox[]{90}{16 RAW}   } & 
      $\vcenter{{(\emph{pub})} The dataset includes 40k synthetic and 8k real depth and thermal stereo images for human behaviour recognition and person detection tasks, with bounding box annotations.} $ & 
      \makecell{\rotatebox[]{90}{Mil. \& Sur.}   } \\
     \midrule
   
     \makecell{\rotatebox[]{90}{SENSIAC} \rotatebox[]{90}{\cite{SENSIAC}}} & 
      $\vcenter{ \includegraphics[width=1.0\linewidth]{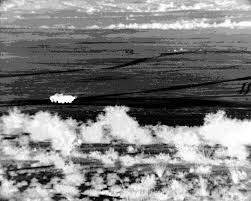}}$  & 
      $\vcenter{people, vehicles, etc} $ & 
      {\rotatebox[]{90}{\makecell{not publicly\\specified}}} &
      {\rotatebox[]{90}{\makecell{not publicly\\specified}}} &
      {\rotatebox[]{90}{\makecell{not publicly\\specified}}}
       &
      {\rotatebox[]{90}{\makecell{not publicly\\specified}}} &
      $\vcenter{{(\emph{pri})} The dataset contains 207 GB of videos in  the IR domain and 106 GB of videos in the  visible domain with ground truth data for automatic target recognition (ATR) tasks. Collected by the US Army Night Vision and  Electronic Sensors Directorate (NVESD).}$  & 
      \makecell{\rotatebox[]{90}{Mil. \& Sur.}   } \\
     \midrule

    \makecell{\rotatebox[]{90}{Server} \rotatebox[]{90}{\cite{liu2019research}}} & 
      $\vcenter{ \includegraphics[width=1.0\linewidth]{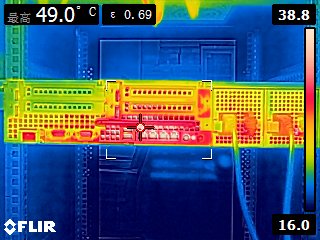}}$  & 
      {server hardware}  & 
      \makecell{\rotatebox[]{90}{1351}  } & 
      \makecell{\rotatebox[]{90}{320x240} } & 
      \makecell{\rotatebox[]{90}{\begin{tabular}[c]{@{}c@{}}FLIR\\E8\end{tabular}}} &
      \makecell{\rotatebox[]{90}{8* HE}    } & 
      $\vcenter{{(\emph{pub})} The dataset contains thermal images of servers for detecting the overheated area of the server surface, with five categories: normal status,  main fan failure, vice-fan failure,  air vent blockage and low-load status.}$  & 
     \makecell{\rotatebox[]{90}{Industrial} } \\
     \midrule
    
     \makecell{\rotatebox[]{90}{SG-Ship} \rotatebox[]{90}{\cite{7812788}}}  & 
      $\vcenter{ \includegraphics[width=1.0\linewidth]{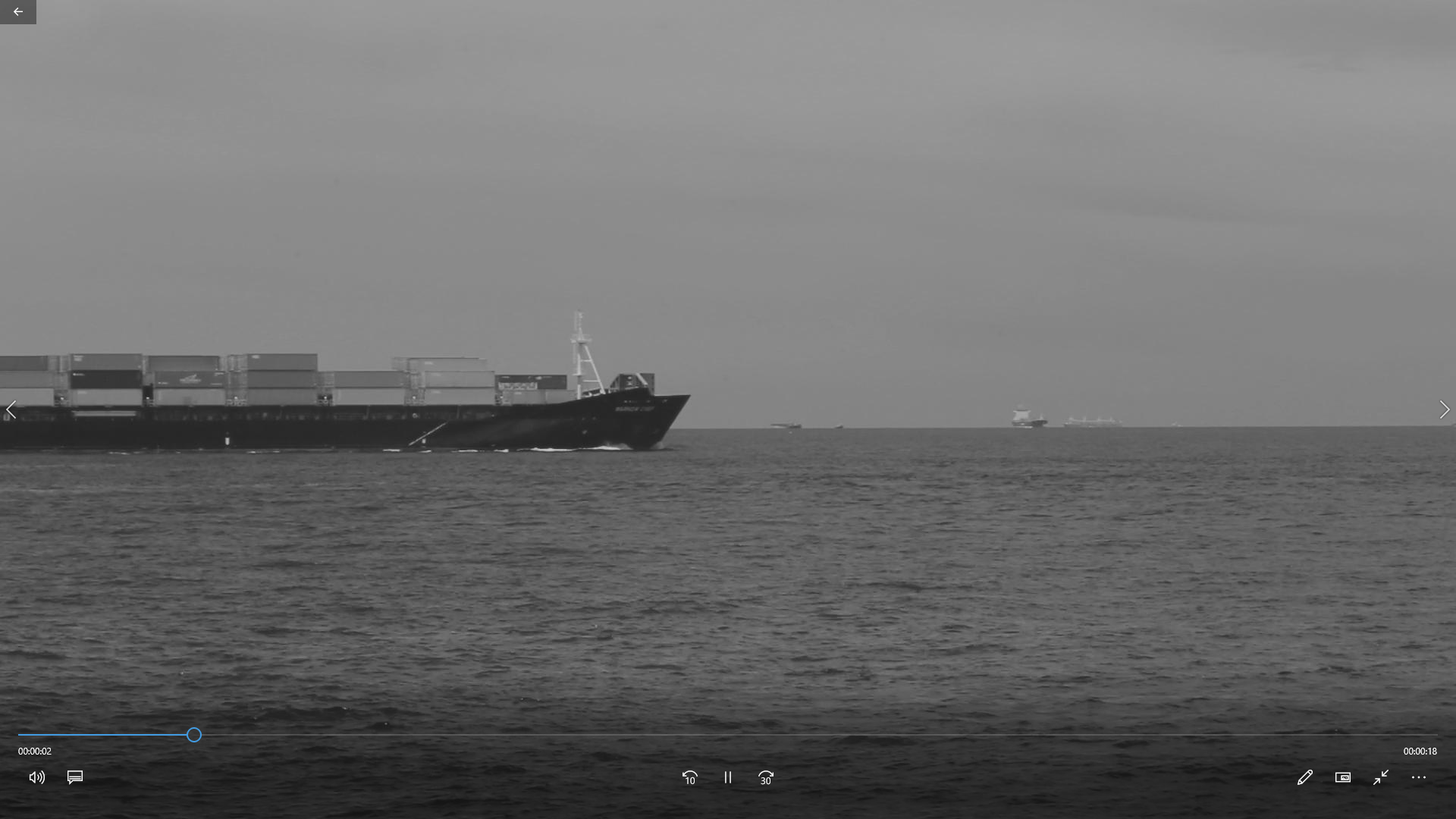}}$  & 
      {ship}   & 
      \makecell{\rotatebox[]{90}{24K} } &
      \makecell{\rotatebox[]{90}{1080x1920} } & 
      \makecell{\rotatebox[]{90}{Canon  70D}  } & 
      \makecell{\rotatebox[]{90}{8* HE}    } & 
      $\vcenter{{(\emph{pub})} Singapore Maritime dataset The ship videos, are captured  at various locations around Singapore  waters. The set is divided into three parts,  40 videos for visible on-shore, 11 videos for visible on-board and 30 videos for NIR on-shore shots. The dataset provides annotations in .mat format in 3 folders: horizon, object and track.}$  & 
      \makecell{\rotatebox[]{90}{Mil. \& Sur.}  } \\
     \midrule
     
      \makecell{\rotatebox[]{90}{Soccer} \rotatebox[]{90}{\cite{gade2018constrained}}} & 
      $\vcenter{ \includegraphics[width=1.0\linewidth]{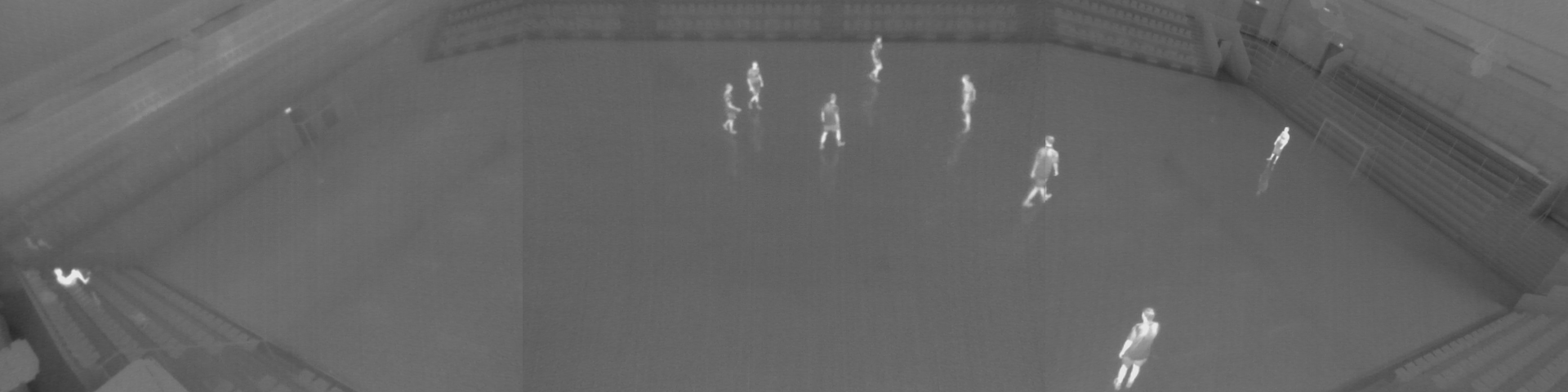}}$  & 
      {people}  & 
      \makecell{\rotatebox[]{90}{3000}  } & 
      \makecell{\rotatebox[]{90}{1920x480} } & 
      \makecell{\rotatebox[]{90}{\begin{tabular}[c]{@{}c@{}}AXIS\\Q1922\end{tabular}}} &
      \makecell{\rotatebox[]{90}{8 HE}    } & 
      $\vcenter{{(\emph{pub})} The dataset provides four thermal infrared  video sequences of eight soccer players at  an indoor sports arena.}$  & 
     \makecell{\rotatebox[]{90}{Mil. \& Sur.}  } \\
     \midrule
     
     \makecell{\rotatebox[]{90}{Spindle} \rotatebox[]{90}{\cite{spindle}}}   & 
      $\vcenter{ \includegraphics[width=1.0\linewidth]{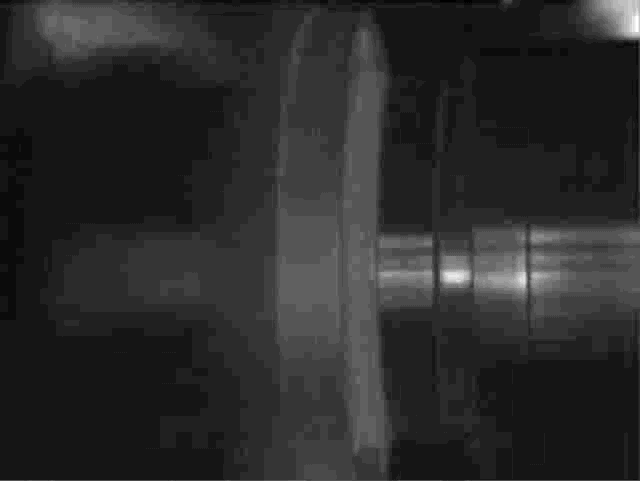}}$  & 
      {lathe milling}  & 
      \makecell{\rotatebox[]{90}{1500 }} & 
      \makecell{\rotatebox[]{90}{640x480} } & 
      \makecell{\rotatebox[]{90}{\begin{tabular}[c]{@{}c@{}}not\\specified\end{tabular}}} &
      \makecell{\rotatebox[]{90}{8* HE}  } & 
      $\vcenter{{(\emph{rr})} Spindle thermal error prediction dataset contains all the thermal images and error data of a spindle, taken on a lathe and a milling machine.}$ & 
     \makecell{\rotatebox[]{90}{Industrial}  } \\
     \midrule
   
   \makecell{\rotatebox[]{90}{Surf-Coat} \rotatebox[]{90}{\cite{coated}}}  & 
      $\vcenter{ \includegraphics[width=1.0\linewidth]{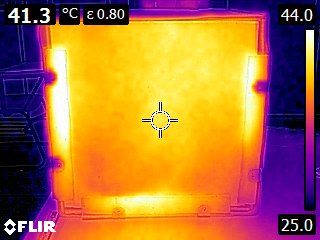}}$  & 
      $\vcenter{various materials}$  & 
      \makecell{\rotatebox[]{90}{449}   } & 
      \makecell{\rotatebox[]{90}{320x240} } & 
      \makecell{\rotatebox[]{90}{FLIR E4} } & 
      \makecell{\rotatebox[]{90}{8* HE}     } & 
      $\vcenter{{(\emph{pub})} This dataset contains raw temperature measurement data of external, and internal surfaces and air in the interior of a structure under different conditions. It has also surface thermographic images of different types of coatings and diffuses reflectance data for these materials.} $ & 
     \makecell{\rotatebox[]{90}{Industrial} } \\
     \midrule

    \end{tabular}%
  \label{tab:dataset14}%
\end{table*}%

\begin{table*}[p]
  \centering
  \footnotesize
    \begin{tabular}{p{30pt}|p{43pt}|p{38pt}|p{16pt}|p{15pt}|p{26pt}|p{17pt}|p{135pt}|p{15pt}}
    \toprule
    Table \ref{datasetList}   &  continued... & 
    \makecell{} & 
    \makecell{} & 
    & 
    \makecell{} & 
    \makecell{} & 
    \makecell{} & 
    \makecell{} \\
   \midrule

   \makecell{\rotatebox[]{90}{Terravic-F} \rotatebox[]{90}{\cite{terravic}}} & 
      $\vcenter{ \includegraphics[width=1.0\linewidth]{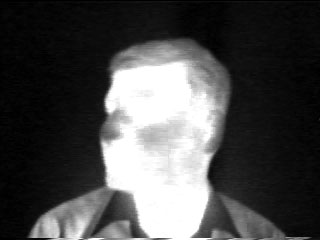}}$  & 
      {face}   & 
      \makecell{\rotatebox[]{90}{23K}   } & 
      \makecell{\rotatebox[]{90}{320x240} } & 
      \makecell{\rotatebox[]{90}{\begin{tabular}[c]{@{}c@{}}Raytheon\\L3\end{tabular}}} &
      \makecell{\rotatebox[]{90}{8 HE}     } & 
      $\vcenter{{(\emph{pub})} The Terravic Facial dataset consists of 20 thermal infrared facial image sequences, captured from the front,  left and right at indoor and outdoor environments.} $ & 
     \makecell{\rotatebox[]{90}{Mil. \& Sur.} } \\
     \midrule
     
     \makecell{\rotatebox[]{90}{Terravic-M} \rotatebox[]{90}{\cite{terravic}}}  & 
      $\vcenter{ \includegraphics[width=1.0\linewidth]{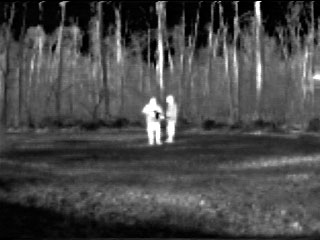}}$  & 
      $\vcenter{human, plane, dock} $ & 
      \makecell{\rotatebox[]{90}{22K} } & 
      \makecell{\rotatebox[]{90}{320x240} } & 
      \makecell{\rotatebox[]{90}{\begin{tabular}[c]{@{}c@{}}Raytheon\\L3\end{tabular}}} &
      \makecell{\rotatebox[]{90}{8 HE}} & 
      $\vcenter{{(\emph{pub})} The Terravic Motion dataset has 18 sequences of thermal infrared  images, captured at  indoor and outdoor environments for object detection and tracking tasks.}$ & 
     \makecell{\rotatebox[]{90}{Mil. \& Sur.}  } \\
     \midrule
     
      \makecell{\rotatebox[]{90}{Terravic-W} \rotatebox[]{90}{\cite{terravic}}} & 
      $\vcenter{ \includegraphics[width=1.0\linewidth]{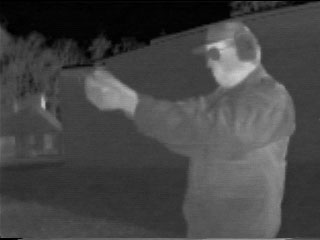}}$  & 
      $\vcenter{person, weapon} $ & 
      \makecell{\rotatebox[]{90}{3555}  } & 
      \makecell{\rotatebox[]{90}{320x240} } & 
      \makecell{\rotatebox[]{90}{\begin{tabular}[c]{@{}c@{}}Raytheon\\L3\end{tabular}}} &
      \makecell{\rotatebox[]{90}{8 HE} } & 
      $\vcenter{{(\emph{pub})} The dataset includes 5 thermal IR image sequences for weapon detection and weapon evacuation detection tasks.} $ & 
     \makecell{\rotatebox[]{90}{Mil. \& Sur.}  } \\
     \midrule
   
   \makecell{\rotatebox[]{90}{TestisT} \rotatebox[]{90}{\cite{testis}} } & 
      $\vcenter{ \includegraphics[width=1.0\linewidth]{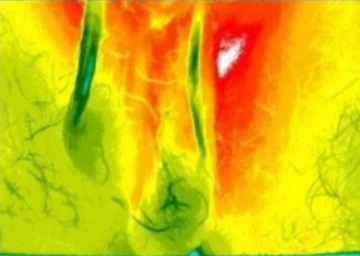}}$  & 
      {testis}  & 
      \makecell{\rotatebox[]{90}{50} } & 
      \makecell{\rotatebox[]{90}{640x320} } & 
      \makecell{\rotatebox[]{90}{VIS-IR  640} } & 
      \makecell{\rotatebox[]{90}{8 HE} } & 
      $\vcenter{{(\emph{pub})} The Testis Thermography dataset contains 50 thermal testis images to detect varicocele.} $& 
     \makecell{\rotatebox[]{90}{Medical}  } \\
     \midrule
     
     \makecell{\rotatebox[]{90}{THALES} \rotatebox[]{90}{\cite{robin}}}  & 
      $\vcenter{ \includegraphics[width=1.0\linewidth]{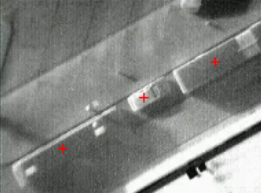}}$  & 
      {vehicle, boat}  & 
      \makecell{\rotatebox[]{90}{more  than  6K} } & 
      {\rotatebox[]{90}{\makecell{320x240\\640x512}}} &
      \makecell{\rotatebox[]{90}{\begin{tabular}[c]{@{}c@{}}not\\specified\end{tabular}}} &
      \makecell{\rotatebox[]{90}{\begin{tabular}[c]{@{}c@{}}8/16\\HE/RAW\end{tabular}}} &
      $\vcenter{{(\emph{rr})} Contains aerial images obtained from six hours of video recorded from a helicopter in different environments, such as urban, expressway, rural, etc.}$ & 
     \makecell{\rotatebox[]{90}{Mil. \& Sur.}  } \\
     \midrule
     
     \makecell{\rotatebox[]{90}{The Flame} \rotatebox[]{90}{\cite{flame}}} & 
      $\vcenter{ \includegraphics[width=1.0\linewidth]{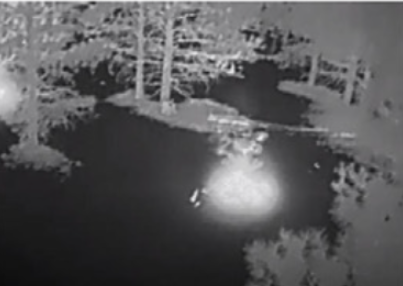}}$  & 
      {fire}  & 
      \makecell{\rotatebox[]{90}{49K } } &
      {\rotatebox[]{90}{\begin{tabular}[c]{@{}c@{}}640x512\\3840x2160\end{tabular}}} &
      \makecell{\rotatebox[]{90}{\begin{tabular}[c]{@{}c@{}}FLIR  Vue  Pro R\\DJI Phantom 3\end{tabular}}} &
      \makecell{\rotatebox[]{90}{8* HE}    } & 
      $\vcenter{{(\emph{rr})} The dataset consists of raw aerial video  and raw heat map footage captured by drones, during a pile burn in Northern  Arizona. Collected for wildfire detection and includes 47992 labelled frames for fire classification and  2003 ground truth masked frames for fire segmentation.}$  & 
     \makecell{\rotatebox[]{90}{Mil. \& Sur.}} \\
     \midrule

   \makecell{\rotatebox[]{90}{TIDOC} \rotatebox[]{90}{\cite{catcarman}}}  & 
      $\vcenter{ \includegraphics[width=0.7\linewidth]{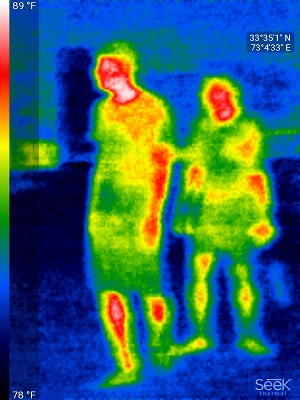}}$  & 
      {car, cat, pedestrian}  & 
      \makecell{\rotatebox[]{90}{6892} } & 
      {\rotatebox[]{90}{\begin{tabular}[c]{@{}c@{}}300x400\\1080x1440\end{tabular}}} &
      \makecell{\rotatebox[]{90}{\begin{tabular}[c]{@{}c@{}}FLIR\\Seek Thermal\end{tabular}}} &
      \makecell{\rotatebox[]{90}{8* HE} } & 
      $\vcenter{{(\emph{pub})} The Thermal Image dataset for Object Classification (TIDOC) includes thermal images of three classes, car, cat, and pedestrian.} $ & 
     \makecell{\rotatebox[]{90}{Mil. \& Sur.}   } \\
     \midrule

       \makecell{\rotatebox[]{90}{{TIMo}} \rotatebox[]{90}{\cite{timo}}}  & 
      $\vcenter{ \includegraphics[width=0.7\linewidth]{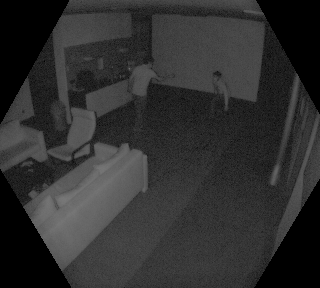}}$  & 
      {person}  & 
      \makecell{\rotatebox[]{90}{635.6K} } & 
      {\rotatebox[]{90}{\begin{tabular}[c]{@{}c@{}}512x512\\288x320\end{tabular}}} &
      \makecell{\rotatebox[]{90}{\begin{tabular}[c]{@{}c@{}}Microsoft Azure\\ Kinect RGB-D\end{tabular}}} &
      \makecell{\rotatebox[]{90}{16 RAW} } & 
      $\vcenter{{(\emph{rr})} The time-of-Flight Indoor Monitoring dataset presents IR and depth videos. Totally, 612 K frames for anomaly detection and 23.6 K frames for person detection, with  annotations for both tasks. For person detection, bounding boxes and segmentation masks are available.} $ & 
     \makecell{\rotatebox[]{90}{Mil. \& Sur.}   } \\
     \midrule
     
     \makecell{\rotatebox[]{90}{TIRDRD} \rotatebox[]{90}{\cite{yoon2016thermal}} } & 
      $\vcenter{ \includegraphics[width=1.0\linewidth]{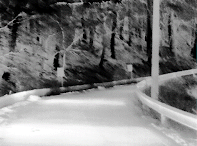}}$  & 
      {road}   & 
      \makecell{\rotatebox[]{90}{6000}  } & 
      \makecell{\rotatebox[]{90}{640x480} } & 
      \makecell{\rotatebox[]{90}{\begin{tabular}[c]{@{}c@{}}FLIR\\A655SC\end{tabular}}} &
      \makecell{\rotatebox[]{90}{8 HE}     } & 
      $\vcenter{{(\emph{pub})} Thermal IR-based Drivable Region Detection dataset consists of about  6000 manually annotated images, with road scenarios, such  as on-road, off-road, and cluttered road for drivable region detection tasks.}$  & 
     \makecell{\rotatebox[]{90}{Mil. \& Sur.} } \\
     \midrule

    \makecell{\rotatebox[]{90}{{TNO}} \rotatebox[]{90}{\cite{TNO}}}  & 
      $\vcenter{ \includegraphics[width=1.0\linewidth]{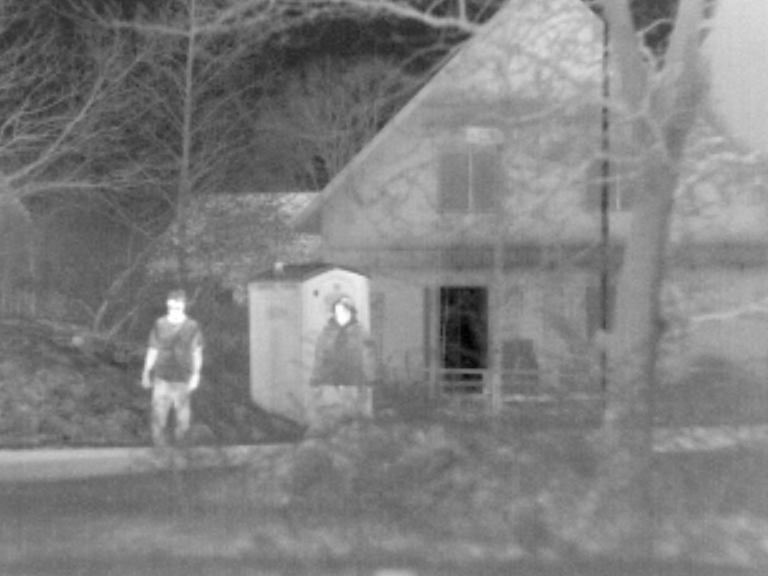}}$  & 
      $\vcenter{objects, person, vehicle} $   & 
      \makecell{\rotatebox[]{90}{400}  } & 
      \makecell{\rotatebox[]{90}{256x256} } & 
      \makecell{\rotatebox[]{90}{\begin{tabular}[c]{@{}c@{}}Amber \\Radiance 1\end{tabular}}} &
      \makecell{\rotatebox[]{90}{8* HE}     } & 
      $\vcenter{{(\emph{pub})} The TNO Image Fusion Dataset consists of visual and IR pairs of images captured at night time. }$  & 
     \makecell{\rotatebox[]{90}{Mil. \& Sur.} } \\
     \midrule
    
      \makecell{\rotatebox[]{90}{Transformer} \rotatebox[]{90}{\cite{Transformer}}} & 
      $\vcenter{ \includegraphics[width=1.0\linewidth]{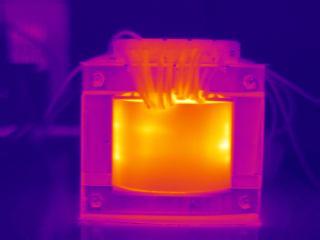}}$  & 
      $\vcenter{transformer, induction motors} $ & 
      \makecell{\rotatebox[]{90}{255}  } & 
      \makecell{\rotatebox[]{90}{320x240} } & 
       \makecell{\rotatebox[]{90}{\begin{tabular}[c]{@{}c@{}}Dali-tech\\T4/T8\end{tabular}}} &
      \makecell{\rotatebox[]{90}{8* HE}    } & 
      $\vcenter{{(\emph{pub})} The dataset contains thermal images of induction motors and transformers for the purpose of condition monitoring.}$  & 
     \makecell{\rotatebox[]{90}{Industrial}   } \\
     \midrule

    \end{tabular}%
  \label{tab:dataset14}%
\end{table*}%

\begin{table*}[h]
  \centering
  \footnotesize
    \begin{tabular}{p{30pt}|p{43pt}|p{38pt}|p{16pt}|p{15pt}|p{26pt}|p{17pt}|p{135pt}|p{15pt}}
    \toprule
    Table \ref{datasetList}   &  continued... & 
    \makecell{} & 
    \makecell{} & 
    & 
    \makecell{} & 
    \makecell{} & 
    \makecell{} & 
    \makecell{} \\
   \midrule

   \makecell{\rotatebox[]{90}{TRICLOBS } \rotatebox[]{90}{\cite{triclobs}}}  & 
      $\vcenter{ \includegraphics[width=1.0\linewidth]{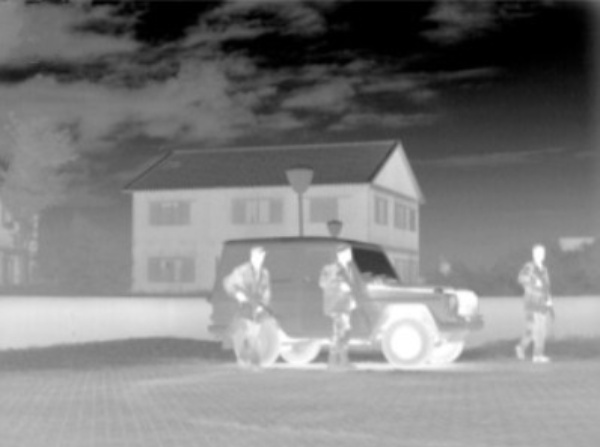}}$  & 
      {people, vehicle}   & 
      \makecell{\rotatebox[]{90}{57K} } & 
      \makecell{\rotatebox[]{90}{640x480} } & 
      {\rotatebox[]{90}{\begin{tabular}[c]{@{}c@{}}XenICs\\Gobi\\ 384\end{tabular}}} &
      \makecell{\rotatebox[]{90}{8 HE} } & 
     $\vcenter{{(\emph{pub})} The TRICLOBS (TRI-band Color Low  Light OBServation) dataset contains  video sequences in visible, NIR  and LWIR bands. The  main purpose of this dataset is to use image fusion and colour-mapping algorithms for surveillance applications. }$ & 
     \makecell{\rotatebox[]{90}{Mil. \& Sur.} } \\
     \midrule

     \makecell{\rotatebox[]{90}{Tufts-Face } \rotatebox[]{90}{\cite{panetta2018comprehensive}}}  & 
      $\vcenter{ \includegraphics[width=1.0\linewidth]{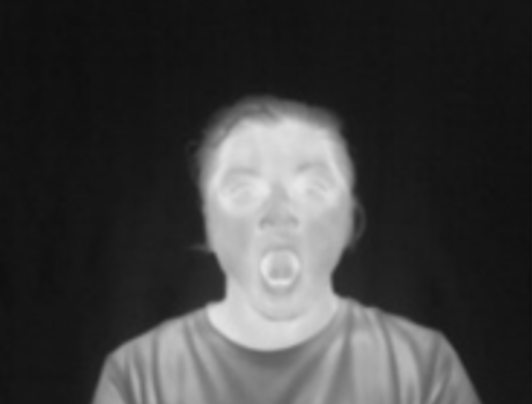}}$ & 
      {face }  & 
      \makecell{\rotatebox[]{90}{100K } } & 
      \makecell{\rotatebox[]{90}{not  specified} } & 
      {\rotatebox[]{90}{\begin{tabular}[c]{@{}c@{}}various\\cameras\end{tabular}}} &
      \makecell{\rotatebox[]{90}{not  specified }} & 
      $\vcenter{{(\emph{rr})} Tufts-Face-Database has more than 100K face images of 7 image modes: visible, NIR, thermal, computerised  sketch, video, plenoptic and 3D images.}$  & 
     \makecell{\rotatebox[]{90}{Mil. \& Sur.}  } \\
     \midrule
     
     \makecell{\rotatebox[]{90}{UL-FMTV} \rotatebox[]{90}{\cite{fmtv}}}  & 
      $\vcenter{ \includegraphics[width=0.7\linewidth]{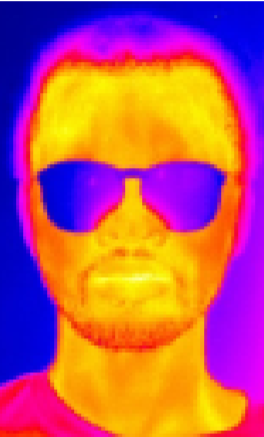}}$  & 
      {face} & 
      \makecell{\rotatebox[]{90}{71400} } & 
      \makecell{\rotatebox[]{90}{640x512} } & 
      {\rotatebox[]{90}{\begin{tabular}[c]{@{}c@{}c@{}}Indigo\\Phoenix\\Thermal\end{tabular}}} &
      \makecell{\rotatebox[]{90}{8 HE} } & 
      $\vcenter{{(\emph{rr})} The ULFMT video dataset includes MWIR band facial videos of 238 subjects for facial pose and  expression recognition applications.} $ & 
     \makecell{\rotatebox[]{90}{Mil. \& Sur.}  } \\
     \midrule
     
     \makecell{\rotatebox[]{90}{UNIRI-TID} \rotatebox[]{90}{\cite{9133581}}} & 
      $\vcenter{ \includegraphics[width=1.0\linewidth]{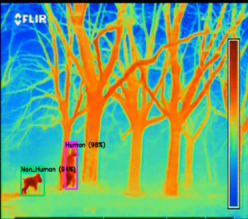}}$  & 
      {person}  & 
      \makecell{\rotatebox[]{90}{11K}   } & 
      \makecell{\rotatebox[]{90}{1280x960} } & 
      {\rotatebox[]{90}{\begin{tabular}[c]{@{}c@{}c@{}}FLIR\\Therma\\-Cam P10\end{tabular}}} &
      \makecell{\rotatebox[]{90}{8 HE}     } & 
      $\vcenter{{(\emph{rr})} This is a collection of thermal videos and images that simulate illegal border crossings and movements in protected areas. The videos were shot in and around the forest at night and in various weather conditions.}$  & 
     \makecell{\rotatebox[]{90}{Mil. \& Sur.}   } \\
     \midrule
     
    \makecell{\rotatebox[]{90}{VAIS  } \rotatebox[]{90}{\cite{7301291}}}  & 
      $\vcenter{ \includegraphics[width=1.0\linewidth]{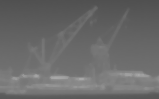}}$  & 
      {ship}   & 
      \makecell{\rotatebox[]{90}{1242}  } & 
      \makecell{\rotatebox[]{90}{1024x680} } & 
      {\rotatebox[]{90}{\begin{tabular}[c]{@{}c@{}c@{}}Sofradir\\ EC Atom \\1024\end{tabular}}} &
      \makecell{\rotatebox[]{90}{8 HE}     } & 
      $\vcenter{{(\emph{pub})} VAIS includes IR and visible image sequences of actual ship images captured from piers, along with annotations.}$ & 
     \makecell{\rotatebox[]{90}{Mil. \& Sur.}   } \\
     \midrule
     
     \makecell{\rotatebox[]{90}{Valle-Aerial} \rotatebox[]{90}{\cite{aerial}}}  & 
      $\vcenter{ \includegraphics[width=1.0\linewidth]{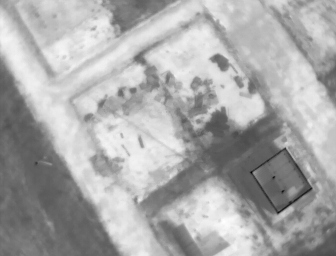}}$  & 
      {road, car}  & 
       \makecell{\rotatebox[]{90}{110} } & 
       \makecell{\rotatebox[]{90}{336x256} } & 
      \makecell{\rotatebox[]{90}{\begin{tabular}[c]{@{}c@{}}Zenmuse\\XT\end{tabular}}} &
       \makecell{\rotatebox[]{90}{8* HE}     } & 
       $\vcenter{{(\emph{pub})} The dataset is made up of thermal and visible aerial images of a planar scene captured using a UAV at Universidad del Valle in Cali, Colombia.}$  & 
     \makecell{\rotatebox[]{90}{Mil. \& Sur.}  } \\
     \midrule
    
    \makecell{\rotatebox[]{90}{VAP} \rotatebox[]{90}{\cite{RGB_depth}}} & 
      $\vcenter{ \includegraphics[width=1.0\linewidth]{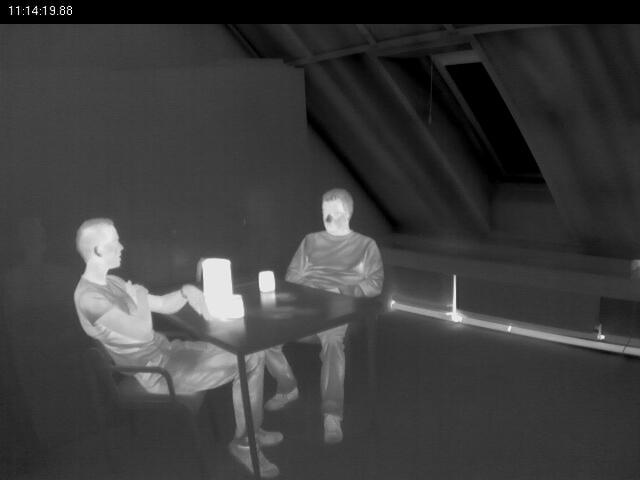}}$  & 
      {people}  & 
      \makecell{\rotatebox[]{90}{11537} } & 
      \makecell{\rotatebox[]{90}{640x480} } & 
      \makecell{\rotatebox[]{90}{\begin{tabular}[c]{@{}c@{}}AXIS\\Q1922\end{tabular}}} &
       \makecell{\rotatebox[]{90}{\begin{tabular}[c]{@{}c@{}}16/32\\HE/RAW\end{tabular}}} &
      $\vcenter{{(\emph{pub})} The dataset contains RGB (32bit), depth (16bit), and thermal (32bit) images of people for human detection, human  segmentation, person re-identiﬁcation  tasks, with 5724 annotated frames.} $ & 
     \makecell{\rotatebox[]{90}{Mil. \& Sur.}  } \\
     \midrule
     
     \makecell{\rotatebox[]{90}{ViViD++} \rotatebox[]{90}{\cite{vivid}}}   & 
      $\vcenter{ \includegraphics[width=1.0\linewidth]{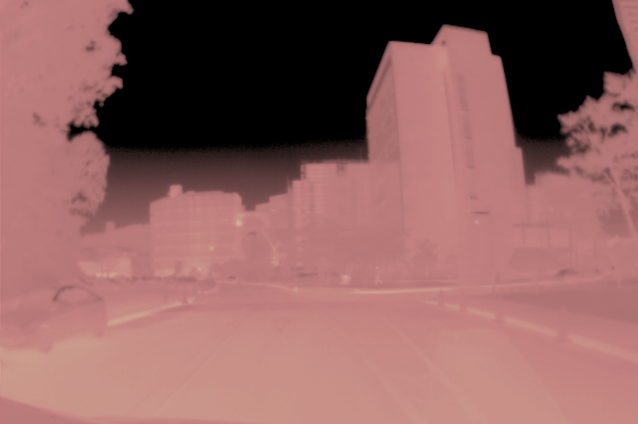}}$  & 
      {uncate-gorised}   & 
      \makecell{\rotatebox[]{90}{9290} } & 
      \makecell{\rotatebox[]{90}{640×480} } & 
      \makecell{\rotatebox[]{90}{FLIR A65} } & 
      \makecell{\rotatebox[]{90}{8 HE}     } & 
      $\vcenter{{(\emph{rr})} The dataset provides normal and poor illumination sequences, captured by thermal,  depth, and temporal difference sensors for indoor and outdoor environments.}$  & 
     \makecell{\rotatebox[]{90}{Scientific} } \\
     \midrule
    
    \makecell{\rotatebox[]{90}{{VOT-RGBTIR}} \rotatebox[]{90}{\cite{vot}}}   & 
      $\vcenter{ \includegraphics[width=1.0\linewidth]{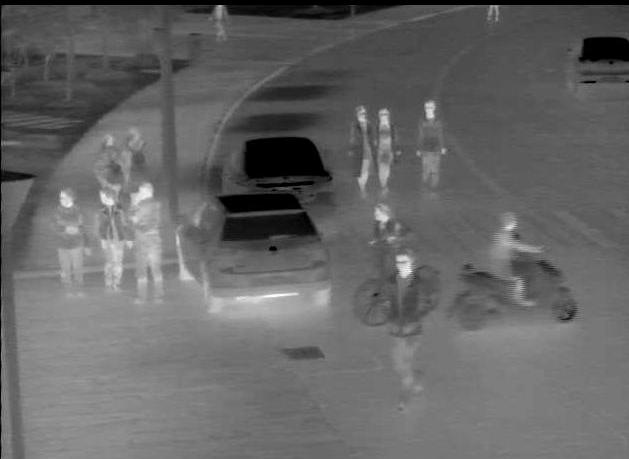}}$  & 
      {pedestrian, bike, car, dog, motorcycle}   & 
      \makecell{\rotatebox[]{90}{20K} } & 
      \makecell{\rotatebox[]{90}{around 600x400} } & 
      \makecell{\rotatebox[]{90}{not specified} } & 
      \makecell{\rotatebox[]{90}{8* HE}     } & 
      $\vcenter{{(\emph{pub})} VOT-RGB TIR 2019 dataset includes infrared images in 60 sequences. Each sequence contains various numbers of images with  objects like cars, pedestrians, motorcycles, etc. The dataset has different object annotations, such as baby, child or cars with colour, etc.}$  & 
     \makecell{\rotatebox[]{90}{Mil. \& Sur.} } \\
     \midrule
     
    \end{tabular}%
  \label{tab:dataset14}%
\end{table*}%
\end{center}


\section{Conclusions and Future Directions}

In this survey, we compile a list of publicly available IR image and video
sets for artificial intelligence and computer vision researchers. We mainly focus on IR image and video sets, which are collected and labelled for computer vision applications such as object detection, object segmentation, classification, and motion detection. We categorize 109 publicly available or private sets according to their sensor types, image resolution, and scale. The list includes brief descriptions for each set. The statistical details of the entire corpus of IR image \& video sets are provided in terms of applications fields, including object categories, resolution, annotations, sensor types and preprocessing details.

We believe that this survey, with solid introductory references to the fundamentals of IR imagery, will be a guideline for computer vision and artificial intelligence researchers who want to delve into working with the spectra beyond the visible domain.  {Today, consumer electronics are integrating IR cameras with smartphones, making IR imaging a reality within the consumer market. Within a short time, the IR domain will host a large number of pre-trained deep learning models. Therefore, this collection can be used to research deep learning models for vision problems like IR domain adaptation, multi-modal vision, and fusion in the future. Such an approach may result in IR subband-specific deep feature extractors, which can be used for a variety of vision tasks. These models would need very large-scale sets.  A crucial practice in the future would be the ongoing updating of this survey, especially in light of the possibility that annotated IR sets may soon be made available in vast quantities.}



\section*{Declarations}
The authors have no conflicts of interest to declare that are relevant to the content of this article.

\section*{Data Availability Statement}
The dataset generated during the current study is available from the corresponding author upon reasonable request.

\Urlmuskip=0mu plus 1mu\relax
\bibliographystyle{sn-basic}
\bibliography{ref.bib}

\section*{Appendix 1}
\subsection{List of Abbreviations}
\begin{tabular}{p{50pt} p{155pt}}

\textbf{CT}  & Computerised Tomography \\
\textbf{CTE} & Coefficient of Thermal Expansion \\
\textbf{D*} & Detectivity \\
\textbf{E} & Emissivity \\
\textbf{ES} & Electromagnetic Spectrum \\
\textbf{FHD} & Full High Definition \\
\textbf{FIR} & Far-Infrared \\
\textbf{FLIR} & Forward Looking Infrared \\
\textbf{FOV} & Field-of-View \\
\textbf{FPA} & Focal Plane Array \\
\textbf{HD} & High Definition \\
\textbf{HE} & Histogram Equalization \\
\textbf{IR}  & Infrared \\
\textbf{LD} & Low Definition \\
\textbf{LWIR}  & Long-Wave Infrared \\
\textbf{Mil.\&Sur.}  & Military \& Surveillance \\
\textbf{MR} & Magnetic Resonance \\
\textbf{MWIR} & Mid-Wave Infrared \\
\textbf{NEP} & Noise-Equivalent-Power \\
\textbf{NIR}  & Near-Infrared \\
\textbf{pri} & Private Dataset \\
\textbf{pub} & Public Dataset \\
\textbf{RGB} & Red-Green-Blue \\
\textbf{rr} & Dataset that Requires Registration \\
\textbf{SAR} & Synthetic Aperture Radar \\
\textbf{SD} & Standard Definition \\
\textbf{SNR} & Signal-to-Noise Ratio \\
\textbf{SWIR} & Short-Wave Infrared \\
\textbf{UHD} & Ultra High Definition \\

\end{tabular}

\end{document}